\documentclass{article}
\pdfoutput=1
\usepackage[final]{colm2026_conference}

\usepackage{algorithm}
\usepackage{algpseudocode}
\usepackage{xcolor}

\definecolor{patchblue}{RGB}{46,92,180}
\definecolor{patchteal}{RGB}{0,120,120}
\definecolor{patchgray}{RGB}{95,95,95}

\algrenewcommand\algorithmicrequire{\textbf{Input:}}
\algrenewcommand\algorithmicensure{\textbf{Output:}}
\algrenewcommand\algorithmiccomment[1]{\hfill\textcolor{patchgray}{\(\triangleright\)~#1}}

\newcommand{\PatchAdd}[1]{\State \textcolor{patchblue}{\textbf{+}~#1}}
\newcommand{\PatchComment}[1]{\textcolor{patchteal}{\emph{#1}}}

\usepackage[T1]{fontenc}
\usepackage[utf8]{inputenc}
\usepackage{kotex}

\usepackage{hyperref}
\usepackage{url}
\usepackage{booktabs}
\usepackage{lineno}

\usepackage{graphicx}
\usepackage{enumitem}
\usepackage{array}
\usepackage{amsmath}
\usepackage{float}
\usepackage{pifont}
\usepackage{wrapfig}
\usepackage{caption}
\usepackage{subcaption}
\usepackage[section]{placeins}
\usepackage[most]{tcolorbox}

\definecolor{darkblue}{rgb}{0, 0, 0.5}
\hypersetup{colorlinks=true, citecolor=darkblue, linkcolor=darkblue, urlcolor=darkblue}

\title{Headroom-Drift Replay: A Primitive for Principled Replay Control in GRPO}
\author{
Hyun Bin Park \& Du-Seong Chang \\
Department of Artificial Intelligence, Sogang University \\
Seoul, Republic of Korea \\
\texttt{tjrtkroal@sogang.ac.kr} \quad \texttt{duseong.chang@gmail.com}
}

\NewDocumentCommand{\replayfigure}{O{} m m m}{%
  \begin{figure}[htbp]
    \centering
    \IfFileExists{#2}{%
      \includegraphics[width=#3\linewidth]{#2}%
    }{%
      \fbox{\parbox[c][0.22\textheight][c]{0.9\linewidth}{\centering
      Missing figure file:\\ \texttt{#2}}}%
    }
    \caption{#4}
    \if\relax\detokenize{#1}\relax\else\label{#1}\fi
  \end{figure}
}

\newtcolorbox{responseblock}{
  enhanced,
  breakable,
  colback=white,
  colframe=black!65,
  boxrule=0.4pt,
  arc=1.2mm,
  left=2mm,
  right=2mm,
  top=1.5mm,
  bottom=1.5mm
}

\newtcblisting{rawresponse}{
  enhanced,
  breakable,
  listing only,
  colback=black!1,
  colframe=black!40,
  boxrule=0.3pt,
  arc=0.8mm,
  left=1.5mm,
  right=1.5mm,
  top=1mm,
  bottom=1mm
}

\newtcolorbox{rqbox}{
  enhanced,
  breakable,
  colback=blue!2,
  colframe=blue!45!black,
  boxrule=0.5pt,
  arc=1mm,
  left=2mm,
  right=2mm,
  top=1mm,
  bottom=1mm
}

\newcommand{\best}[1]{\textbf{#1}}
\newcommand{\second}[1]{\underline{#1}}

\newcolumntype{L}[1]{>{\raggedright\arraybackslash}p{#1}}

\begin{document}

\ifcolmsubmission
\linenumbers
\fi

\maketitle

\begin{abstract}
  RL-based post-training for reasoning models is increasingly bottlenecked by repeated fresh rollout generation, particularly in agentic settings where environment interaction dominates wall-clock cost. Replay can reduce this burden by reusing past trajectories, but existing methods typically embed it within larger training pipelines involving exploration, experience restructuring, or mixed-policy optimization. This makes replay's own contribution difficult to isolate. We ask a focused question: how far can principled replay selection alone go? We introduce Headroom-Drift Replay, a group-level replay control primitive for GRPO that separates reuse into two decisions. Headroom ranks stored groups by remaining learning value, while Drift gates them by compatibility with the current policy. The fresh on-policy stream remains unchanged, and the method adds no auxiliary generation or training machinery. Across mathematical reasoning, multimodal reasoning, and Agentic Search benchmarks, this single intervention outperforms naive replay and matches or exceeds broader replay methods on Avg Mean@32. In Agentic Search, where environment interaction dominates cost, it delivers comparable quality at materially lower wall-clock time.
\end{abstract}

\begin{figure}[htbp]
  \centering
  \includegraphics[width=\linewidth]{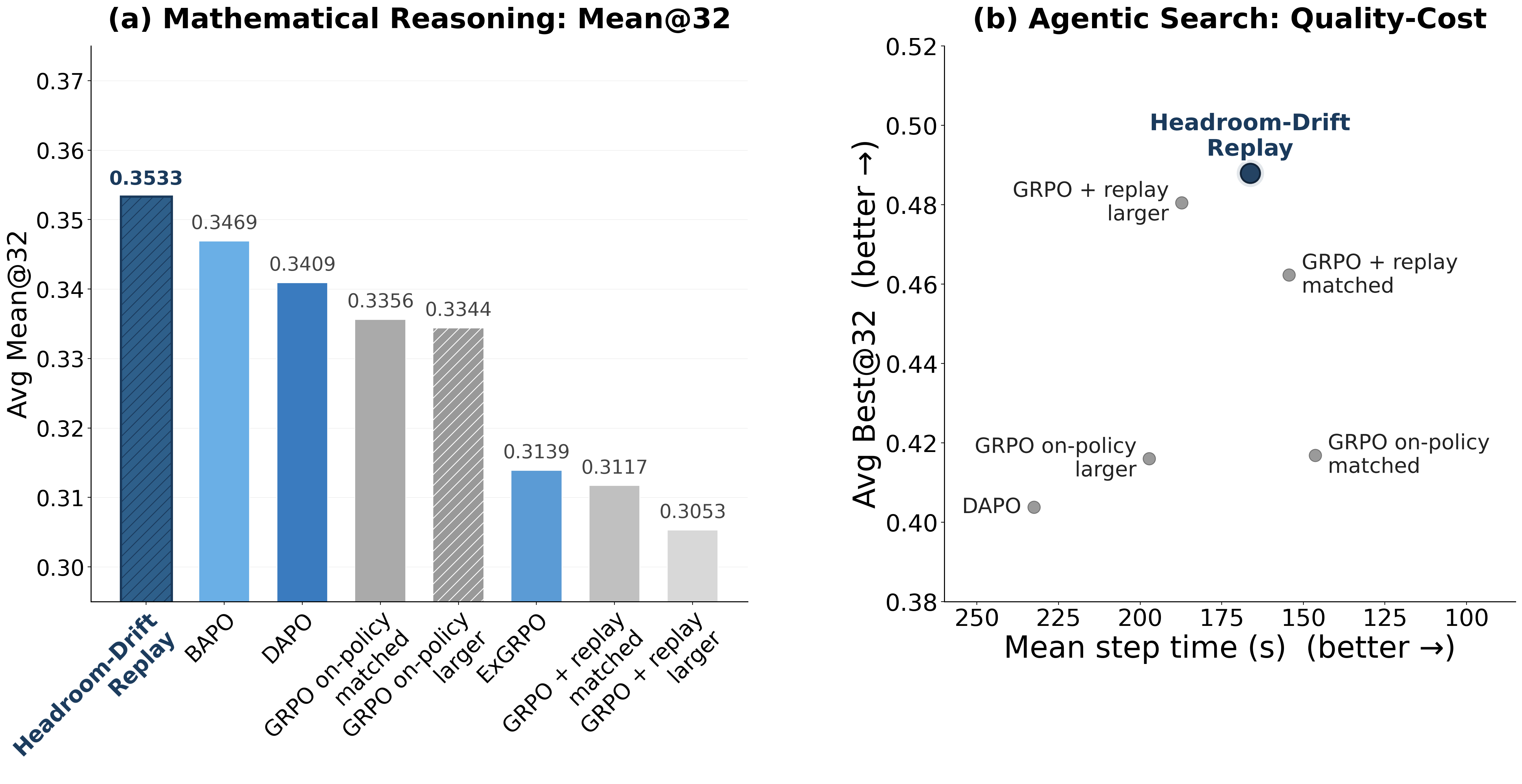}
  \caption{Summary of two representative views. (a) In mathematical reasoning, Headroom-Drift Replay leads on Avg Mean@32 over all baselines. (b) In agentic search, Headroom-Drift combines the highest Avg Best@32 with competitive wall-clock cost.}
  \label{fig:main_hero}
\end{figure}

\section{Introduction}

Reinforcement learning (RL) has become a central ingredient in post-training language models for reasoning~\citep{ouyang2022instructgpt,bai2022constitutional}, but it often incurs substantial cost through repeated fresh rollout generation. This burden is especially high in GRPO-style training and grows further in agentic settings, where interaction with external environments adds significant wall-clock overhead~\citep{igpo2026,istar2026,gem2026}. Replay offers a natural remedy by reusing previously collected trajectories. Yet naive reuse is often insufficient for stable learning. The question is not whether to replay, but how to control replay itself: which stored experience should re-enter training, and under what conditions.

A stored replay group can fail for two distinct reasons. It may still contain useful learning signal but have become too stale under the current policy. Conversely, it may remain close to the current policy while offering little room for further learning. Reliable replay therefore requires two separate judgments: whether a group still offers learning value and whether it remains compatible with the current policy.

Prior work has shown that replay can improve sample efficiency in reasoning-oriented RL, but existing methods typically couple it with auxiliary mechanisms. This makes replay itself difficult to study as a standalone control problem. We therefore isolate replay-side control within GRPO~\citep{shao2024deepseekmath,guo2025deepseekr1} and ask how far it can go on its own.

Headroom-Drift Replay is a group-level replay control primitive for GRPO built around separate judgments of learning value and current-policy compatibility. Headroom measures the remaining room to shift probability toward good actions and away from bad ones. Policy Drift measures how far the current policy has moved from the policy that generated the group and gates out stale groups. Reuse operates on full groups rather than individual responses, preserving the within-group comparison structure central to GRPO. Together, these two controls form the only addition to standard GRPO; the fresh on-policy stream remains unchanged. This design isolates the effect of replay-side control on training dynamics.

Empirically, Headroom-Drift outperforms naive replay across mathematical reasoning, multimodal reasoning, and Agentic Search, and matches or exceeds methods that combine replay with additional training machinery. The strongest evidence comes from mathematical reasoning under the fullest baseline set. In Agentic Search, where environment interaction dominates wall-clock cost, the benefit takes a different form: Headroom-Drift outperforms on-policy scaling in both quality and per-step efficiency because the selected trajectories remain informative enough to substitute for costly fresh environment interaction. Multimodal reasoning shows that the same control principle extends beyond text-only settings.

Our contributions are threefold:\par
\noindent\textbf{(i)} We formulate replay in GRPO as a two-part control problem---retaining learning value while ensuring current-policy compatibility---and instantiate it in Headroom-Drift Replay, a group-level replay control primitive that requires no auxiliary training machinery.\par
\noindent\textbf{(ii)} We construct role-aligned baselines that isolate distinct control questions---on-policy budget matching, fresh-data scaling, naive replay volume, strong non-replay alternatives, and broader replay methods with auxiliary machinery---and show across mathematical reasoning, Agentic Search, and multimodal reasoning that this primitive outperforms naive replay and matches or exceeds broader replay methods.\par
\noindent\textbf{(iii)} We analyze how principled replay control behaves across training, covering same-buffer counterfactual comparisons, replay-age compatibility and lifetime exposure concentration, multi-reward ingress dynamics, and the relationship between replay and entropy collapse.

\section{Related work}

Classical reinforcement learning motivates the two control axes in our design. Prioritized Experience Replay showed that stored samples differ substantially in learning value, and that replay is more effective when guided by expected training utility rather than random selection~\citep{schaul2016per}. A complementary line of off-policy correction methods addresses policy mismatch, showing that past experience can remain useful when reuse is appropriately weighted or truncated under the current policy~\citep{munos2016retrace,espeholt2018impala}. These works operate on transitions or trajectories in classical RL rather than on groups in LLM post-training, but they establish the same two requirements that drive our design: replay should account for both learning value and current-policy compatibility. Headroom provides a PER-style learning-value priority over full GRPO groups; Headroom-Drift applies this priority only among groups admitted by the Drift-based current-policy compatibility gate.

In reasoning-oriented LLM training, replay has moved from a peripheral heuristic to a more central optimization component. RePO~\citep{li2025repo}, for example, incorporates replay into the GRPO training loop to improve sample efficiency beyond fully on-policy updates. This shift confirms that replay carries real value in modern reasoning settings where rollout generation is expensive.

Several recent methods strengthen replay by coupling it with broader training mechanisms. EFRame~\citep{wang2025eframe} combines replay with exploration and filtering to determine which trajectories are retained and revisited. ExGRPO~\citep{zhan2026exgrpo} organizes reasoning trajectories by correctness and entropy signals before mixed-policy reuse. BAPO~\citep{wan2026buffermatters} adopts a buffer-centric off-policy design that pairs historical hard samples with adaptive batch construction. These methods reinforce the value of replay in modern reasoning RL, but because replay is embedded within larger pipelines, the two control questions central to our work---which stored groups still offer learning value, and which remain compatible with the current policy---are rarely made explicit as separable design choices.

Our work treats these two questions as the primary intervention. The goal is not to replace broader replay-centric designs, but to provide a composable primitive within that larger design space: one that can stand alone and whose effects can be independently evaluated.

\section{Method}

\begin{figure}[!htbp]
  \centering
  \includegraphics[width=\linewidth,height=0.32\textheight,keepaspectratio]{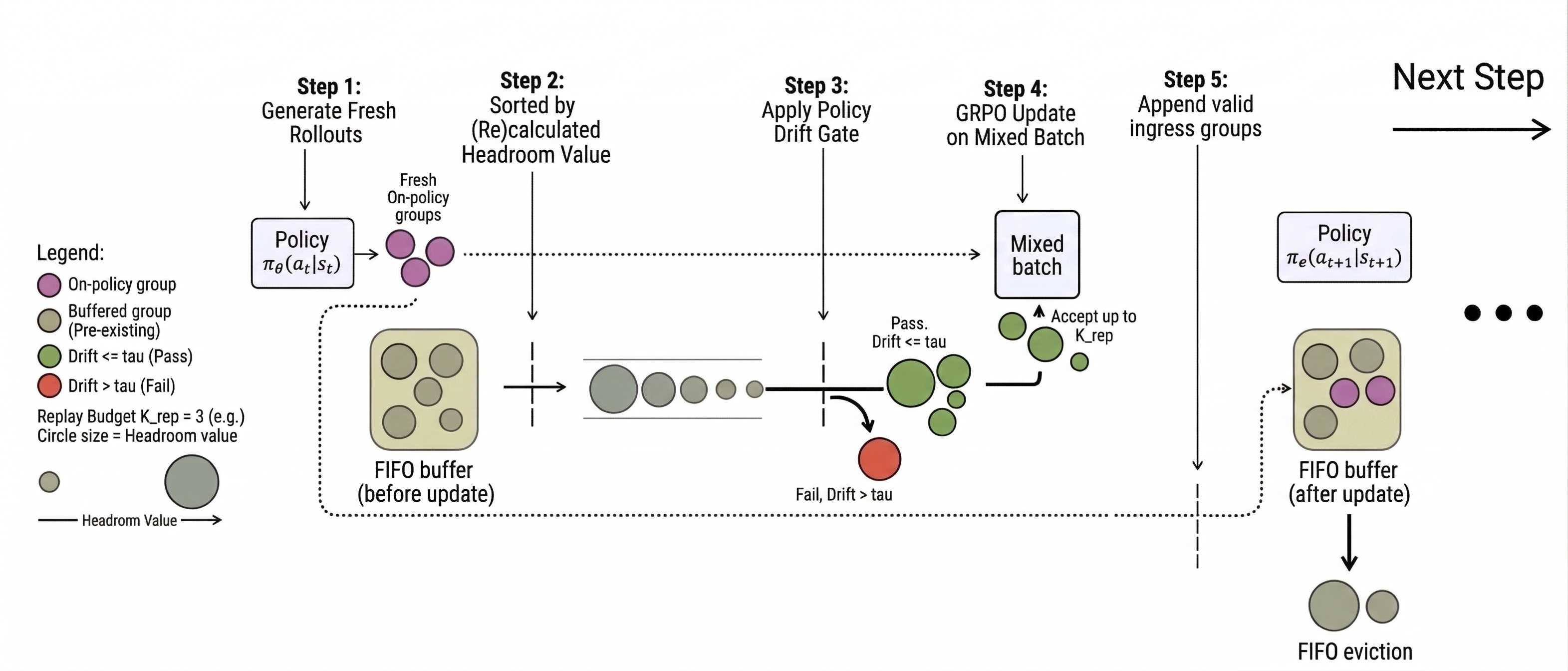}
  \caption{Headroom-Drift Replay in one GRPO step. The current policy generates fresh on-policy groups. Pre-existing buffered groups are ordered by Headroom and filtered by the Policy Drift gate; up to \(K_{\mathrm{rep}}\) accepted replay groups are merged with the fresh groups to form the mixed batch for the GRPO update. After the update, fresh groups with non-identical rewards are appended to the FIFO replay buffer.}
  \label{fig:main_figure}
\end{figure}

\subsection{Method overview.}
Headroom-Drift Replay is a group-level replay control primitive for GRPO that decomposes reuse into two explicit control axes. Rather than treating reuse as a single inclusion decision, it separates learning-value prioritization from current-policy compatibility control through Headroom-based ranking and Policy-Drift gating. The minimum unit of reuse is the full group rather than individual responses, preserving the within-group comparison structure central to GRPO. The fresh on-policy stream is left untouched: the only modification to standard GRPO is that selected replay groups are added to the mixed batch. Any change in training dynamics therefore traces to replay-side control alone.

\subsection{GRPO setup and replay state.}
We first specify the GRPO state needed to express replay groups within the same update form as fresh on-policy groups. For a prompt $Q$, the current policy $\pi_\theta$ generates $n$ responses $G_1,\dots,G_n$, which together form a group $g=(Q,G_1,\dots,G_n)$. Let $\mathcal{T}(g)$ denote the set of token positions in the group, and let $a_{i,j}$ and $s_{i,j}$ denote the $j$-th generated token in response $G_i$ and its conditioning context. The current policy assigns token probability $\pi_\theta(a_{i,j}\mid s_{i,j})$. GRPO computes rewards at the response level and derives a group-relative advantage $A_i$ for each response $G_i$; this same $A_i$ is then applied to all token positions belonging to that response.

What distinguishes replay from the fresh on-policy case is that each stored group is paired with the policy state under which it was originally generated. We denote this reference policy by $\pi_{\mathrm{gen}(g)}$. For a freshly generated group, $\pi_{\mathrm{gen}(g)}$ coincides with the usual rollout-time old policy in GRPO. For a replayed group, however, it is the generation-time policy stored with that group in the buffer. Accordingly, the replay buffer maintains immutable per-group reference state: the stored actions, their generation-time log-probabilities under $\pi_{\mathrm{gen}(g)}$, and the response-level advantages $A_i$. This frozen reference state anchors each replayed group to the rollout under which it was originally collected.

Using this reference state, both fresh and replayed groups can be handled within the same GRPO update form. For a group $g$, the token-level importance ratio is
\[
  r_{i,j}(\theta;g)=\frac{\pi_\theta(a_{i,j}\mid s_{i,j})}{\pi_{\mathrm{gen}(g)}(a_{i,j}\mid s_{i,j})},
\]
so the difference between fresh and replayed groups lies only in which reference policy appears in the denominator.

\subsection{Headroom: learning-value prioritization.}\label{subsec:replay-components}
Headroom provides the learning-value signal in our replay-side control primitive. It measures the remaining directional correction room of a stored group under a reference policy. Intuitively, replay is more useful when positively advantaged stored actions still have room for probability increase, or negatively advantaged stored actions still have room for probability decrease. For a policy $\pi$ and a stored token position $(i,j)$, define the token-level Headroom contribution by
\[
  h_{i,j}(\pi;g)=
  \begin{cases}
    1-\pi(a_{i,j}\mid s_{i,j}), & A_i>0,\\
    \pi(a_{i,j}\mid s_{i,j}), & A_i<0,\\
    0, & A_i=0,
  \end{cases}
\]
and aggregate it over the group as
\[
  \mathrm{Headroom}(g;\pi)=\frac{1}{|\mathcal{T}(g)|}\sum_{(i,j)\in\mathcal{T}(g)} h_{i,j}(\pi;g).
\]
When evaluated under the generation-time policy $\pi_{\mathrm{gen}(g)}$, Headroom defines the value-side replay priority stored for group $g$.

\subsection{Policy drift: current-policy compatibility control.}
Policy Drift provides the current-policy compatibility control in our replay-side control primitive. A stored group may still have high Headroom while already being too far from the current policy for reliable reuse~\citep{munos2016retrace,espeholt2018impala}.

To measure this mismatch for a stored group $g$, define the token-level log-probability drift on the stored action by
\[
  \Delta_{i,j}(g;\theta)=\log \pi_\theta(a_{i,j}\mid s_{i,j})-\log \pi_{\mathrm{gen}(g)}(a_{i,j}\mid s_{i,j}).
\]
We then aggregate these tokenwise deviations into a group-level Policy Drift score by averaging their squared values over the group:
\[
  \mathrm{Drift}(g;\theta)=\frac{1}{|\mathcal{T}(g)|}\sum_{(i,j)\in \mathcal{T}(g)} \Delta_{i,j}(g;\theta)^2.
\]
A group is replayed only if its Policy Drift does not exceed a threshold $\tau$.

This choice is motivated by the role of Policy Drift as a replay-compatibility gate rather than a full distribution-distance objective. GRPO commonly monitors policy deviation through approximate-KL proxies~\citep{guo2025deepseekr1,yu2025dapo} built from sampled log-ratios on generated tokens. For replay gating, however, what matters is mismatch magnitude on the stored token sequence, not a signed average: if some tokenwise shifts are positive while others are negative, they can offset each other in the average even when the overall sequence has moved substantially under the current policy. We therefore use a non-cancelling aggregation of tokenwise drift, and adopt a squared (L2-style) form so that larger local mismatches are penalized more strongly. An L1 variant yields a tighter formal bound but is less sensitive to concentrated per-token mismatch; in practice, the L2 gate achieves higher late-stage validation while filling less of the replay budget, indicating that it wins through better-compatible selection rather than replay volume (Appendix~\ref{app:l1-vs-l2-ablation}). This squared aggregation also provides a formal link between the two control axes: when a group passes the Drift gate, its current-policy Headroom can deviate from the stored Headroom priority by at most $\sqrt{\tau}$, so the gate directly limits how much staleness can distort the stored Headroom priority (proof in Appendix~\ref{app:headroom-drift-formal-v2}, Proposition~2).

\subsection{One GRPO step with replay-side control}

\newcommand{\ParallelNote}[1]{\textcolor{orange!85!red}{#1}}

\begin{algorithm}[t]
  \caption{\textsc{Headroom-Drift Replay} as a replay-side control augmentation to standard GRPO.
  Blue lines mark replay-specific additions; the fresh on-policy rollout path remains unchanged.}
  \label{alg:hdr_grpo_patch}
  \small
  \begin{algorithmic}[1]
    \Require current policy $\pi_{\theta_t}$, prompt batch $\mathcal{X}_t$, replay buffer $\mathbf{B}_t$, replay budget $K_{\mathrm{rep}}$, drift threshold $\tau$
    \For{each training step $t$}
    \State Sample prompt batch $\mathcal{X}_t$
    \State Generate grouped fresh rollouts under $\pi_{\theta_t}$
    \State Compute rewards and group-relative advantages for the fresh groups
    \PatchAdd{Identify mixed-outcome fresh groups as replay ingress candidates $\mathcal{I}_t$}
    \PatchAdd{Order pre-existing buffered groups by \emph{Headroom} (descending)}
    \For{each candidate group $g$ in Headroom order}
    \If{$|\mathcal{R}_t| \ge K_{\mathrm{rep}}$}
    \State \textbf{break}
    \EndIf
    \PatchAdd{Reevaluate the stored actions of $g$ under $\pi_{\theta_t}$ \Comment{\PatchComment{single forward pass on fixed trajectories}}}
    \PatchAdd{Compute \emph{Policy Drift} and refresh \emph{Headroom} for $g$ from the same log-probability pass}
    \PatchAdd{\textbf{if} Policy Drift$(g) \le \tau$ \textbf{then} add $g$ to $\mathcal{R}_t$}
    \EndFor
    \PatchAdd{Form mixed actor batch $\mathcal{A}_t \gets \mathcal{G}^{\mathrm{on}}_t \cup \mathcal{R}_t$}
    \State Run the standard GRPO update on $\mathcal{A}_t$
    \PatchAdd{Append ingress groups $\mathcal{I}_t$ to $\mathbf{B}_t$ after selection \Comment{\PatchComment{no same-step replay}}}
    \PatchAdd{If capacity is exceeded, evict the oldest groups from $\mathbf{B}_t$ (FIFO)}
    \EndFor
  \end{algorithmic}
\end{algorithm}

\noindent\textbf{Phase A: Generate fresh groups and identify replay ingress candidates.}
At each training step, the current policy generates fresh grouped rollouts, and rewards and group-relative advantages are computed as in standard GRPO. Replay ingress candidates \(\mathcal{I}_t\) are then identified from these fresh groups.\par
\noindent\textbf{Phase B: Order pre-existing buffered groups by Headroom.}
The pre-existing buffered groups are retrieved and ordered by Headroom, which defines the value-side priority for replay candidate scanning.\par
\noindent\textbf{Phase C: Apply the Policy Drift gate.}
The Headroom-ordered candidates are scanned under the current policy. For each scanned group, the model reevaluates the group via parallel teacher-forced log-probability computation, and admits the group for replay only if its Policy Drift does not exceed \(\tau\). Scanning stops once the replay budget \(K_{\mathrm{rep}}\) is filled or no further admissible groups remain.\par
\noindent\textbf{Phase D: Form the mixed batch and run the GRPO update.}
The accepted replay groups are merged with the fresh on-policy groups to form the mixed batch for the GRPO update.\par
\noindent\textbf{Phase E: Append valid ingress groups and maintain the FIFO buffer.}
To prevent same-step replay, the previously identified ingress candidates \(\mathcal{I}_t\) are appended to the replay buffer only after the GRPO update. The buffer is maintained as a fixed-capacity FIFO queue, so once capacity is exceeded, the oldest groups are evicted regardless of Policy Drift or Headroom.\par
\noindent\textbf{Operational note.} Replay requires no fresh autoregressive generation and no fresh environment interaction. Because reevaluation is teacher-forced on fixed stored token sequences, it can be parallelized over stored tokens. For efficiency, the current-policy log-probabilities computed for Policy Drift are reused to refresh Headroom in the same pass. Because scanning stops once the replay budget is filled, cached Headroom for unscanned groups can become stale. The two mechanisms therefore act complementarily: Policy Drift gating protects admission from overly stale groups, while same-pass Headroom refresh keeps the cached priority accurate for groups that are actually revisited. Appendix~\ref{app:formal-headroom-drift} makes the cached/reference/runtime Headroom distinction precise and formalizes replay selection in a form that directly mirrors the implementation above. Appendix~\ref{app:impl-details} details the engineering choices required for stable and efficient execution, including frozen reference-state management, replay-prefix batch layout, and sequence-length balancing across GPUs.

\section{Experiments}

\subsection{Experimental setup}

We evaluate Headroom-Drift Replay in three GRPO-style settings with fixed, verifiable rewards: mathematical reasoning, Agentic Search, and multimodal reasoning. For mathematical reasoning, we use AIME24, AMC23, MATH500, Minerva, and OlympiadBench. For Agentic Search, we evaluate on NQ, TriviaQA, PopQA, HotpotQA, 2WikiMultiHopQA, Musique, and Bamboogle. For multimodal reasoning, we report a compact three-benchmark view covering Geometry3K (Geo3K), MathVista, and MathVision. We use Mean@32 as the headline metric for the main cross-domain comparisons. For their cross-benchmark summaries, we report Avg Mean@32. Best@32 and Weighted Mean@32 are reported in Appendix~\ref{app:detailed-benchmark-tables} for completeness. We additionally include a 7B Agentic Search scale check.

Our baselines are chosen to answer distinct control questions, and we refer to them by role rather than by raw budget alone. GRPO on-policy matched keeps the fresh rollout budget aligned with replay and isolates the effect of replay reuse itself. GRPO on-policy larger tests whether replay gains can be explained by simple fresh-data scaling. GRPO + replay matched and, where available, GRPO + replay larger test whether buffered reuse alone is sufficient, or whether principled replay control is necessary. DAPO serves as a strong non-replay comparator, and where directly comparable we also include broader replay/off-policy methods such as ExGRPO and BAPO. In mathematical reasoning, we include the full comparison set. In multimodal reasoning, we report GRPO on-policy matched, GRPO on-policy larger, and DAPO in a simplified three-benchmark view. In Agentic Search, the main role-aligned comparisons are GRPO on-policy matched, GRPO on-policy larger, GRPO + replay matched, and DAPO. We do not include ExGRPO or BAPO in this setting because their published implementations target prompt-conditioned reasoning trajectories rather than externally interleaved tool-calling trajectories, and a like-for-like adaptation would require substantial re-engineering rather than a clean baseline comparison.

Because replay cost varies across settings---negligible in single-turn mathematical reasoning but substantial in Agentic Search where environment interaction dominates wall-clock time---we interpret Mean@32 together with acquisition cost. Table~\ref{tab:search_timing} reports mean per-step wall-clock time as the main cost-side comparator.

\subsection{Evaluation across domains}

\begin{figure*}[!t]
  \centering
  \includegraphics[width=0.95\linewidth]{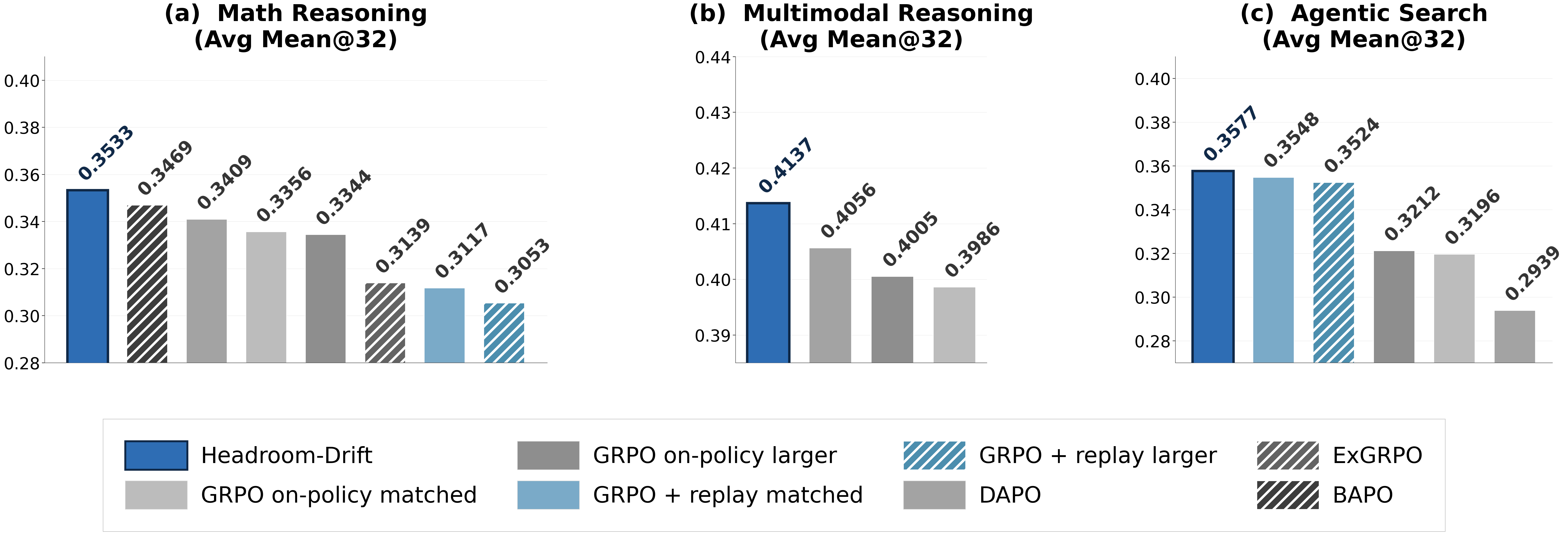}
  \caption{Cross-domain summary with Mean@32.}
  \label{fig:exp_overview}
\end{figure*}

\subsubsection{Mathematical reasoning}

Figure~\ref{fig:exp_overview} reports the mathematical-reasoning held-out results under the fullest comparison set. Headroom-Drift leads on Avg Mean@32 across all baselines. Relative to GRPO on-policy matched, the improvement confirms that principled replay adds value beyond purely on-policy training at matched fresh-rollout budget. Relative to GRPO on-policy larger, Headroom-Drift still leads despite using fewer fresh responses, ruling out simple fresh-data scaling as an explanation. Against GRPO + replay matched and GRPO + replay larger, the margin widens further, indicating that replay quality depends on principled subset selection rather than on replay volume. Among the broader comparators, DAPO, ExGRPO, and BAPO are all surpassed on Avg Mean@32; this is the one domain where the primitive matches or exceeds every baseline under a single headline metric.

\replayfigure[fig:math_training_score_curves]{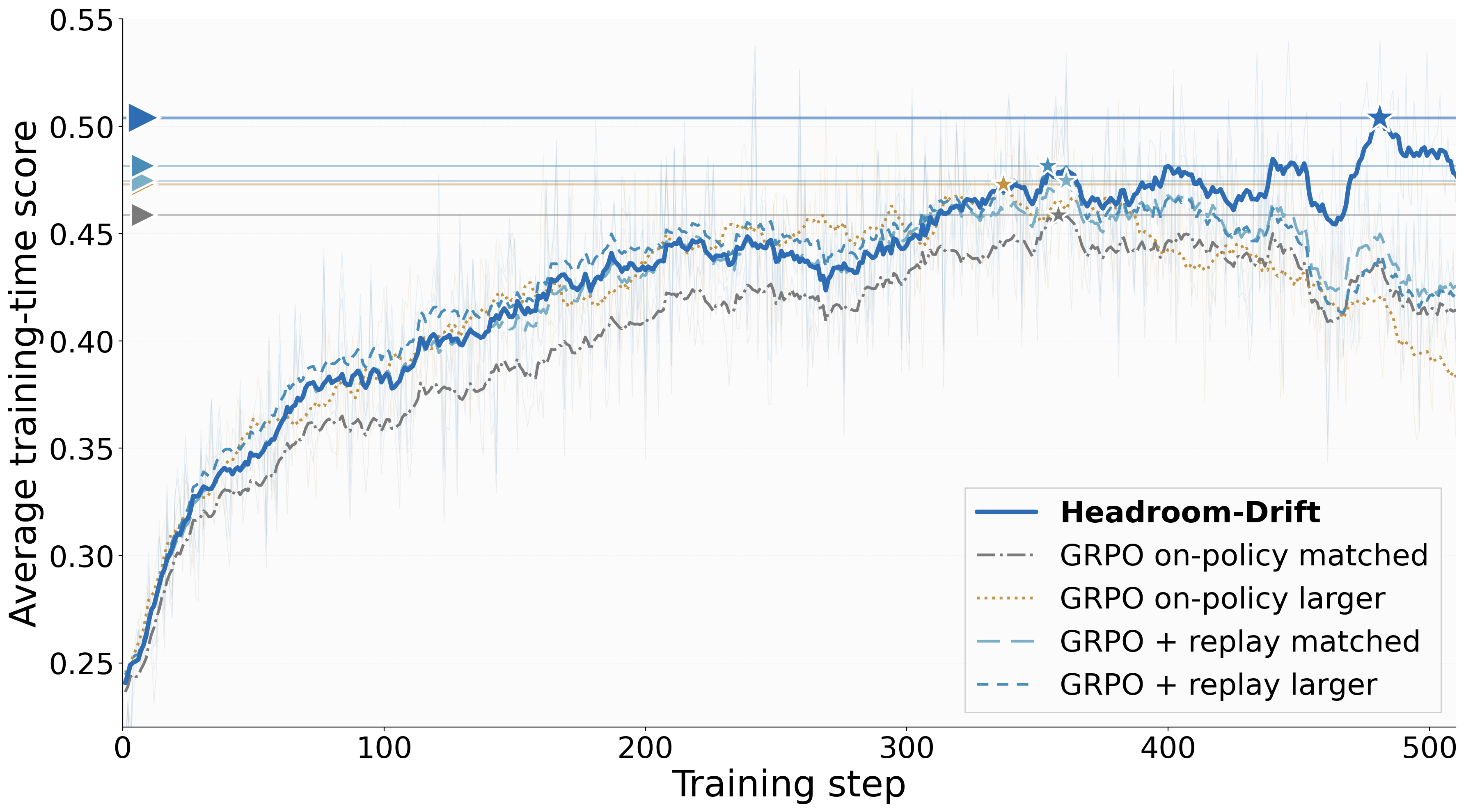}{0.92}{Training-score trajectories on mathematical reasoning.}

Figure~\ref{fig:math_training_score_curves} provides supporting training-dynamics evidence. The replay trajectory stays above GRPO on-policy matched over a broad mid-to-late window and tracks above GRPO on-policy larger through much of the same range, indicating that the Avg Mean@32 advantage is reflected in the training trajectory rather than confined to a single checkpoint. A training-score-matched analysis in Appendix~\ref{app:entropy-collapse} further shows that replay delays the onset of entropy collapse: at comparable score levels, the replay run enters the low-entropy regime later and remains there longer than GRPO on-policy larger, suggesting that principled replay contributes to more sustained learning dynamics.

\subsubsection{Agentic search}

Agentic search serves a different role from mathematical reasoning: the main question is not the largest standalone gain in Mean@32, but whether replay-side control remains worthwhile when training is dominated by multi-turn environment interaction cost.

Against naive replay baselines, the picture depends on the metric. On the headline Avg Mean@32, Headroom-Drift leads GRPO + replay matched by only 0.0029 (0.3577 vs.\ 0.3548), a gap within noise range. But on Avg Best@32, the margin is substantially larger (0.4879 vs.\ 0.4623, as shown in Figure~\ref{fig:main_hero}(b)), suggesting that principled selection shifts the upper tail of trajectory quality more than it shifts the average.

As an additional scale check, we conduct a controlled Search-R1 comparison between Headroom-Drift and GRPO + replay matched with Qwen2.5-7B-Instruct. To make 7B training and agentic evaluation feasible within the available compute budget, both methods use an 8-bit optimizer and we evaluate four samples per input; we therefore report Mean@4 and Best@4 rather than their @32 counterparts. Under this matched setup, Headroom-Drift improves Avg Mean@4 from 0.3737 to 0.3955 and Weighted Mean@4 from 0.4158 to 0.4298, leading on six of seven benchmarks. Full per-benchmark Mean@4 and Best@4 results are reported in Appendix~\ref{app:detailed-benchmark-tables}, Table~\ref{tab:search_results_7b}.

\begin{table}[H]
  \centering
  \small
  \setlength{\tabcolsep}{6pt}
  \renewcommand{\arraystretch}{1.05}
  \begin{tabular}{lr}
    \toprule
    Method & Step Time (s) \\
    \midrule
    GRPO on-policy matched (b128) & 146.4 \\
    GRPO + replay matched (b128+r64) & 154.4 \\
    Headroom-Drift (b128+r64) & 166.3 \\
    GRPO + replay larger (b128+r128) & 187.3 \\
    GRPO on-policy larger (b192) & 197.2 \\
    DAPO & 232.5 \\
    \bottomrule
  \end{tabular}
  \caption{Mean per-step wall-clock time on agentic search (Search-R1).}
  \label{tab:search_timing}
\end{table}

The more revealing comparison is against on-policy scaling. GRPO on-policy larger uses 1.5$\times$ the fresh rollout budget yet achieves lower Avg Mean@32 (0.3212 vs.\ 0.3577) at higher per-step cost (197.2\,s vs.\ 166.3\,s). Headroom-Drift surpasses it on both axes because principled selection substitutes reevaluation of stored trajectories for costly fresh environment interaction (Table~\ref{tab:search_timing}). This is a Pareto improvement over on-policy scaling.

\subsubsection{Multimodal reasoning}

Multimodal reasoning serves as breadth evidence. The question is whether the same replay-side control principle generalizes beyond text-only settings. In this simplified three-benchmark view, Headroom-Drift leads on Avg Mean@32 over GRPO on-policy matched, GRPO on-policy larger, and DAPO (0.4137 vs.\ 0.3986, 0.4005, and 0.4056 respectively).

\subsection{Mechanism analysis of replay selection}

\begin{figure}[t]

  \centering

  \includegraphics[width=\linewidth]{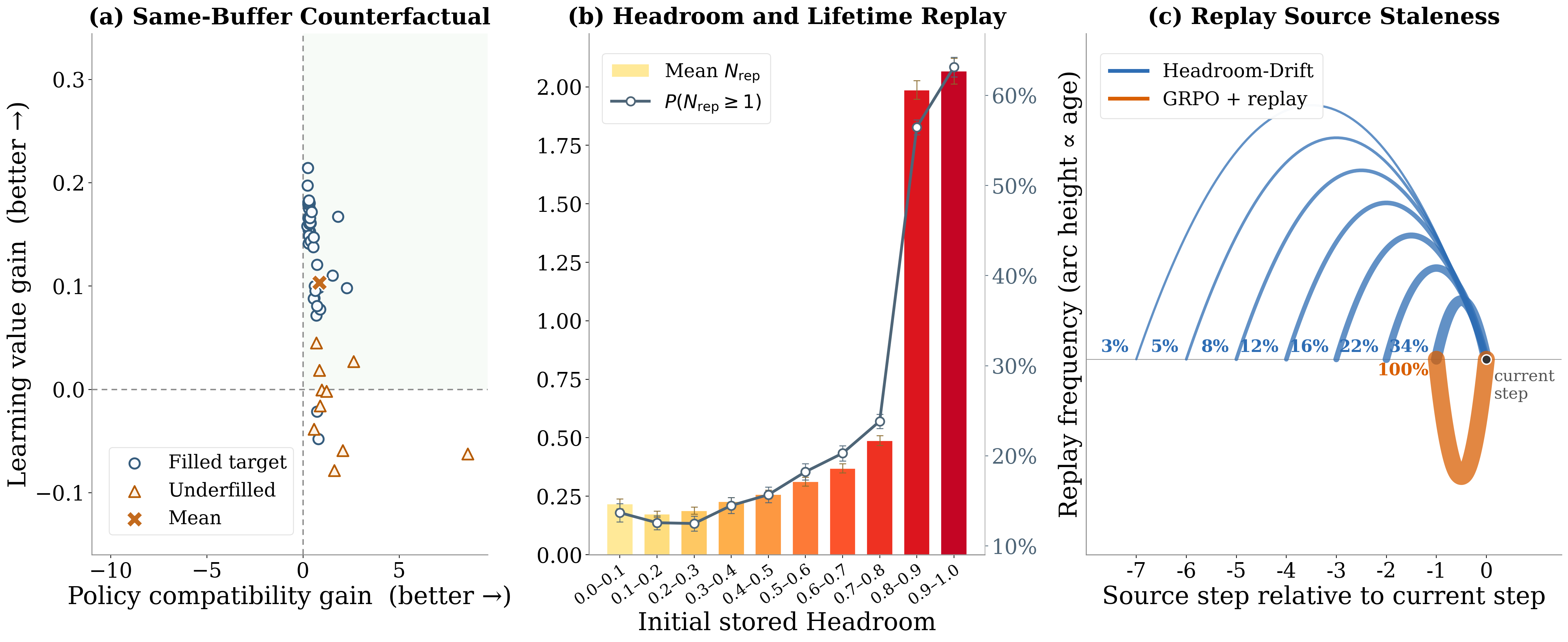}

  \caption{\textbf{(A)} Same-buffer counterfactual against GRPO + replay. \textbf{(B)} Initial stored Headroom and lifetime replay. Bars show mean replay count \(N_{\mathrm{rep}}\), and the line shows \(P(N_{\mathrm{rep}} \ge 1)\). \textbf{(C)} Conceptual view of replay source staleness under Headroom-Drift versus GRPO + replay.}

  \label{fig:replay_selection_in_practice}

\end{figure}

We now ask whether the two control axes operate as designed. Figure~\ref{fig:replay_selection_in_practice} examines this question through three complementary views: step-level subset quality, lifetime reuse concentration, and temporal diversity of replay sources.

Figure~\ref{fig:replay_selection_in_practice}(a) compares, at each training step, the replay subset actually selected by Headroom-Drift against the subset that GRPO + replay would select by recency alone, on matched pre-step buffer states and matched subset sizes. Each point plots the compatibility gain ($x$: positive means our subset has lower Policy Drift) against the learning-value gain ($y$: positive means higher current-policy Headroom). The majority of steps fall in the jointly improved quadrant, confirming systematic advantage on both axes. The gain is asymmetric: policy compatibility improvement is consistent, while learning-value improvement weakens in late-stage steps where the admissible pool contracts. This pattern is expected---Policy Drift acts as a hard gate, while Headroom prioritizes within the feasible set that the gate admits.

Figure~\ref{fig:replay_selection_in_practice}(b) shows that these step-level decisions accumulate into a sharp lifetime reuse pattern. Groups with initial stored Headroom below 0.8 are rarely replayed (mean $N_{\mathrm{rep}} < 0.5$; replay probability ${<}\,24\%$), while groups above 0.8 are replayed frequently (mean $N_{\mathrm{rep}} \approx 2.0$; ${>}\,56\%$). Headroom ranking preferentially surfaces high-value groups, but lower-Headroom groups are not excluded; they still enter replay when the Policy Drift gate judges them sufficiently compatible.

Figure~\ref{fig:replay_selection_in_practice}(c) reveals the temporal consequence. GRPO + replay draws exclusively from the immediately preceding step (100\% at age 1), since each step's buffer ingress exceeds the replay target. Headroom-Drift distributes replay across ages 1--7, with 44\% coming from sources three or more steps back. This multi-age mixture arises because Headroom ranking surfaces older high-value groups that still pass the Policy Drift gate, producing a temporally diverse corrective signal rather than uniform one-step reuse.

In practice, we calibrate the Drift threshold \(\tau\) with a short log-spaced sweep instead of a fine-grained full-training search. On Search-R1, three decade-spaced settings yield mean replay acceptance rates of \(6.4\%\), \(40.2\%\), and \(94.4\%\); only the middle setting filters selectively while keeping replay KL stable. Collapsing acceptance under a tight threshold and rising replay KL under a loose one provide early signals for ruling out unsuitable values without running every candidate to completion. Across the evaluated runs, the selected values fall into two task-family settings, and the Search-R1 threshold is reused from 3B to 7B without scale-specific retuning. Appendix~\ref{app:tau-sensitivity} reports the exact thresholds and full analysis.

Appendix~\ref{app:replay-age-reuse-stats} provides further empirical detail on these reuse patterns. Three findings stand out: age alone does not determine replay compatibility, as the Drift gate rejects a non-negligible fraction of even the youngest groups; scan depth grows over training as the admissible set contracts; and most buffered groups are never replayed, while a small repeatedly reused subset accounts for the majority of total replay exposure. A component ablation supports the role of both axes. On MATH-500, the full Headroom-Drift combination achieves the highest Mean@32, followed by Headroom-only (the PER-style priority-only analogue) and then Drift-only (Appendix~\ref{app:math-component-ablation}, Table~\ref{tab:math_ablation}).

\section{Conclusion and future directions}
We introduced Headroom-Drift Replay, a replay-side control primitive that decomposes reuse into learning-value prioritization (Headroom) and current-policy compatibility control (Drift). Across mathematical reasoning, agentic search, and multimodal reasoning, this single intervention outperforms naive replay. In mathematical reasoning under the fullest baseline set, it matches or exceeds broader replay methods on Avg Mean@32. In agentic search, it delivers a Pareto improvement over on-policy scaling. Replay, when controlled along the right axes, functions as a capable optimization primitive in its own right.

We studied this primitive in isolated form, but its two explicit control axes are designed to compose with the exploration, filtering, and adaptive batch construction strategies in recent replay-centric methods~\citep{wang2025eframe,zhan2026exgrpo,wan2026buffermatters}. The primitive does not compete with these methods; it provides a separable control layer within them. Preliminary training-score evidence on a CISPO-style objective suggests that the two-axis decomposition transfers beyond GRPO (Appendix~\ref{app:cispo-portability}), though held-out evaluation across objective families, including PPO-, GSPO-, and CISPO-style formulations~\citep{zheng2025gspo,minimax2025m1,yuan2026cappo,shen2026bapo}, remains open.

\clearpage

\section*{Acknowledgments}
This research was conducted as part of the Sovereign AI Foundation Model Project (GPU Track), organized by the Ministry of Science and ICT (MSIT) and supported by the National IT Industry Promotion Agency (NIPA), S. Korea (PJT-26-010017). This research was supported by the MSIT (Ministry of Science, ICT), Korea, under the Top-Tier AI Global HRD invitation program (RS-2025-25461932) supervised by the IITP (Institute for Information \& Communications Technology Planning \& Evaluation).

\section*{Ethics Statement}
This work studies replay selection mechanisms for reinforcement-learning post-training of reasoning models. The experiments use standard benchmark datasets and simulated training environments; no new human-subject data were collected. Potential risks include misuse of improved reasoning capability and increased compute consumption in large-scale training. To mitigate overclaiming, we report results with explicit scope limits and emphasize quality--cost trade-offs under the evaluated settings. During camera-ready preparation, the authors used a large language model to revise prose and assist with drafting. The authors reviewed the resulting text and take responsibility for all content in the paper.

\bibliography{main_final_refs}

\begin{thebibliography}{22}
\providecommand{\natexlab}[1]{#1}
\providecommand{\url}[1]{\texttt{#1}}
\expandafter\ifx\csname urlstyle\endcsname\relax
  \providecommand{\doi}[1]{doi: #1}\else
  \providecommand{\doi}{doi: \begingroup \urlstyle{rm}\Url}\fi

\bibitem[Bai et~al.(2022)Bai, Kadavath, Kundu, et~al.]{bai2022constitutional}
Yuntao Bai, Saurav Kadavath, Sandipan Kundu, et~al.
\newblock Constitutional {AI}: Harmlessness from {AI} feedback, 2022.
\newblock URL \url{https://arxiv.org/abs/2212.08073}.

\bibitem[Espeholt et~al.(2018)Espeholt, Soyer, Munos, Simonyan, Mnih, Ward,
  Doron, Firoiu, Harley, Dunning, Legg, and Kavukcuoglu]{espeholt2018impala}
Lasse Espeholt, Hubert Soyer, R{\'e}mi Munos, Karen Simonyan, Vlad Mnih, Tom
  Ward, Yotam Doron, Vlad Firoiu, Tim Harley, Iain Dunning, Shane Legg, and
  Koray Kavukcuoglu.
\newblock {IMPALA}: Scalable distributed deep-{RL} with importance weighted
  actor-learner architectures.
\newblock In \emph{Proceedings of the 35th International Conference on Machine
  Learning}, volume~80 of \emph{Proceedings of Machine Learning Research}, pp.\
   1407--1416, 2018.
\newblock URL \url{https://proceedings.mlr.press/v80/espeholt18a.html}.

\bibitem[Guo et~al.(2025)Guo, Yang, Zhang, Song, Wang, Zhu, Xu, Zhang, Ma, Bi,
  et~al.]{guo2025deepseekr1}
Daya Guo, Dejian Yang, Haowei Zhang, Junxiao Song, Peiyi Wang, Qihao Zhu,
  Runxin Xu, Ruoyu Zhang, Shirong Ma, Xiao Bi, et~al.
\newblock {DeepSeek-R1} incentivizes reasoning in {LLMs} through reinforcement
  learning.
\newblock \emph{Nature}, 645\penalty0 (8081):\penalty0 633--638, 2025.
\newblock \doi{10.1038/s41586-025-09422-z}.
\newblock URL \url{https://www.nature.com/articles/s41586-025-09422-z}.

\bibitem[Li et~al.(2025)Li, Zhou, Lam, Yang, and Lu]{li2025repo}
Siheng Li, Zhanhui Zhou, Wai Lam, Chao Yang, and Chaochao Lu.
\newblock {RePO}: Replay-enhanced policy optimization, 2025.
\newblock URL \url{https://arxiv.org/abs/2506.09340}.

\bibitem[Lipkin et~al.(2026)Lipkin, Petrenko, Chen, Wijmans, Cusumano-Towner,
  Giryes, and Kr{\"a}henb{\"u}hl]{petrenko2026entropy}
Ben Lipkin, Aleksei Petrenko, Kevin Chen, Erik Wijmans, Marco Cusumano-Towner,
  Raja Giryes, and Philipp Kr{\"a}henb{\"u}hl.
\newblock Entropy-preserving reinforcement learning.
\newblock In \emph{International Conference on Learning Representations
  (ICLR)}, 2026.
\newblock \doi{10.48550/arXiv.2603.11682}.
\newblock URL \url{https://iclr.cc/virtual/2026/poster/10010707}.

\bibitem[Liu et~al.(2026{\natexlab{a}})Liu, Wang, Wu, Huang, Li, Jiao, and
  Zhang]{istar2026}
Xiaoqian Liu, Ke~Wang, Yuchuan Wu, Fei Huang, Yongbin Li, Jianbin Jiao, and
  Junge Zhang.
\newblock Agentic reinforcement learning with implicit step rewards.
\newblock In \emph{International Conference on Learning Representations
  (ICLR)}, 2026{\natexlab{a}}.
\newblock \doi{10.48550/arXiv.2509.19199}.
\newblock URL \url{https://iclr.cc/virtual/2026/poster/10007373}.

\bibitem[Liu et~al.(2026{\natexlab{b}})Liu, Sims, Duan, Chen, Yu, Zhou, Xu,
  Xiong, Liu, Tan, Wang, Zhu, Shi, Yang, Shieh, Teh, Lee, and Lin]{gem2026}
Zichen Liu, Anya Sims, Keyu Duan, Changyu Chen, Simon Yu, Xiangxin Zhou,
  Haotian Xu, Shaopan Xiong, Bo~Liu, Chenmien Tan, Weixun Wang, Hao Zhu, Weiyan
  Shi, Diyi Yang, Michael~Qizhe Shieh, Yee~Whye Teh, Wee~Sun Lee, and Min Lin.
\newblock {GEM}: A gym for generalist {LLMs}.
\newblock In \emph{International Conference on Learning Representations
  (ICLR)}, 2026{\natexlab{b}}.
\newblock URL \url{https://openreview.net/forum?id=vsqQ1lG52a}.

\bibitem[Miao et~al.(2025)Miao, Zhang, Ding, Zhang, Zhang, and
  Tao]{miao2025energy}
Yuchun Miao, Sen Zhang, Liang Ding, Yuqi Zhang, Lefei Zhang, and Dacheng Tao.
\newblock The energy loss phenomenon in {RLHF}: A new perspective on mitigating
  reward hacking.
\newblock In \emph{Proceedings of the 42nd International Conference on Machine
  Learning}, volume 267 of \emph{Proceedings of Machine Learning Research},
  pp.\  44076--44105, 2025.
\newblock URL \url{https://proceedings.mlr.press/v267/miao25c.html}.

\bibitem[{MiniMax} et~al.(2025){MiniMax}, Chen, Li, Gong, Jiang, Fei, Yang,
  et~al.]{minimax2025m1}
{MiniMax}, Aili Chen, Aonian Li, Bangwei Gong, Binyang Jiang, Bo~Fei, Bo~Yang,
  et~al.
\newblock {MiniMax-M1}: Scaling test-time compute efficiently with lightning
  attention, 2025.
\newblock URL \url{https://arxiv.org/abs/2506.13585}.

\bibitem[Munos et~al.(2016)Munos, Stepleton, Harutyunyan, and
  Bellemare]{munos2016retrace}
R{\'e}mi Munos, Tom Stepleton, Anna Harutyunyan, and Marc~G. Bellemare.
\newblock Safe and efficient off-policy reinforcement learning.
\newblock In \emph{Advances in Neural Information Processing Systems},
  volume~29, 2016.
\newblock URL
  \url{https://proceedings.neurips.cc/paper/2016/hash/c3992e9a68c5ae12bd18488bc579b30d-Abstract.html}.

\bibitem[Ouyang et~al.(2022)Ouyang, Wu, Jiang, Almeida, Wainwright, Mishkin,
  Zhang, Agarwal, Slama, Ray, Schulman, Hilton, Kelton, Miller, Simens, Askell,
  Welinder, Christiano, Leike, and Lowe]{ouyang2022instructgpt}
Long Ouyang, Jeffrey Wu, Xu~Jiang, Diogo Almeida, Carroll Wainwright, Pamela
  Mishkin, Chong Zhang, Sandhini Agarwal, Katarina Slama, Alex Ray, John
  Schulman, Jacob Hilton, Fraser Kelton, Luke Miller, Maddie Simens, Amanda
  Askell, Peter Welinder, Paul~F. Christiano, Jan Leike, and Ryan Lowe.
\newblock Training language models to follow instructions with human feedback.
\newblock In \emph{Advances in Neural Information Processing Systems},
  volume~35, pp.\  27730--27744, 2022.
\newblock \doi{10.52202/068431-2011}.
\newblock URL
  \url{https://proceedings.neurips.cc/paper_files/paper/2022/hash/b1efde53be364a73914f58805a001731-Abstract-Conference.html}.

\bibitem[Schaul et~al.(2016)Schaul, Quan, Antonoglou, and
  Silver]{schaul2016per}
Tom Schaul, John Quan, Ioannis Antonoglou, and David Silver.
\newblock Prioritized experience replay.
\newblock In \emph{International Conference on Learning Representations
  (ICLR)}, 2016.
\newblock \doi{10.48550/arXiv.1511.05952}.
\newblock URL \url{https://arxiv.org/abs/1511.05952}.

\bibitem[Shao et~al.(2024)Shao, Wang, Zhu, Xu, Song, Bi, Zhang, Zhang, Li, Wu,
  and Guo]{shao2024deepseekmath}
Zhihong Shao, Peiyi Wang, Qihao Zhu, Runxin Xu, Junxiao Song, Xiao Bi, Haowei
  Zhang, Mingchuan Zhang, Y.~K. Li, Y.~Wu, and Daya Guo.
\newblock {DeepSeekMath}: Pushing the limits of mathematical reasoning in open
  language models, 2024.
\newblock URL \url{https://arxiv.org/abs/2402.03300}.

\bibitem[Wan et~al.(2026)Wan, Wang, Huang, and Sun]{wan2026buffermatters}
Xu~Wan, Yansheng Wang, Wenqi Huang, and Mingyang Sun.
\newblock {Buffer Matters}: Unleashing the power of off-policy reinforcement
  learning in large language model reasoning.
\newblock In \emph{International Conference on Learning Representations
  (ICLR)}, 2026.
\newblock \doi{10.48550/arXiv.2602.20722}.
\newblock URL \url{https://iclr.cc/virtual/2026/poster/10009473}.

\bibitem[Wang et~al.(2025)Wang, Wei, Zhang, Shao, Dan, Huang, Zhang, and
  Wang]{wang2025eframe}
Chen Wang, Lai Wei, Yanzhi Zhang, Chenyang Shao, Zedong Dan, Weiran Huang,
  Yuzhi Zhang, and Yue Wang.
\newblock {EFRame}: Deeper reasoning via exploration-filter-replay
  reinforcement learning framework, 2025.
\newblock URL \url{https://arxiv.org/abs/2506.22200}.

\bibitem[Wang et~al.(2026{\natexlab{a}})Wang, Dai, Ye, Gan, Yao, Deng, Wu, and
  Ying]{igpo2026}
Guoqing Wang, Sunhao Dai, Guangze Ye, Zeyu Gan, Wei Yao, Yong Deng, Xiaofeng
  Wu, and Zhenzhe Ying.
\newblock Information gain-based policy optimization: A simple and effective
  approach for multi-turn {LLM} agents.
\newblock In \emph{International Conference on Learning Representations
  (ICLR)}, 2026{\natexlab{a}}.
\newblock \doi{10.48550/arXiv.2510.14967}.
\newblock URL \url{https://iclr.cc/virtual/2026/poster/10007215}.

\bibitem[Wang et~al.(2026{\natexlab{b}})Wang, Xie, Zhang, Sun, Chen, Li, and
  Zhang]{liu2026entropy}
Shumin Wang, Yuexiang Xie, Wenhao Zhang, Yuchang Sun, Yanxi Chen, Yaliang Li,
  and Yanyong Zhang.
\newblock On the entropy dynamics in reinforcement fine-tuning of large
  language models, 2026{\natexlab{b}}.
\newblock URL \url{https://arxiv.org/abs/2602.03392}.

\bibitem[Xi et~al.(2026)Xi, Guo, Nan, Zhou, Shen, Chen, Liu, Huang, Zhang, Guo,
  Deng, Lei, Zheng, Wang, Zhang, Sun, Zheng, Yan, Gui, Zhang, and
  Huang]{shen2026bapo}
Zhiheng Xi, Xin Guo, Yang Nan, Enyu Zhou, Junrui Shen, Wenxiang Chen, Jiaqi
  Liu, Jixuan Huang, Zhihao Zhang, Honglin Guo, Xun Deng, Zhikai Lei, Miao
  Zheng, Guoteng Wang, Shuo Zhang, Peng Sun, Rui Zheng, Hang Yan, Tao Gui,
  Qi~Zhang, and Xuanjing Huang.
\newblock {BAPO}: Stabilizing off-policy reinforcement learning for {LLMs} via
  balanced policy optimization with adaptive clipping.
\newblock In \emph{International Conference on Learning Representations
  (ICLR)}, 2026.
\newblock URL \url{https://openreview.net/forum?id=jIeJJqG7dz}.

\bibitem[Yu et~al.(2025)Yu, Zhang, Zhu, Yuan, Zuo, Yue, Dai, Fan, Liu, Liu,
  et~al.]{yu2025dapo}
Qiying Yu, Zheng Zhang, Ruofei Zhu, Yufeng Yuan, Xiaochen Zuo, Yu~Yue, Weinan
  Dai, Tiantian Fan, Gaohong Liu, Juncai Liu, et~al.
\newblock {DAPO}: An open-source {LLM} reinforcement learning system at scale.
\newblock In \emph{Advances in Neural Information Processing Systems},
  volume~38, pp.\  113222--113244, 2025.
\newblock \doi{10.52202/085713-3775}.
\newblock URL
  \url{https://proceedings.neurips.cc/paper_files/paper/2025/hash/a4277440d50f1f15d2cb4c14f7e0c0d2-Abstract-Conference.html}.

\bibitem[Yuan et~al.(2026)Yuan, Wu, and Liu]{yuan2026cappo}
Dun Yuan, Di~Wu, and Xue Liu.
\newblock Escaping policy contraction: Contraction-aware {PPO} ({CaPPO}) for
  stable language model fine-tuning.
\newblock In \emph{International Conference on Learning Representations
  (ICLR)}, 2026.
\newblock URL \url{https://openreview.net/forum?id=vDlkJewkDu}.

\bibitem[Zhan et~al.(2026)Zhan, Li, Wang, Qu, Liu, Shao, Wong, and
  Cheng]{zhan2026exgrpo}
Runzhe Zhan, Yafu Li, Zhi Wang, Xiaoye Qu, Dongrui Liu, Jing Shao, Derek~F.
  Wong, and Yu~Cheng.
\newblock {ExGRPO}: Learning to reason from prior successes.
\newblock In \emph{International Conference on Learning Representations
  (ICLR)}, 2026.
\newblock \doi{10.48550/arXiv.2510.02245}.
\newblock URL \url{https://iclr.cc/virtual/2026/poster/10011339}.

\bibitem[Zheng et~al.(2025)Zheng, Liu, Li, Chen, Yu, Gao, Dang, Liu, Men, Yang,
  Zhou, and Lin]{zheng2025gspo}
Chujie Zheng, Shixuan Liu, Mingze Li, Xiong-Hui Chen, Bowen Yu, Chang Gao, Kai
  Dang, Yuqiong Liu, Rui Men, An~Yang, Jingren Zhou, and Junyang Lin.
\newblock Group sequence policy optimization, 2025.
\newblock URL \url{https://arxiv.org/abs/2507.18071}.

\end{thebibliography}
\bibliographystyle{colm2026_conference}

\clearpage

\appendix

\section{Formal interpretation, mixed-batch objective, and replay-step formalization}
\label{app:formal-headroom-drift}

This appendix makes precise the quantities and replay-step construction used by Headroom-Drift Replay. We first fix replay-state conventions, including the distinction between reference, cached, current-policy, and runtime Headroom. We then formalize the mixed-batch objective, define Headroom and Policy Drift, prove a reference-to-current Headroom mismatch bound, introduce lifetime replay exposure quantities, and finally write the replay-augmented GRPO step in set notation.

\subsection{Notation and replay-state conventions}
\label{app:notation-state-conventions}

\begin{table}[p]
  \centering
  \small
  \setlength{\tabcolsep}{3pt}
  \renewcommand{\arraystretch}{1.08}
  \begin{tabular}{L{0.18\linewidth}L{0.20\linewidth}L{0.54\linewidth}}
    \toprule
    \textbf{Symbol} & \textbf{Type} & \textbf{Meaning} \\
    \midrule
    \multicolumn{3}{c}{\textit{Core GRPO and replay state}} \\
    \midrule
    $g$ & group & grouped rollout associated with one prompt \\
    $\mathcal{T}(g)$ & index set & token positions contained in group $g$ \\
    $A_i$ & scalar & response-level group-relative advantage for response $i$ \\
    $\pi_{\mathrm{gen}(g)}$ & policy & generation-time reference policy that originally produced group $g$ \\
    $\pi_{\theta_t}$ & policy & current policy at training step $t$ \\
    $t_0(g)$ & integer & step at which group $g$ is freshly generated \\
    $\mathrm{age}_t(g)$ & scalar & age of buffered group $g$ at step $t$, defined as $t-t_0(g)$ \\
    \midrule
    \multicolumn{3}{c}{\textit{Headroom and Policy Drift}} \\
    \midrule
    $\mathrm{Headroom}(g;\pi)$ & scalar score & policy-conditioned remaining directional correction room of group $g$ under policy $\pi$ \\
    $H^{\mathrm{ref}}(g)$ & scalar score & reference Headroom $\mathrm{Headroom}(g;\pi_{\mathrm{gen}(g)})$ \\
    $H_t^{\mathrm{cur}}(g)$ & scalar score & conceptual current-policy Headroom $\mathrm{Headroom}(g;\pi_{\theta_t})$ \\
    $\widehat H_t(g)$ & scalar score & cached Headroom priority stored in the buffer at the \emph{beginning} of step $t$ and used for buffer ordering \\
    $H_t^{\mathrm{run}}(g)$ & scalar score & runtime Headroom actually computed for a group scanned under the current policy at step $t$ \\
    $r_t(g)$ & integer & last refresh step index of the cached Headroom of $g$, so that $\widehat H_t(g)=\mathrm{Headroom}(g;\pi_{\theta_{r_t(g)}})$ \\
    $\Delta_{i,j}(g;\theta_t)$ & scalar & token-level stored-action log-probability drift under the current policy \\
    $\mathrm{Drift}(g;\theta_t)$ & scalar score & group-level Policy Drift, i.e., squared-average tokenwise log-probability mismatch relative to $\pi_{\mathrm{gen}(g)}$ \\
    \midrule
    \multicolumn{3}{c}{\textit{Replay-step notation}} \\
    \midrule
    $\mathbf{B}_t$ & ordered queue & replay buffer at the beginning of step $t$ \\
    $\mathcal{G}_t^{\mathrm{on}}$ & set & freshly generated on-policy groups at step $t$ \\
    $\mathcal{I}_t$ & set & ingress candidate set from $\mathcal{G}_t^{\mathrm{on}}$ determined by the task-specific ingress rule \\
    $\mathcal{S}_t$ & ordered set & Headroom-ordered scan prefix actually reevaluated at step $t$ \\
    $\mathcal{F}_t(\tau)$ & set & Drift-feasible buffered groups satisfying $\mathrm{Drift}(g;\theta_t)\le \tau$ \\
    $\mathcal{R}_t$ & set & accepted replay groups at step $t$ \\
    $\mathcal{A}_t$ & batch-level collection & mixed actor batch combining fresh on-policy groups and accepted replay groups \\
    $K_{\mathrm{rep}}$ & scalar & replay budget per training step \\
    $\tau$ & scalar & Policy Drift threshold \\
    $C$ & scalar & replay-buffer capacity \\
    \midrule
    \multicolumn{3}{c}{\textit{Lifetime reuse and exposure}} \\
    \midrule
    $R_t(g)$ & indicator & equals $1$ if group $g$ is replayed at step $t$, else $0$ \\
    $N_T^{\mathrm{rep}}(g)$ & scalar & cumulative replay count of group $g$ up to step $T$ \\
    $N_T^{\mathrm{use}}(g)$ & scalar & total training uses of group $g$ up to step $T$, including its initial fresh use \\
    $W_T(g)$ & scalar & cumulative mixed-objective exposure weight of group $g$ up to step $T$ \\
    \bottomrule
  \end{tabular}
  \caption{Notation used in Appendix~\ref{app:formal-headroom-drift}. The key state distinction is between reference Headroom $H^{\mathrm{ref}}(g)$, cached Headroom $\widehat H_t(g)$ used for current-step ordering, conceptual current-policy Headroom $H_t^{\mathrm{cur}}(g)$, and runtime Headroom $H_t^{\mathrm{run}}(g)$ obtained only for groups actually scanned at step $t$.}
  \label{tab:notation-headroom-drift}
\end{table}

For any buffered group $g$ and step $t$, we distinguish four Headroom-related quantities. The \emph{reference Headroom} is the immutable generation-time score
\begin{equation}
  H^{\mathrm{ref}}(g)
  :=
  \mathrm{Headroom}(g;\pi_{\mathrm{gen}(g)}).
  \label{eq:reference-headroom-v2}
\end{equation}
The \emph{current-policy Headroom} is the conceptual value of the same policy-conditioned quantity evaluated at the current policy,
\begin{equation}
  H_t^{\mathrm{cur}}(g)
  :=
  \mathrm{Headroom}(g;\pi_{\theta_t}).
  \label{eq:current-headroom-v2}
\end{equation}
The \emph{cached Headroom} $\widehat H_t(g)$ is the priority stored in the buffer at the beginning of step $t$ and used to order replay candidates:
\begin{equation}
  \widehat H_t(g)
  :=
  \mathrm{Headroom}(g;\pi_{\theta_{r_t(g)}}),
  \label{eq:cached-headroom-v2}
\end{equation}
where $r_t(g)$ is the last step at which the cached priority of $g$ was refreshed. Finally, for a buffered group $g$ that is actually scanned under the current policy at step $t$, the \emph{runtime Headroom} is
\begin{equation}
  H_t^{\mathrm{run}}(g)
  :=
  \mathrm{Headroom}(g;\pi_{\theta_t}).
  \label{eq:runtime-headroom-v2}
\end{equation}
Thus, whenever $g$ is scanned at step $t$, we have
\begin{equation}
  H_t^{\mathrm{run}}(g)=H_t^{\mathrm{cur}}(g).
  \label{eq:runtime-equals-current}
\end{equation}
At insertion time, the cached priority is initialized at the generation-time value, so a newly buffered group starts with $\widehat H_t(g)=H^{\mathrm{ref}}(g)$. If a buffered group is scanned at step $t$, then its cached priority is refreshed \emph{after} selection using the runtime value from the same current-policy reevaluation pass; otherwise the cached value remains unchanged. Consequently, current-step replay ordering is governed by $\widehat H_t(g)$, not by a freshly recomputed current-policy Headroom over the entire buffer.

\subsection{Mixed-batch objective as constrained replay augmentation}
\label{app:mixed-batch-objective-v2}

Let $\mathcal{G}_t^{\mathrm{on}}$ denote the fresh on-policy groups generated at step $t$, and let $\mathcal{R}_t$ denote the accepted replay groups selected from the pre-existing buffer. The mixed actor batch is
\begin{equation}
  \mathcal{A}_t
  =
  \mathcal{G}_t^{\mathrm{on}} \uplus \mathcal{R}_t,
  \label{eq:mixed-actor-batch-new-v2}
\end{equation}
where $\uplus$ denotes disjoint union.

For any group $g$, let
\begin{equation}
  \ell_{\mathrm{GRPO}}\bigl(g;\theta,\pi_{\mathrm{gen}(g)}\bigr)
  \label{eq:group-grpo-loss-v2}
\end{equation}
be the usual group-level GRPO objective evaluated with the generation-time reference policy attached to $g$. The one-step mixed-batch objective is then
\begin{equation}
  \mathcal{L}_t^{\mathrm{mix}}(\theta)
  =
  \frac{1}{|\mathcal{A}_t|}
  \sum_{g\in\mathcal{A}_t}
  \ell_{\mathrm{GRPO}}\bigl(g;\theta,\pi_{\mathrm{gen}(g)}\bigr).
  \label{eq:mixed-objective-v2}
\end{equation}

Define the fresh and replay parts separately as
\begin{equation}
  \mathcal{L}_t^{\mathrm{on}}(\theta)
  =
  \frac{1}{|\mathcal{G}_t^{\mathrm{on}}|}
  \sum_{g\in\mathcal{G}_t^{\mathrm{on}}}
  \ell_{\mathrm{GRPO}}\bigl(g;\theta,\pi_{\mathrm{gen}(g)}\bigr),
  \label{eq:on-policy-objective-v2}
\end{equation}
\begin{equation}
  \mathcal{L}_t^{\mathrm{rep}}(\theta)
  =
  \frac{1}{|\mathcal{R}_t|}
  \sum_{g\in\mathcal{R}_t}
  \ell_{\mathrm{GRPO}}\bigl(g;\theta,\pi_{\mathrm{gen}(g)}\bigr)
  \qquad (|\mathcal{R}_t|>0),
  \label{eq:replay-objective-v2}
\end{equation}
with
\begin{equation}
  \alpha_t
  :=
  \frac{|\mathcal{G}_t^{\mathrm{on}}|}{|\mathcal{A}_t|}.
  \label{eq:alpha-t-v2}
\end{equation}

\paragraph{Proposition 1 (Exact mixed-objective decomposition).}
If $|\mathcal{R}_t|>0$, then
\begin{equation}
  \mathcal{L}_t^{\mathrm{mix}}(\theta)
  =
  \alpha_t\,\mathcal{L}_t^{\mathrm{on}}(\theta)
  +
  (1-\alpha_t)\,\mathcal{L}_t^{\mathrm{rep}}(\theta).
  \label{eq:mixed-decomposition-v2}
\end{equation}
If $|\mathcal{R}_t|=0$, then $\mathcal{L}_t^{\mathrm{mix}}(\theta)=\mathcal{L}_t^{\mathrm{on}}(\theta)$.

\emph{Proof.}
The identity follows immediately by splitting the sum in \eqref{eq:mixed-objective-v2} over the disjoint union \eqref{eq:mixed-actor-batch-new-v2} and normalizing by $|\mathcal{A}_t|$.

This decomposition makes explicit that replay augments rather than replaces the fresh on-policy stream: the fresh rollout path remains intact, while replay contributes an additional constrained term determined by buffer-side selection.

\subsection{Headroom and policy drift}
\label{app:headroom-drift-formal-v2}

For a stored group $g$ and any policy $\pi$, define the token-level Headroom contribution as
\begin{equation}
  h_{i,j}(\pi;g)
  =
  \begin{cases}
    1-\pi(a_{i,j}\mid s_{i,j}), & A_i>0,\\
    \pi(a_{i,j}\mid s_{i,j}), & A_i<0,\\
    0, & A_i=0,
  \end{cases}
  \label{eq:token-headroom-contribution-v2}
\end{equation}
and the group-level Headroom as
\begin{equation}
  \mathrm{Headroom}(g;\pi)
  =
  \frac{1}{|\mathcal{T}(g)|}
  \sum_{(i,j)\in\mathcal{T}(g)} h_{i,j}(\pi;g).
  \label{eq:headroom-definition-v2}
\end{equation}
Thus Headroom is a policy-conditioned measure of remaining directional correction room on the stored action sequence. The reference Headroom $H^{\mathrm{ref}}(g)$ is obtained by evaluating \eqref{eq:headroom-definition-v2} at the immutable generation-time policy.

To measure current-policy mismatch on the stored actions, define the token-level log-probability drift
\begin{equation}
  \Delta_{i,j}(g;\theta_t)
  :=
  \log \pi_{\theta_t}(a_{i,j}\mid s_{i,j})
  -
  \log \pi_{\mathrm{gen}(g)}(a_{i,j}\mid s_{i,j}),
  \label{eq:token-drift-definition-v2}
\end{equation}
and the group-level Policy Drift
\begin{equation}
  \mathrm{Drift}(g;\theta_t)
  :=
  \frac{1}{|\mathcal{T}(g)|}
  \sum_{(i,j)\in\mathcal{T}(g)}
  \Delta_{i,j}(g;\theta_t)^2.
  \label{eq:group-drift-definition-v2}
\end{equation}
Replay admissibility is controlled by the hard gate $\mathrm{Drift}(g;\theta_t)\le \tau$.

\paragraph{Remark (Policy Drift as a sampled log-ratio magnitude signal).}
The term $\Delta_{i,j}(g;\theta_t)$ is the tokenwise sampled log-ratio on the stored action sequence underlying approximate-KL-style diagnostics in GRPO-like training. For replay gating, what matters is mismatch magnitude on the stored sequence rather than a signed average that can cancel across tokens. This is why Policy Drift uses the non-cancelling squared aggregation in \eqref{eq:group-drift-definition-v2}.

\paragraph{Proposition 2 (Policy-Drift control of Headroom mismatch).}
For any stored group $g$ and current policy $\pi_{\theta_t}$,
\begin{equation}
  \left|
  H_t^{\mathrm{cur}}(g)-H^{\mathrm{ref}}(g)
  \right|
  \le
  \sqrt{\mathrm{Drift}(g;\theta_t)}.
  \label{eq:headroom-mismatch-bound-v2}
\end{equation}

\emph{Proof.}
By the definitions of $H_t^{\mathrm{cur}}(g)$ and $H^{\mathrm{ref}}(g)$,
\begin{align*}
  \left|H_t^{\mathrm{cur}}(g)-H^{\mathrm{ref}}(g)\right|
  &\le
  \frac{1}{|\mathcal{T}(g)|}
  \sum_{(i,j)\in\mathcal{T}(g)}
  \left|
  h_{i,j}(\pi_{\theta_t};g)-h_{i,j}(\pi_{\mathrm{gen}(g)};g)
  \right|.
\end{align*}
For any token $(i,j)$, the case definition in \eqref{eq:token-headroom-contribution-v2} implies
\begin{equation*}
  \left|
  h_{i,j}(\pi_{\theta_t};g)-h_{i,j}(\pi_{\mathrm{gen}(g)};g)
  \right|
  \le
  \left|
  \pi_{\theta_t}(a_{i,j}\mid s_{i,j})-
  \pi_{\mathrm{gen}(g)}(a_{i,j}\mid s_{i,j})
  \right|.
\end{equation*}
Now write
\begin{equation*}
  x=\log \pi_{\theta_t}(a_{i,j}\mid s_{i,j}),
  \qquad
  y=\log \pi_{\mathrm{gen}(g)}(a_{i,j}\mid s_{i,j}).
\end{equation*}
Because $x,y\le 0$ and the derivative of $u\mapsto e^u$ is at most $1$ on $(-\infty,0]$, the exponential map is $1$-Lipschitz on this domain. Hence,
\begin{equation*}
  \left|
  \pi_{\theta_t}(a_{i,j}\mid s_{i,j})-
  \pi_{\mathrm{gen}(g)}(a_{i,j}\mid s_{i,j})
  \right|
  \le
  |x-y|
  =
  \left|\Delta_{i,j}(g;\theta_t)\right|.
\end{equation*}
Substituting this bound and applying Cauchy--Schwarz gives
\begin{align*}
  \left|H_t^{\mathrm{cur}}(g)-H^{\mathrm{ref}}(g)\right|
  &\le
  \frac{1}{|\mathcal{T}(g)|}
  \sum_{(i,j)\in\mathcal{T}(g)}
  \left|\Delta_{i,j}(g;\theta_t)\right| \\
  &\le
  \sqrt{
    \frac{1}{|\mathcal{T}(g)|}
    \sum_{(i,j)\in\mathcal{T}(g)}
    \Delta_{i,j}(g;\theta_t)^2
  }
  =
  \sqrt{\mathrm{Drift}(g;\theta_t)}.
\end{align*}
This proves \eqref{eq:headroom-mismatch-bound-v2}.

\paragraph{Corollary 2.1 (Ranking certificate).}
Let $g$ and $h$ be two stored groups. If
\begin{equation}
  H^{\mathrm{ref}}(g)-H^{\mathrm{ref}}(h)
  >
  \sqrt{\mathrm{Drift}(g;\theta_t)}
  +
  \sqrt{\mathrm{Drift}(h;\theta_t)},
  \label{eq:ranking-certificate-headroom-v2}
\end{equation}
then
\begin{equation}
  H_t^{\mathrm{cur}}(g)>H_t^{\mathrm{cur}}(h).
  \label{eq:runtime-ranking-preserved-v2}
\end{equation}
Thus a sufficiently large margin in reference Headroom guarantees preservation of the current-policy Headroom ordering.

\paragraph{Corollary 2.2 (Drift-gated Headroom distortion bound).}
If a replayed group $g$ passes the Policy Drift gate at step $t$, i.e., if $g\in\mathcal{R}_t$, then
\begin{equation}
  \left|H_t^{\mathrm{cur}}(g)-H^{\mathrm{ref}}(g)\right|
  \le
  \sqrt{\tau}.
  \label{eq:tau-distortion-bound-v2}
\end{equation}
Equivalently,
\begin{equation}
  H_t^{\mathrm{cur}}(g)
  \ge
  H^{\mathrm{ref}}(g)-\sqrt{\tau}.
  \label{eq:tau-lower-bound-v2}
\end{equation}
This gives a conservative lower bound on current-policy Headroom for accepted replay groups.

\paragraph{Remark (reference Headroom versus cached and runtime Headroom).}
Proposition~2 and Corollaries~2.1--2.2 are stated against the immutable reference quantity $H^{\mathrm{ref}}(g)$ and the conceptual current-policy quantity $H_t^{\mathrm{cur}}(g)$ because Policy Drift is defined relative to the generation-time policy. The buffer ordering actually used at step $t$ is determined by the cached priority $\widehat H_t(g)$, which is initialized at $H^{\mathrm{ref}}(g)$ and refreshed opportunistically only when a group is scanned. For scanned groups, the runtime value equals the conceptual current-policy value by \eqref{eq:runtime-equals-current}. Since current-step selection does \emph{not} recompute current-policy Headroom for the entire buffer, Proposition~2 should be read as a principled control of reference-to-current mismatch, not as a direct theorem about a uniformly refreshed buffer cache.

\subsection{Replay multiplicity and lifetime exposure}
\label{app:replay-multiplicity-v2}

A buffered group can contribute to training more than once: it is used once as a fresh on-policy group at generation time, and may later re-enter training multiple times via replay. To make this explicit, define the replay indicator
\begin{equation}
  R_t(g)
  :=
  \mathbf{1}[g\in\mathcal{R}_t],
  \label{eq:replay-indicator-v2}
\end{equation}
which equals $1$ if group $g$ is replayed at step $t$ and $0$ otherwise.

The cumulative replay count of group $g$ up to step $T$ is
\begin{equation}
  N_T^{\mathrm{rep}}(g)
  :=
  \sum_{t=t_0(g)+1}^{T} R_t(g),
  \label{eq:lifetime-replay-count-v2}
\end{equation}
where the sum starts at $t_0(g)+1$ because same-step replay is disallowed. The corresponding total training-use count is
\begin{equation}
  N_T^{\mathrm{use}}(g)
  :=
  1 + N_T^{\mathrm{rep}}(g),
  \label{eq:total-use-count-v2}
\end{equation}
where the leading $1$ accounts for the initial fresh on-policy use at generation time.

To connect replay reuse to the mixed-batch objective, define the cumulative mixed-objective exposure weight
\begin{equation}
  W_T(g)
  :=
  \sum_{t\le T}
  \frac{\mathbf{1}[g\in\mathcal{A}_t]}{|\mathcal{A}_t|}.
  \label{eq:cumulative-exposure-v2}
\end{equation}
This quantity records how much objective mass group $g$ receives over time under the step-wise mixed-batch objective.

These definitions make explicit that replay selection induces a structured reuse process over buffered groups. A group's lifetime replay count and cumulative exposure are jointly shaped by (i) the ingress rule that determines whether it enters the buffer, (ii) finite FIFO residence, (iii) Headroom-ordered prefix truncation during scanning, and (iv) Policy Drift admissibility under the current policy. In later appendices, these quantities are summarized empirically through replay age, replay count, and scan-prefix statistics.

\subsection{Replay-step formalization}
\label{app:replay-step-formalization-v2}

Let $\mathcal{I}_t\subseteq\mathcal{G}_t^{\mathrm{on}}$ denote the ingress candidate set produced by the task-appropriate ingress rule at step $t$. Replay selection operates only on the pre-existing buffer $\mathbf{B}_t$; current-step ingress is appended only after replay selection and the GRPO update.

Let
\begin{equation}
  \mathbf{C}_t
  =
  \mathrm{Sort}_{\downarrow \widehat H_t}(\mathbf{B}_t)
  =
  \bigl(g_t^{(1)},g_t^{(2)},\dots,g_t^{(|\mathbf{B}_t|)}\bigr)
  \label{eq:ordered-buffer-by-cached-headroom-v2}
\end{equation}
be the buffer ordered by descending cached Headroom at the beginning of step $t$. Define the Drift-feasible buffered subset
\begin{equation}
  \mathcal{F}_t(\tau)
  :=
  \{g\in\mathbf{B}_t \mid \mathrm{Drift}(g;\theta_t)\le \tau\}.
  \label{eq:drift-feasible-subset-v2}
\end{equation}
The accepted replay set is the top-$K_{\mathrm{rep}}$ subset of $\mathcal{F}_t(\tau)$ under the ordering induced by $\widehat H_t$:
\begin{equation}
  \mathcal{R}_t
  :=
  \operatorname{TopK}_{K_{\mathrm{rep}}}^{\downarrow \widehat H_t}\bigl(\mathcal{F}_t(\tau)\bigr).
  \label{eq:accepted-replay-topk-v2}
\end{equation}
By construction,
\begin{equation}
  |\mathcal{R}_t|\le K_{\mathrm{rep}},
  \qquad
  \forall g\in\mathcal{R}_t,\; \mathrm{Drift}(g;\theta_t)\le \tau.
  \label{eq:accepted-replay-basic-properties-v2}
\end{equation}

\paragraph{Proposition 3 (Equivalent constrained characterization of replay selection).}
At each training step $t$, the accepted replay set is equivalently characterized as a solution of
\begin{equation}
  \mathcal{R}_t
  \in
  \arg\max_{\mathcal{S}\subseteq\mathcal{F}_t(\tau),\; |\mathcal{S}|\le K_{\mathrm{rep}}}
  \sum_{g\in\mathcal{S}} \widehat H_t(g).
  \label{eq:constrained-headroom-maximization-v2}
\end{equation}
That is, replay selection chooses the cardinality-constrained subset with maximum total cached Headroom among the Drift-feasible candidates.

\emph{Proof.}
Ordering $\mathcal{F}_t(\tau)$ by descending $\widehat H_t$ and taking the first at most $K_{\mathrm{rep}}$ elements is exactly the solution of \eqref{eq:constrained-headroom-maximization-v2}.

For the implementation-aware view, define the cumulative admissible count over the ordered buffer prefix as
\begin{equation}
  c_t(m)
  :=
  \sum_{u=1}^{m}
  \mathbf{1}\bigl[\mathrm{Drift}(g_t^{(u)};\theta_t)\le \tau\bigr].
  \label{eq:cumulative-admissible-count-v2}
\end{equation}
Let the stopping index be
\begin{equation}
  m_t^{\star}
  :=
  \begin{cases}
    \min\{m : c_t(m)=K_{\mathrm{rep}}\}, & \text{if such an } m \text{ exists},\\
    |\mathbf{B}_t|, & \text{otherwise.}
  \end{cases}
  \label{eq:stopping-index-v2}
\end{equation}
Then the actual scan prefix is
\begin{equation}
  \mathcal{S}_t
  :=
  \bigl\{g_t^{(1)},\dots,g_t^{(m_t^{\star})}\bigr\},
  \label{eq:scan-prefix-v2}
\end{equation}
and the accepted replay set can equivalently be written as
\begin{equation}
  \mathcal{R}_t
  =
  \bigl\{g\in\mathcal{S}_t : \mathrm{Drift}(g;\theta_t)\le \tau\bigr\}.
  \label{eq:accepted-replay-from-scan-prefix-v2}
\end{equation}
This makes explicit the Headroom-ordered prefix truncation induced jointly by cached priority ordering and the replay budget $K_{\mathrm{rep}}$.

For every scanned group $g\in\mathcal{S}_t$, the same current-policy reevaluation pass used to compute $\mathrm{Drift}(g;\theta_t)$ also yields the runtime Headroom $H_t^{\mathrm{run}}(g)=H_t^{\mathrm{cur}}(g)$. The cached priority is then updated after selection according to
\begin{equation}
  \widehat H_t^{+}(g)
  :=
  \begin{cases}
    H_t^{\mathrm{run}}(g), & g\in\mathcal{S}_t,\\
    \widehat H_t(g), & g\in\mathbf{B}_t\setminus\mathcal{S}_t.
  \end{cases}
  \label{eq:cached-headroom-refresh-v2}
\end{equation}
Thus runtime Headroom affects future candidate ordering through cache maintenance, but it does not retroactively alter the current-step selection order.

The mixed actor batch used by the GRPO optimizer is then
\begin{equation}
  \mathcal{A}_t
  =
  \mathcal{G}_t^{\mathrm{on}} \uplus \mathcal{R}_t,
  \label{eq:mixed-actor-batch-repeat-v2}
\end{equation}
and the step objective is the mixed-batch objective in \eqref{eq:mixed-objective-v2}.

Finally, the updated cache state is appended with the current-step ingress groups and maintained under FIFO capacity $C$:
\begin{equation}
  \mathbf{B}_{t+1}
  =
  \mathrm{FIFOAppend}_{C}\bigl(\mathbf{B}_t^{+},\mathrm{Ord}(\mathcal{I}_t)\bigr),
  \label{eq:buffer-transition-new-v2}
\end{equation}
where $\mathbf{B}_t^{+}$ denotes the ordered buffer with refreshed cached priorities from \eqref{eq:cached-headroom-refresh-v2}, and each newly appended ingress group $g\in\mathcal{I}_t$ is initialized with cached Headroom
\begin{equation}
  \widehat H_{t+1}(g)=H^{\mathrm{ref}}(g).
  \label{eq:new-ingress-initialization-v2}
\end{equation}
Because appending happens only after selection, groups generated at step $t$ cannot be replayed in the same step.

\section{Detailed benchmark tables}
\label{app:detailed-benchmark-tables}

\subsection{Metric definitions}

Let each input instance produce \(n\) sampled responses with scalar scores in \([0,1]\).

\begin{itemize}[leftmargin=1.5em]
  \item \textbf{Best@\(n\)}: for each input, take the maximum score among its \(n\) samples, then average over inputs.
  \item \textbf{Mean@\(n\)}: for each input, average the \(n\) sample scores, then average over inputs.
  \item \textbf{Macro Avg Mean@\(n\)}: unweighted average of benchmark-level Mean@\(n\) values across benchmarks.
  \item \textbf{Weighted Mean@\(n\)}: weighted average of benchmark-level Mean@\(n\) values, with benchmark weights proportional to benchmark size.
\end{itemize}

Unless noted otherwise, the paper's \textit{Avg Mean@\(n\)} notation is identical to \textit{Macro Avg Mean@\(n\)}.

\begin{table}[t]
  \centering
  \small
  \setlength{\tabcolsep}{3pt}
  \renewcommand{\arraystretch}{1.24}
  \begin{tabular}{lcccc}
    \toprule
    Metric &
    \shortstack{GRPO on-pol.\\matched\\(b256)} &
    \shortstack{GRPO on-pol.\\larger\\(b384)} &
    DAPO &
    ExGRPO \\
    \midrule
    AIME24$\uparrow$ & \second{0.3083} / 0.0888 & 0.2833 / 0.0833 & 0.2750 / \best{0.0966} & \best{0.3417} / 0.0758 \\
    AMC23$\uparrow$ & 0.8000 / 0.4531 & 0.8500 / 0.4633 & 0.8500 / 0.5023 & \second{0.8750} / 0.4469 \\
    MATH500$\uparrow$ & 0.8760 / 0.6927 & 0.8800 / 0.6895 & \best{0.9100} / 0.6944 & \second{0.9060} / 0.6701 \\
    Minerva$\uparrow$ & 0.3529 / \second{0.1994} & 0.3566 / 0.1957 & 0.3824 / 0.1790 & \second{0.3934} / 0.1545 \\
    OlympiadBench$\uparrow$ & 0.4095 / 0.2437 & 0.4095 / 0.2403 & \second{0.4347} / 0.2319 & 0.4318 / 0.2220 \\
    Macro Avg$\uparrow$ & 0.5494 / 0.3356 & 0.5559 / 0.3344 & 0.5704 / 0.3409 & \best{0.5896} / 0.3139 \\
    Weighted Avg$\uparrow$ & 0.5473 / 0.3696 & 0.5486 / 0.3664 & 0.5722 / 0.3636 & \second{0.5772} / 0.3448 \\
    \shortstack{Fresh\\Responses$\downarrow$} & \second{1,843,200} & 2,764,800 & 2,445,312 & \best{1,785,600} \\
    \bottomrule
  \end{tabular}

  \vspace{0.6em}

  \begin{tabular}{lcccc}
    \toprule
    Metric &
    BAPO &
    \shortstack{GRPO+r\\matched\\(b256+r128)} &
    \shortstack{GRPO+r\\larger\\(b256+r256)} &
    Headroom-Drift \\
    \midrule
    AIME24$\uparrow$ & \second{0.3083} / 0.0891 & 0.1417 / 0.0813 & 0.1333 / 0.0792 & 0.2750 / \second{0.0943} \\
    AMC23$\uparrow$ & \best{0.9000} / \second{0.5117} & 0.7000 / \best{0.5312} & 0.6250 / 0.4813 & \second{0.8750} / \second{0.5117} \\
    MATH500$\uparrow$ & \second{0.9060} / 0.7026 & 0.8200 / \best{0.7165} & 0.8160 / \second{0.7160} & 0.8820 / 0.7104 \\
    Minerva$\uparrow$ & \best{0.3971} / 0.1911 & 0.2721 / 0.1976 & 0.2721 / 0.1976 & 0.3456 / \best{0.2022} \\
    OlympiadBench$\uparrow$ & \best{0.4362} / 0.2400 & 0.3323 / 0.2459 & 0.3234 / \second{0.2474} & 0.4050 / \best{0.2477} \\
    Macro Avg$\uparrow$ & \second{0.5895} / \second{0.3469} & 0.4166 / 0.3117 & 0.3922 / 0.3053 & 0.5565 / \best{0.3533} \\
    Weighted Avg$\uparrow$ & \best{0.5778} / \second{0.3712} & 0.4525 / 0.3595 & 0.4421 / 0.3595 & 0.5455 / \best{0.3792} \\
    \shortstack{Fresh\\Responses$\downarrow$} & 2,027,520 & \second{1,843,200} & \second{1,843,200} & \second{1,843,200} \\
    \bottomrule
  \end{tabular}
  \caption{Held-out benchmark results on mathematical reasoning. Each cell reports Best@32 / Mean@32. Arrows indicate metric direction ($\uparrow$ higher is better, $\downarrow$ lower is better). Top-1 values are bold, and Top-2 values are underlined. GRPO + replay matched (b256+r128) and GRPO + replay larger (b256+r256) denote recency-only replay using the latest 128/256 buffered groups without Headroom ranking or Policy Drift gating.}
  \label{tab:math_geo3k_results}
\end{table}

\begin{table}[t]
  \centering
  \small
  \setlength{\tabcolsep}{4pt}
  \renewcommand{\arraystretch}{1.16}
    \begin{tabular}{lccccc}
      \toprule
      Metric & \shortstack{GRPO on-pol.\\matched\\(b128)} & \shortstack{GRPO on-pol.\\larger\\(b192)} & DAPO & Headroom-Drift \\
      \midrule
      Geo3K$\uparrow$ & \second{0.5870} / 0.4927 & 0.5753 / 0.4864 & 0.5827 / \second{0.4952} & \best{0.6245} / \best{0.5111} \\
      MathVista$\uparrow$ & 0.7600 / 0.5695 & \best{0.7740} / 0.5747 & \second{0.7720} / \best{0.5860} & \second{0.7720} / \second{0.5825} \\
      MathVision$\uparrow$ & 0.3727 / 0.1336 & \second{0.3865} / \second{0.1404} & 0.3852 / 0.1357 & \best{0.4086} / \best{0.1475} \\
      Macro Avg$\uparrow$ & 0.5732 / 0.3986 & 0.5786 / 0.4005 & \second{0.5800} / \second{0.4056} & \best{0.6017} / \best{0.4137} \\
      Weighted Avg$\uparrow$ & 0.4839 / 0.2740 & \second{0.4945} / 0.2788 & 0.4941 / \second{0.2793} & \best{0.5148} / \best{0.2883} \\
      \bottomrule
    \end{tabular}
  \caption{Multimodal benchmark results (simplified 3-benchmark view). Each cell reports Best@32 / Mean@32. Top-1 values are bold, and Top-2 values are underlined.}
  \label{tab:geo3k_results}
\end{table}

\begin{table}[t]
  \centering
  \small
  \setlength{\tabcolsep}{3pt}
  \renewcommand{\arraystretch}{1.16}
  \begin{tabular}{lccc}
    \toprule
    Metric &
    \shortstack{GRPO on-pol.\\matched\\(b128)} &
    \shortstack{GRPO on-pol.\\larger\\(b192)} &
    DAPO \\
    \midrule
    NQ$\uparrow$ & 0.4908 / 0.4026 & 0.5020 / 0.4152 & 0.4548 / 0.3461 \\
    TriviaQA$\uparrow$ & 0.6569 / 0.5711 & 0.6420 / 0.5626 & 0.6533 / 0.5488 \\
    PopQA$\uparrow$ & 0.4885 / 0.4055 & 0.4907 / 0.4196 & 0.4786 / 0.3916 \\
    HotpotQA$\uparrow$ & 0.4013 / 0.2895 & 0.4050 / 0.3017 & 0.3906 / 0.2684 \\
    2WikiMultiHopQA$\uparrow$ & 0.3913 / 0.2312 & 0.3702 / 0.2306 & 0.3877 / 0.2200 \\
    Musique$\uparrow$ & 0.1528 / 0.0890 & 0.1546 / 0.0909 & 0.1483 / 0.0838 \\
    Bamboogle$\uparrow$ & 0.3370 / 0.2480 & 0.3477 / 0.2280 & 0.3136 / 0.1985 \\
    Macro Avg$\uparrow$ & 0.4169 / 0.3196 & 0.4160 / 0.3212 & 0.4038 / 0.2939 \\
    Weighted Avg$\uparrow$ & 0.4733 / 0.3674 & 0.4669 / 0.3719 & 0.4646 / 0.3486 \\
    \shortstack{Fresh\\Responses$\downarrow$} & \best{204,800} & \second{307,200} & 309,248 \\
    \bottomrule
  \end{tabular}

  \vspace{0.6em}

  \begin{tabular}{lccc}
    \toprule
    Metric &
    \shortstack{GRPO+r\\matched\\(b128+r64)} &
    \shortstack{GRPO+r\\larger\\(b128+r128)} &
    Headroom-Drift \\
    \midrule
    NQ$\uparrow$ & 0.5319 / \best{0.4496} & \best{0.5507} / \second{0.4413} & \second{0.5449} / 0.4269 \\
    TriviaQA$\uparrow$ & 0.6785 / \best{0.5984} & \second{0.6856} / \second{0.5903} & \best{0.6940} / 0.5889 \\
    PopQA$\uparrow$ & 0.5114 / \second{0.4339} & \second{0.5221} / 0.4239 & \best{0.5301} / \best{0.4375} \\
    HotpotQA$\uparrow$ & 0.4539 / \second{0.3321} & \best{0.4771} / \best{0.3407} & \second{0.4721} / 0.3320 \\
    2WikiMultiHopQA$\uparrow$ & 0.4862 / \second{0.2906} & \second{0.5033} / \second{0.2906} & \best{0.5115} / \best{0.3083} \\
    Musique$\uparrow$ & 0.1986 / 0.1132 & \second{0.2247} / \second{0.1239} & \best{0.2305} / \best{0.1241} \\
    Bamboogle$\uparrow$ & 0.3760 / \second{0.2660} & \second{0.4000} / 0.2560 & \best{0.4320} / \best{0.2860} \\
    Macro Avg$\uparrow$ & 0.4623 / \second{0.3548} & \second{0.4805} / 0.3524 & \best{0.4879} / \best{0.3577} \\
    Weighted Avg$\uparrow$ & 0.5201 / \second{0.4062} & \second{0.5346} / 0.4028 & \best{0.5399} / \best{0.4084} \\
    \shortstack{Fresh\\Responses$\downarrow$} & \best{204,800} & \best{204,800} & \best{204,800} \\
    \bottomrule
  \end{tabular}
  \caption{Agentic Search results. Arrows indicate metric direction ($\uparrow$ higher is better, $\downarrow$ lower is better). Top-1 values are bold, and Top-2 values are underlined.}
  \label{tab:search_results}
\end{table}

\begin{table}[t]
  \centering
  \small
  \setlength{\tabcolsep}{3.5pt}
  \renewcommand{\arraystretch}{1.16}
  \resizebox{\linewidth}{!}{%
    \begin{tabular}{lcccc}
      \toprule
      Benchmark &
      \shortstack{GRPO+r matched\\Mean@4} &
      \shortstack{Headroom-Drift\\Mean@4} &
      \shortstack{GRPO+r matched\\Best@4} &
      \shortstack{Headroom-Drift\\Best@4} \\
      \midrule
      NQ & 0.4057 & \best{0.4432} & 0.5244 & \best{0.5604} \\
      TriviaQA & 0.6107 & \best{0.6228} & 0.7108 & \best{0.7172} \\
      PopQA & 0.4069 & \best{0.4292} & 0.5002 & \best{0.5253} \\
      HotpotQA & 0.3596 & \best{0.3748} & 0.5018 & \best{0.5114} \\
      2WikiMultiHopQA & \best{0.3394} & 0.3378 & \best{0.5789} & 0.5597 \\
      Musique & 0.1436 & \best{0.1587} & 0.2491 & \best{0.2627} \\
      Bamboogle & 0.3500 & \best{0.4020} & \best{0.5280} & 0.5200 \\
      \midrule
      Macro Avg & 0.3737 & \best{0.3955} & 0.5133 & \best{0.5224} \\
      Weighted Avg & 0.4158 & \best{0.4298} & 0.5556 & \best{0.5638} \\
      \bottomrule
    \end{tabular}%
  }
  \caption{Additional 7B Agentic Search scale check on Search-R1 with Qwen2.5-7B-Instruct. Both methods use identical configurations and an 8-bit optimizer, and the controlled comparison is based on one run per method with four samples per input. The better value in each metric pair is bold.}
  \label{tab:search_results_7b}
\end{table}

As shown in Table~\ref{tab:search_timing}, Headroom-Drift achieves a lower per-step time than GRPO on-policy larger (b192) despite including replay-selection overhead, suggesting that in agentic settings where environment interaction dominates rollout cost, replay can reduce wall-clock training time in addition to improving sample quality.

\section{Replay ingress in the multi-reward Geo3K setting}\label{app:geo3k-multireward-ingress}

\subsection{Motivation and problem setup}

In this section, we analyze a multi-reward refinement of the replay-ingress rule in Headroom-Drift Replay, using Geometry3K as a focused case study. As stated in the main method description, the binary-reward setting uses mixed-outcome-only ingress; the present section clarifies how that same ingress principle should be specialized when multiple reward components jointly determine the final reward. The key issue is not dataset-specific: reward heterogeneity can distort the semantics of "success" and therefore distort replay-ingress decisions. To make the scope of this analysis explicit, we frame the central question as follows.

\begin{rqbox}
  \textbf{RQ1.} \textbf{Under a multi-reward setting, which groups should be admitted to the replay buffer in order to preserve a useful training signal for subsequent replay learning?}
\end{rqbox}

The original replay ingress rule was designed to admit a group into the replay buffer only when its success count was neither zero nor full. The motivation is that if all samples within a group receive the same reward, the group-relative advantage in GRPO becomes zero, so the group provides no useful signal for policy updates. Accordingly, our goal was to prioritize replaying groups that still preserve within-group reward differences and therefore retain a useful training signal.

In binary-reward settings, the original sample-level success indicator \(s_i^{+}=\mathbf{1}[r_i>0]\) served as an exact indicator of correctness. Under that setting, excluding groups that were all-failure or all-success under \(s_i^{+}\) functioned as an appropriate replay-ingress criterion. In Geometry3K, however, the final reward combines answer correctness and formatting quality, so \(s_i^{+}\) no longer corresponds one-to-one to answer correctness. The problem is that the same \(s_i^{+}\)-based definition used in single-reward settings was applied unchanged to this multi-reward setting.

\begin{table}[htbp]
  \centering
  \begin{tabular}{ll}
    \hline
    Condition & Final reward \\
    \hline
    wrong answer + format failure & 0.0 \\
    wrong answer + format success & 0.1 \\
    correct answer + format failure & 0.9 \\
    correct answer + format success & 1.0 \\
    \hline
  \end{tabular}
  \caption{Geometry3K reward structure.}
  \label{tab:geo3k_reward_structure}
\end{table}

As shown in Table~\ref{tab:geo3k_reward_structure}, the final reward in Geometry3K jointly reflects answer correctness and formatting quality. Accordingly, \(s_i^{+}=\mathbf{1}[r_i>0]\) no longer corresponds exclusively to correctness: an incorrect response may still receive 0.1 for satisfying the format requirement, whereas a correct response may receive 0.9 even when the format requirement is violated. Therefore, carrying over the \(s_i^{+}\)-based success definition from single-reward settings changes the replay ingress dynamics: the dynamics in a multi-reward setting are no longer the same as those in a single-reward setting.

\subsection{Engineering observation: replay ingress drying}

Applying the original \(s_i^{+}\)-based success definition to Geometry3K leads to a rapid drying of replay ingress. An analysis of the training logs shows that, as training progresses, the number of groups newly entering the replay buffer decreases sharply. Figure~\ref{fig:geo3k-replay-buffer-drying} summarizes this pattern. The same logs also indicate that, in later stages of training, an increasing number of groups are filtered out because all responses in the group are judged either all unsuccessful or all successful under the \(s_i^{+}\)-based definition.

\replayfigure[fig:geo3k-replay-buffer-drying]{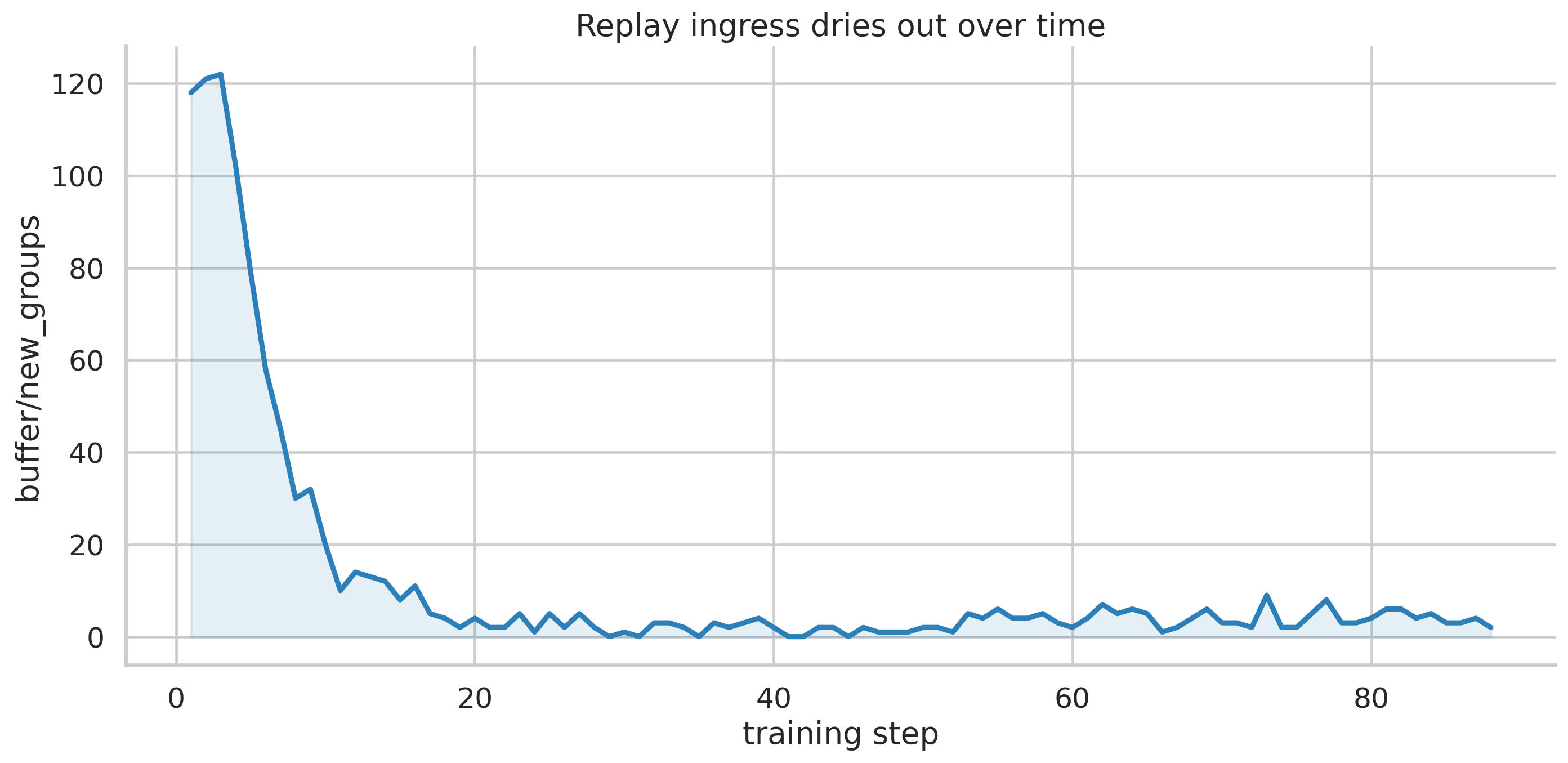}{0.80}{Declining replay-buffer ingress over the course of training in Geometry3K. The batch size is 128.}

This phenomenon did not simply mean that the model had become correct on almost all problems. Figure~\ref{fig:geo3k-reward-mix-over-time} shows that, already after the early stage of training, many groups were classified as all-success groups under the original \(s_i^{+}\)-based definition. Yet these all-success groups were usually not groups in which all responses were actually answer-correct. Rather, many of them contained multiple answer-incorrect samples that nevertheless received positive reward from the format component alone.

\replayfigure[fig:geo3k-reward-mix-over-time]{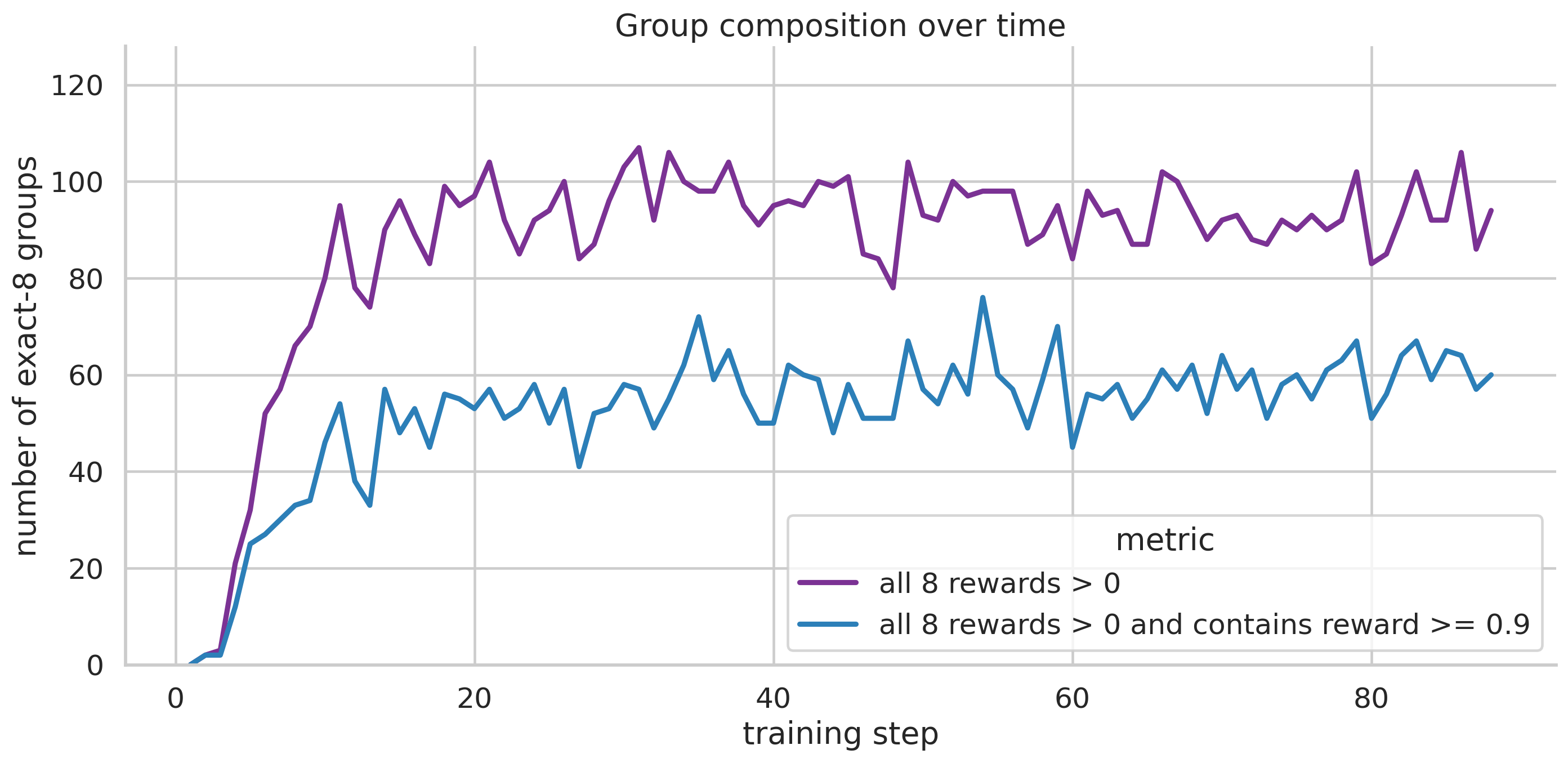}{0.80}{Comparison between the number of exact-8 groups in which all responses satisfy the original success indicator \(s_i^{+}=\mathbf{1}[r_i>0]\) and the number of such groups that contain at least one answer-positive response \((a_i=1\iff r_i\ge 0.9)\). Under the \(s_i^{+}\)-based definition, all such groups are treated uniformly as all-success groups, even though they differ in whether they contain answer-correct responses.}

Thus, in Geometry3K, the label all-success under \(s_i^{+}\) no longer uniquely identifies answer-correct groups; the \(s_i^{+}\)-based definition collapsed the distinction among format-only, answer-only, and fully correct responses by treating all of them uniformly as success. Accordingly, the direct cause of ingress collapse was not that the model had made almost all samples correct, but that the \(s_i^{+}\)-based definition failed to separate answer and format signals and counted format-only positives as success. As a result, even groups that still contained a meaningful mix of correct and incorrect responses in terms of \(a_i\) could be treated as fully positive groups and excluded from replay ingress.

\subsection{Why a correctness-aligned ingress criterion is necessary}

\subsubsection{Format-only reward variance}

In Geometry3K, variation in the final reward is not sufficient by itself to determine whether a group is useful for replay. Because formatting contributes to the final reward, two responses can receive different rewards even when both are answer-incorrect. The example below illustrates this failure mode.

\begin{responseblock}
  \textbf{Example 1. Format-only reward variance without answer-correctness variance}

  \begin{center}
    \begin{tabular}{llll}
      \hline
      Case & Final reward & Answer correctness & Format reward \\
      \hline
      A & \texttt{0.0} & incorrect & fail \\
      B & \texttt{0.1} & incorrect & success \\
      \hline
    \end{tabular}
  \end{center}

  \textbf{Original response A (\texttt{score = 0.0})}
  \begin{rawresponse}
    <think>
    Let's call the missing side length "x". Since the sum of angles in a
    triangle is 180 degrees, we can use the angle sum property to find angle x.
    x + 20° + 135° = 180°
    Simplifying the equation, we get:
    x + 155° = 180°
    Subtracting 155° from both sides, we get:
    x = 25°
    Therefore, angle x is equal to 25°.
    \boxed{25°}
    </think>
  \end{rawresponse}

\end{responseblock}

\begin{responseblock}
  \textbf{Example 1 (continued). Original response B (\texttt{score = 0.1})}

  \begin{rawresponse}
    <think>
    To find the value of x, we can use the sine function in a right triangle.
    The sine function relates the angle of a side (adjacent) to the ratio of
    the opposite side to the hypotenuse. Since angle 20° is adjacent to x,
    we can write:

    sin(20°) = (18) / x

    Now, let's solve for x:

    x = (18) / sin(20°)

    x = (18) / (0.3420)

    x ≈ 52.90
    </think>

    \boxed{52.90}
  \end{rawresponse}
  Case A is format-incorrect because the boxed final answer remains inside the \texttt{<think>} block, whereas Case B is format-correct because the boxed final answer appears outside \texttt{<think>}.
\end{responseblock}

This example highlights the core mismatch. From the perspective of final reward, the group contains non-trivial variation, and such variation would indeed induce a non-zero group-relative advantage in GRPO. However, that advantage is not necessarily aligned with answer correctness. In this case, the resulting update would favor a format-correct but still answer-incorrect response over a response that is both format-incorrect and answer-incorrect. Therefore, a replay criterion based only on final-reward variation can retain groups whose internal preference structure is driven primarily by formatting rather than by answer quality.

The same issue appears at run level. Because GRPO updates the policy from relative reward differences within a group, format-driven variance can be repeatedly reinforced even when answer quality does not improve. Figure~\ref{fig:geo3k-format-reward-saturation-over-time} shows that, even under the baseline on-policy setting, format reward rises quickly whereas answer reward improves more gradually. This suggests that formatting can be learned sufficiently from on-policy data, while replay capacity should be reserved more selectively for groups informative about answer correctness.

\replayfigure[fig:geo3k-format-reward-saturation-over-time]{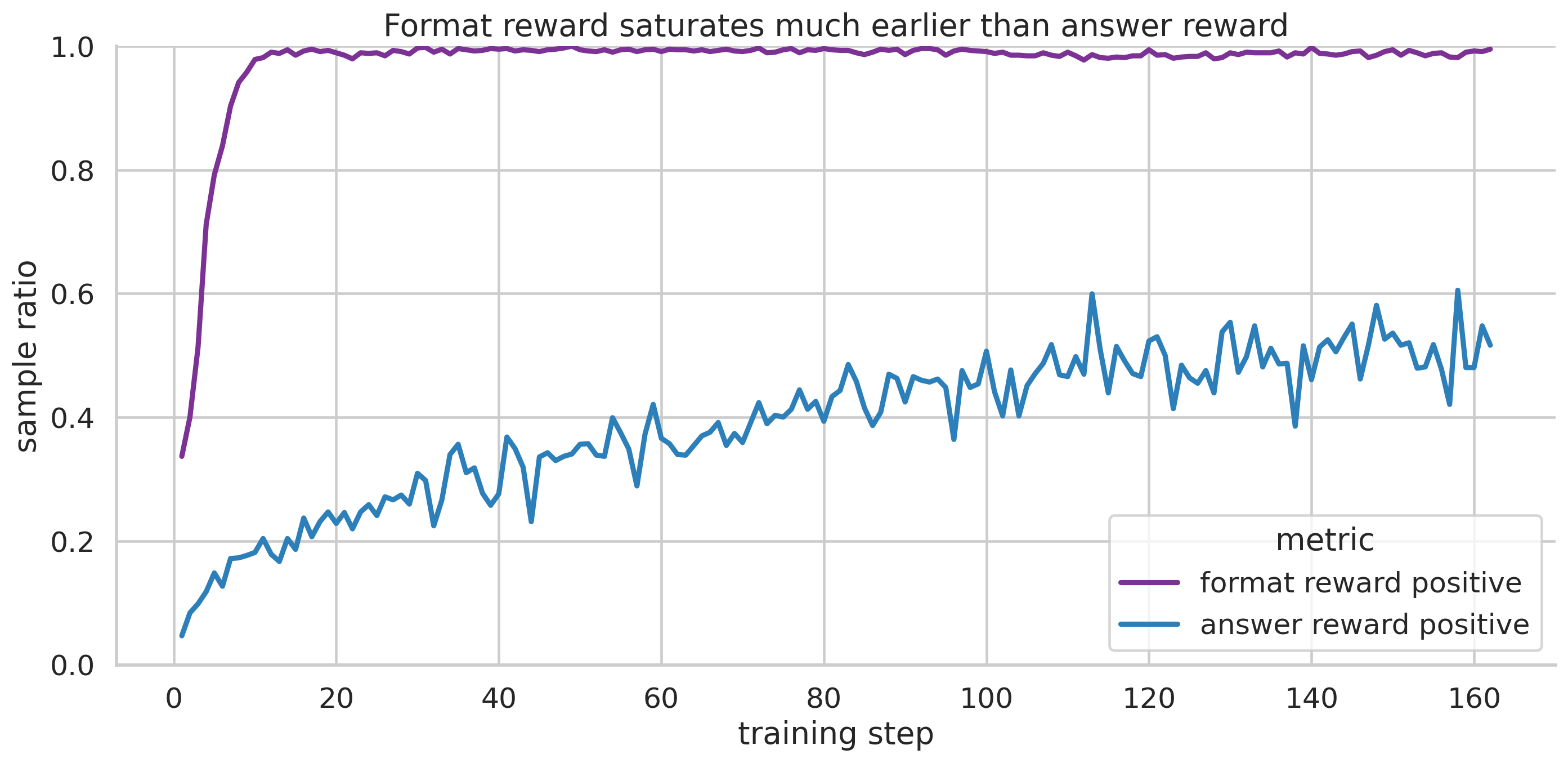}{0.92}{Format reward saturates much earlier than answer reward in Geometry3K.}

Consistent with this interpretation, Figure~\ref{fig:geo3k-ingress-low-reward-groups-over-time} shows that the original ingress rule can continue to admit groups composed only of 0.0/0.1 rewards. These groups contain no answer-correct samples, and their within-group reward variance is driven entirely by formatting. As a result, replay-buffer capacity can be consumed by groups that do not contribute directly to answer-level improvement.

\replayfigure[fig:geo3k-ingress-low-reward-groups-over-time]{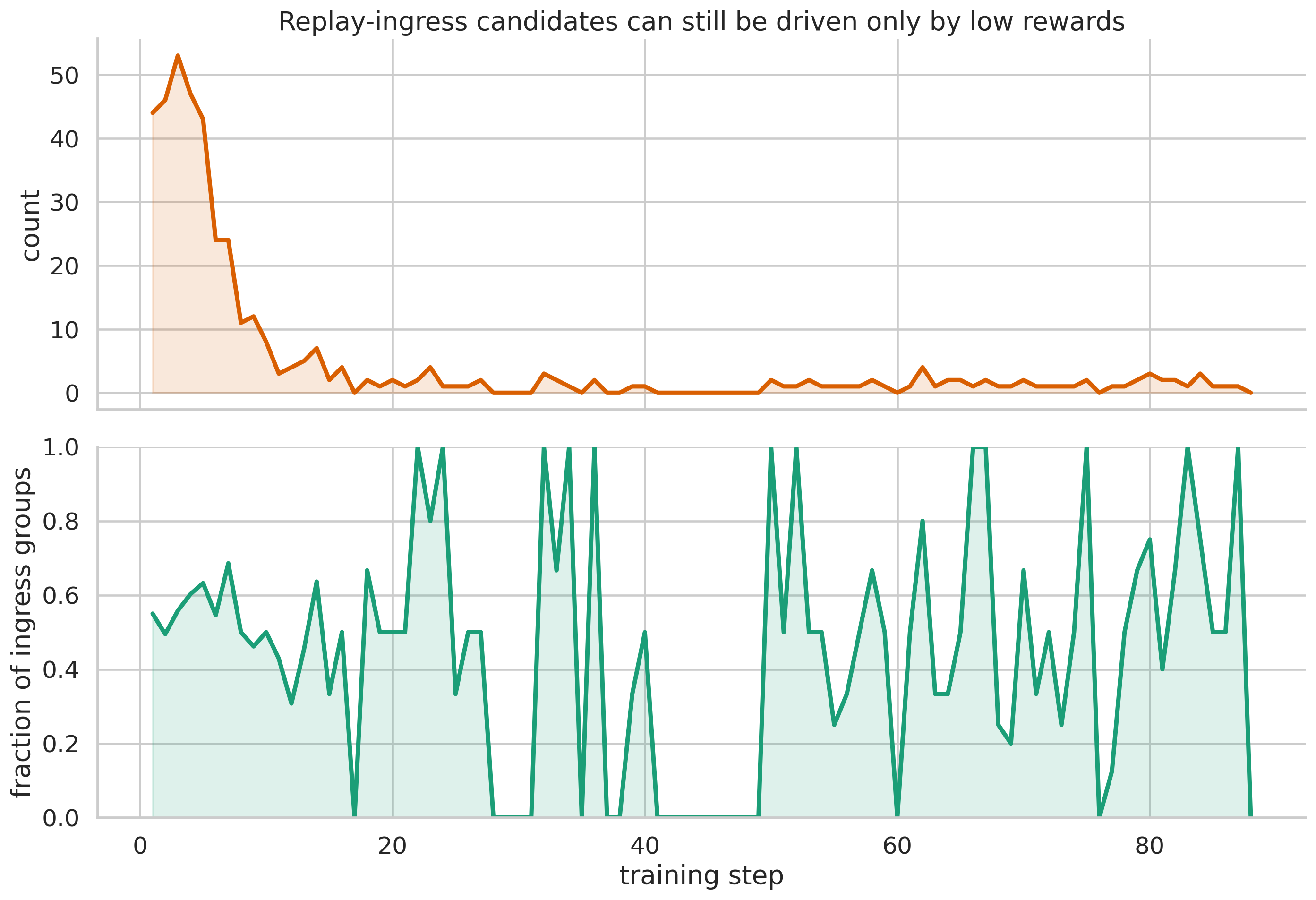}{0.80}{The count and fraction of groups that can enter the replay buffer under the original ingress rule while containing only 0.0 and 0.1 rewards. Such groups contain no answer-correct samples at all, and their within-group reward variation is driven entirely by formatting rather than answer correctness. As the figure shows, these groups are not confined to the very early stage of training, but continue to appear as ingress candidates well into the middle and later stages.}

These observations imply that the main issue is not the existence of a formatting reward, but the ingress criterion itself: the original rule treats answer-driven and format-driven variation as equally replay-worthy. We therefore redesign ingress around the answer-based indicator \(a_i=\mathbf{1}[r_i\ge 0.9]\), so that limited replay capacity is preferentially allocated to groups that contain answer-relevant variation. To make this design choice explicit and reproducible, we now formalize the answer-based ingress rule used in Geometry3K.

\subsection{Formal definition of answer-based ingress}

The discussion above motivates the need for an answer-based ingress rule, but the term \emph{answer-based} must be made precise in order to avoid ambiguity. In particular, our analysis involves both sample-level and group-level notions, as well as both the original \(s_i^{+}\)-based criterion and the revised \(a_i\)-based criterion. We therefore summarize the notation and terminology first, and then define the answer-based ingress rule formally.

\begin{table}[htbp]
  \centering
  \begin{tabular}{L{0.12\linewidth}L{0.18\linewidth}L{0.58\linewidth}}
    \hline
    Level & Symbol / term & Meaning \\
    \hline
    Sample & \(r_i\) & final reward of response \(i\) \\
    Sample & \(s_i^{+}\) & original success indicator, \(s_i^{+}=\mathbf{1}[r_i>0]\) \\
    Sample & \(a_i\) & answer-based indicator, \(a_i=\mathbf{1}[r_i \ge 0.9]\) \\
    Group & \(C^{+}(g)\) & original success count, \(C^{+}(g)=\sum_{i=1}^{G} s_i^{+}\) \\
    Group & \(S(g)\) & answer-based success count, \(S(g)=\sum_{i=1}^{G} a_i\) \\
    Group & answer-mixed group & a group satisfying \(1 \le S(g) \le G-1\) \\
    \hline
  \end{tabular}
  \caption{Notation and terminology used in the Geometry3K replay-ingress analysis.}
  \label{tab:geo3k_ingress_notation}
\end{table}

Using the notation above, we define answer-based ingress at both the sample and group levels. Let a group consist of \(G\) sampled responses, and let \(i \in \{1,\dots,G\}\) index the responses within the group. For each response, we define an answer-based indicator
\[
  a_i \in \{0,1\},
\]
where \(a_i = 1\) if the response is correct under the answer criterion, and \(a_i = 0\) otherwise.

For a group \(g = \{a_i\}_{i=1}^{G}\), we then define the answer-based success count as
\[
  S(g) = \sum_{i=1}^{G} a_i.
\]
Under the answer-based ingress rule, a group is admitted to the replay buffer if and only if
\[
  1 \le S(g) \le G-1.
\]
That is, replay ingress excludes both answer-all-failure groups (\(S(g)=0\)) and answer-all-success groups (\(S(g)=G\)), and retains only answer-mixed groups in which correct and incorrect responses coexist.

In the Geometry3K experiments, the group size is fixed to \(G=8\). The answer-based indicator is instantiated from the task reward structure as
\[
  a_i = \mathbf{1}[r_i \ge 0.9],
\]
where \(r_i\) denotes the final reward of response \(i\). Under the reward structure in Table~\ref{tab:geo3k_reward_structure}, this means that responses with rewards \(0.9\) and \(1.0\) are treated as answer-correct, whereas responses with rewards \(0.0\) and \(0.1\) are treated as answer-incorrect.

This definition makes the role of answer-based ingress precise. It ensures that replay ingress is determined by whether answer correctness actually varies within the group, rather than by whether the final reward varies for any reason. As a result, groups whose internal variation is driven only by formatting do not satisfy the answer-based criterion, whereas groups containing both correct and incorrect responses remain eligible for replay.

\subsection{Empirical effect of answer-based ingress}

\subsubsection{Improved training dynamics}

\replayfigure[fig:geo3k-score-comparison-over-time]{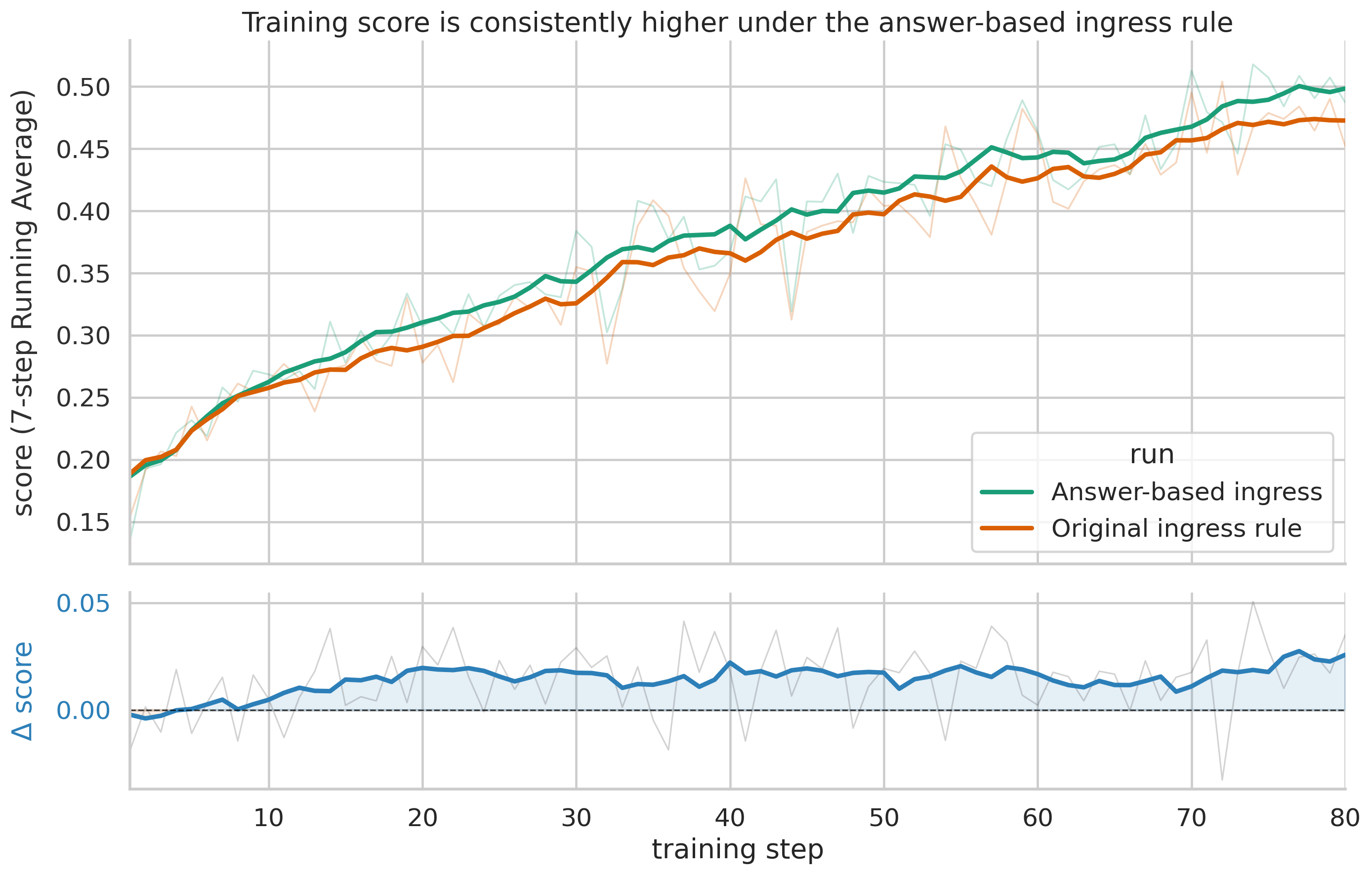}{0.92}{Comparison of training-score trajectories under the original ingress rule and the answer-based ingress rule over the shared early-stage window. The top panel shows the raw trajectories (faint) together with their running averages (bold), while the bottom panel shows the score difference between the two runs over the same window.}

Figure~\ref{fig:geo3k-score-comparison-over-time} shows that over the shared early-stage window, the answer-based ingress rule tends to yield higher training scores than the original ingress rule. The upper panel indicates that the smoothed score trajectory remains above that of the original rule for most of the comparison range, while the lower panel shows a mostly positive gap rather than a single-point spike. These results suggest that aligning replay ingress with answer correctness improves training dynamics in Geometry3K.

\subsubsection{Ingress remains viable}

\replayfigure[fig:geo3k-answer-based-ingress-retention]{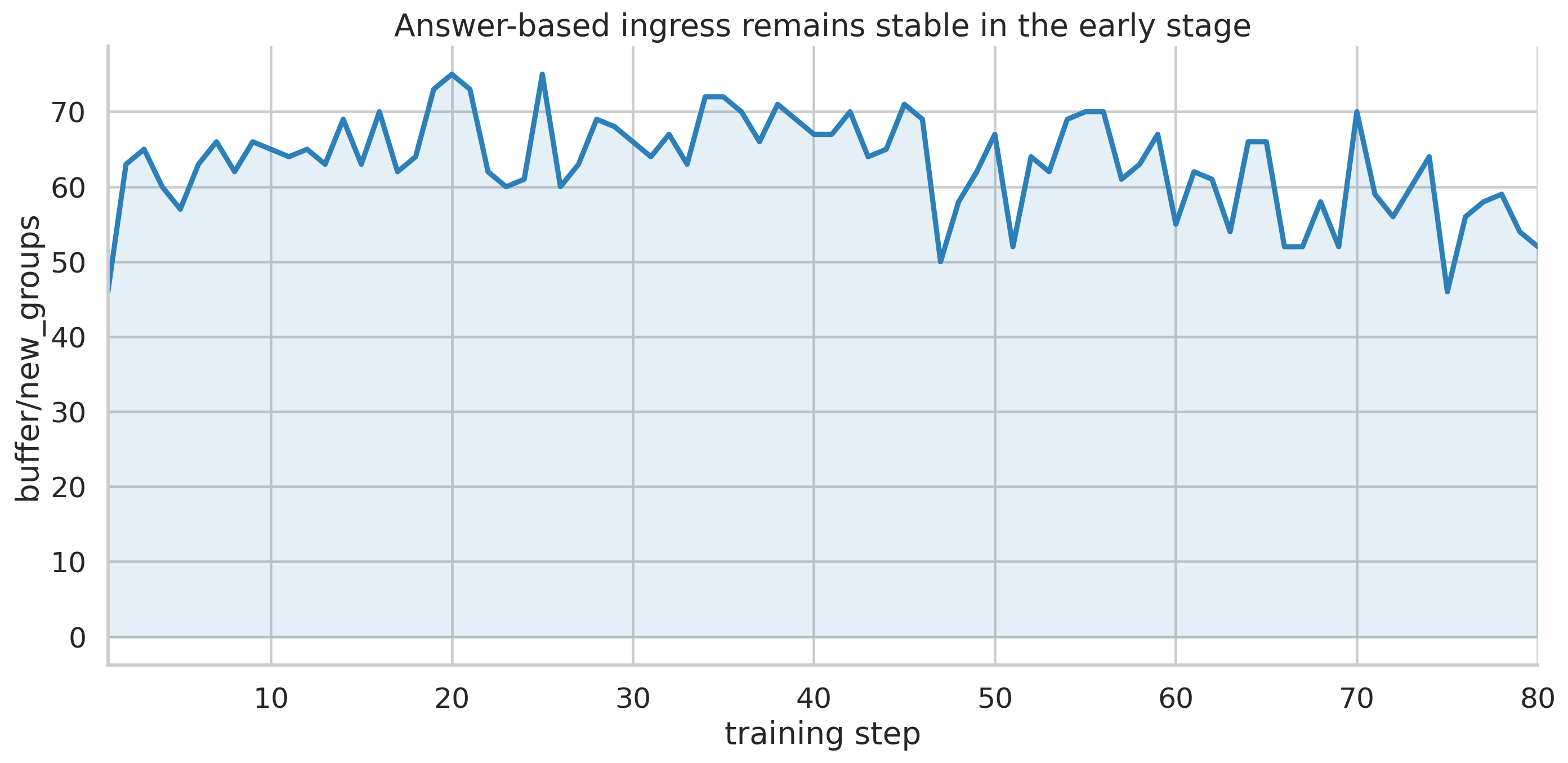}{0.92}{Number of groups newly entering the replay buffer under the answer-based ingress rule during the shared early-stage window. Despite the stricter criterion, replay ingress remains sufficiently active and does not collapse prematurely.}

A natural concern with the answer-based ingress rule is that the stricter \(a_i\)-based definition may reduce ingress so aggressively that replay no longer operates reliably. Figure~\ref{fig:geo3k-answer-based-ingress-retention} shows that this is not what happens. Even in the early stage of training, the number of groups newly entering the replay buffer remains sufficiently large. In other words, strengthening the ingress criterion around answer correctness does not cause replay-buffer supply to dry out prematurely within this window.

These results indicate that answer-based ingress is well motivated in Geometry3K. It is associated with higher training scores while preserving enough replay ingress for replay to remain operational. More broadly, this supports a general design point: when final rewards mix heterogeneous components, replay ingress should be aligned with the target semantic signal (here, answer correctness), not with a coarse positivity test alone.

\begin{samepage}
\begin{rqbox}
  \textbf{Answer to RQ1.} \textbf{In multi-reward settings such as Geometry3K, replay ingress should prioritize groups whose within-group variation reflects the target task signal (answer correctness) rather than auxiliary components (e.g., formatting). This answer-based ingress is associated with better training-score dynamics while maintaining sufficient replay ingress.}
\end{rqbox}
\end{samepage}

\clearpage

\section{Replay age and reuse statistics}
\label{app:replay-age-reuse-stats}

Appendix~\ref{app:formal-headroom-drift} introduced replay age, replay indicators, lifetime replay counts, and cumulative mixed-objective exposure as formal quantities associated with buffer-side replay selection. This section summarizes how those quantities behave empirically in training. Our goal here is descriptive: we make visible how Headroom ranking, Policy Drift gating, and replay-budget truncation shape the actual reuse patterns observed in training. Unless otherwise noted, the statistics below are reported for the mathematical-reasoning run.

\subsection{Replay age and drift compatibility}
\label{app:replay-age-compatibility}

\begin{figure}[t]
  \centering
  \includegraphics[width=\linewidth]{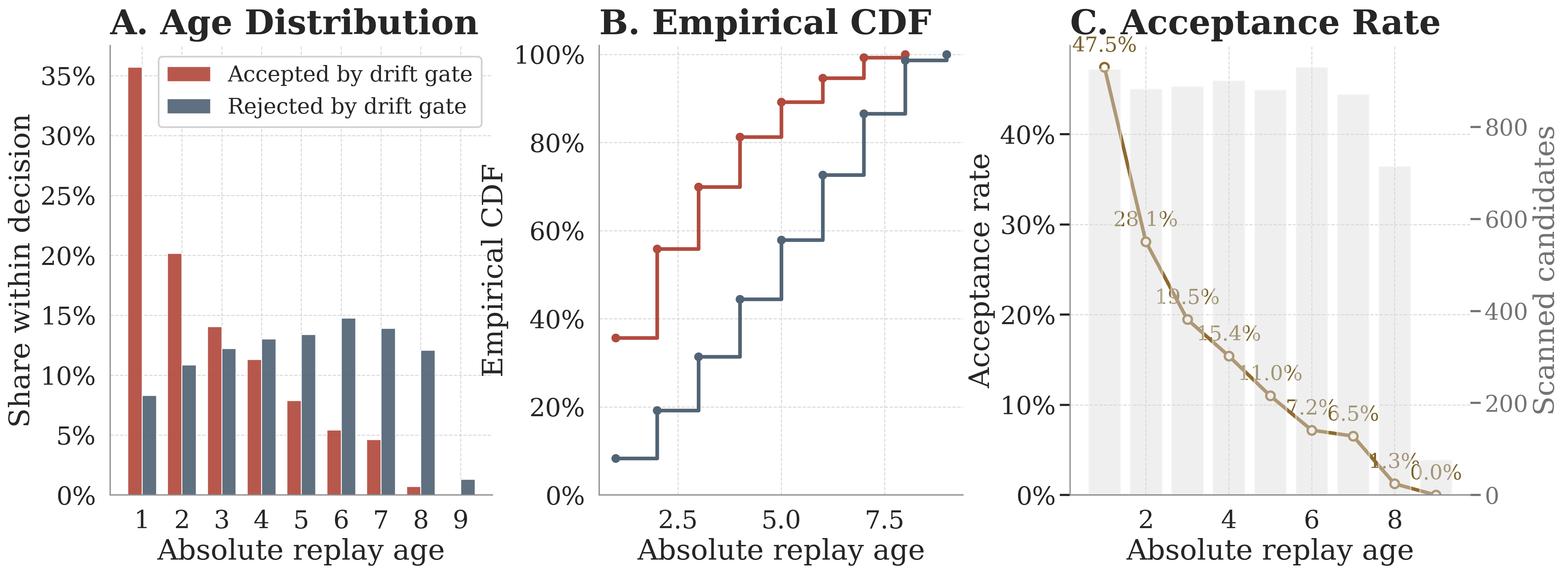}
  \caption{
    Replay age and Drift-gate compatibility in mathematical-reasoning.
    \textbf{(A)} Age histogram normalized separately within the accepted and rejected populations.
    \textbf{(B)} Empirical CDF of replay age for accepted and rejected candidates.
    \textbf{(C)} Age-conditioned acceptance rate, with the gray bars indicating the number of scanned candidates in each age bucket.
    Accepted replay is concentrated on younger groups overall, but the two age distributions overlap substantially, and rejection remains non-negligible even at the smallest replay ages. The raw tail counts at large ages are additionally constrained by finite FIFO turnover, so the age-conditioned acceptance rate in Panel~(C) is the most direct view of compatibility.
  }
  \label{fig:math_replay_absolute_age_checkpoint_aligned_sampled50}
\end{figure}

A natural first question is whether absolute replay age alone is sufficient to characterize replay compatibility. To examine this, we compare the age distributions of replay candidates accepted by the Policy Drift gate and those rejected by it. Figure~\ref{fig:math_replay_absolute_age_checkpoint_aligned_sampled50} summarizes the resulting age statistics in three complementary views.

Figure~\ref{fig:math_replay_absolute_age_checkpoint_aligned_sampled50} shows a clear age gradient in compatibility: older groups are less likely to pass the Policy Drift gate. Accepted groups have a lower mean absolute replay age than rejected-by-drift groups, and the age-conditioned acceptance rate decreases sharply as replay age increases. This is consistent with the intended role of Policy Drift as a current-policy compatibility filter.

At the same time, the figure also shows that age alone is not sufficient to determine compatibility. The accepted and rejected empirical CDFs overlap substantially in the low-age regime, and rejection remains visible even for the youngest replay ages. Thus, while older groups tend to be less reusable, recent groups are not automatically safe. This is precisely why replay compatibility is controlled by Policy Drift rather than by absolute recency alone.

The thin tail at large replay ages is not produced only by the Drift gate; it is also mechanically constrained by finite FIFO residence in the replay buffer. Accordingly, the main message of Figure~\ref{fig:math_replay_absolute_age_checkpoint_aligned_sampled50} is not that old groups disappear solely because of gating, but that \emph{conditional on being scanned}, older groups are less likely to remain compatible, while age alone still does not fully determine acceptance.

\subsection{Headroom-ordered prefix truncation in practice}
\label{app:prefix-truncation-practice}

The replay rule in Appendix~\ref{app:formal-headroom-drift} can be viewed as a Headroom-ordered prefix truncation procedure: buffered groups are ordered by cached Headroom, scanned in that order, and admitted only if they remain Drift-compatible, with selection terminating once the replay budget is filled. Figure~\ref{fig:math_prefix_truncation} shows how this mechanism behaves in practice.

\replayfigure[fig:math_prefix_truncation]{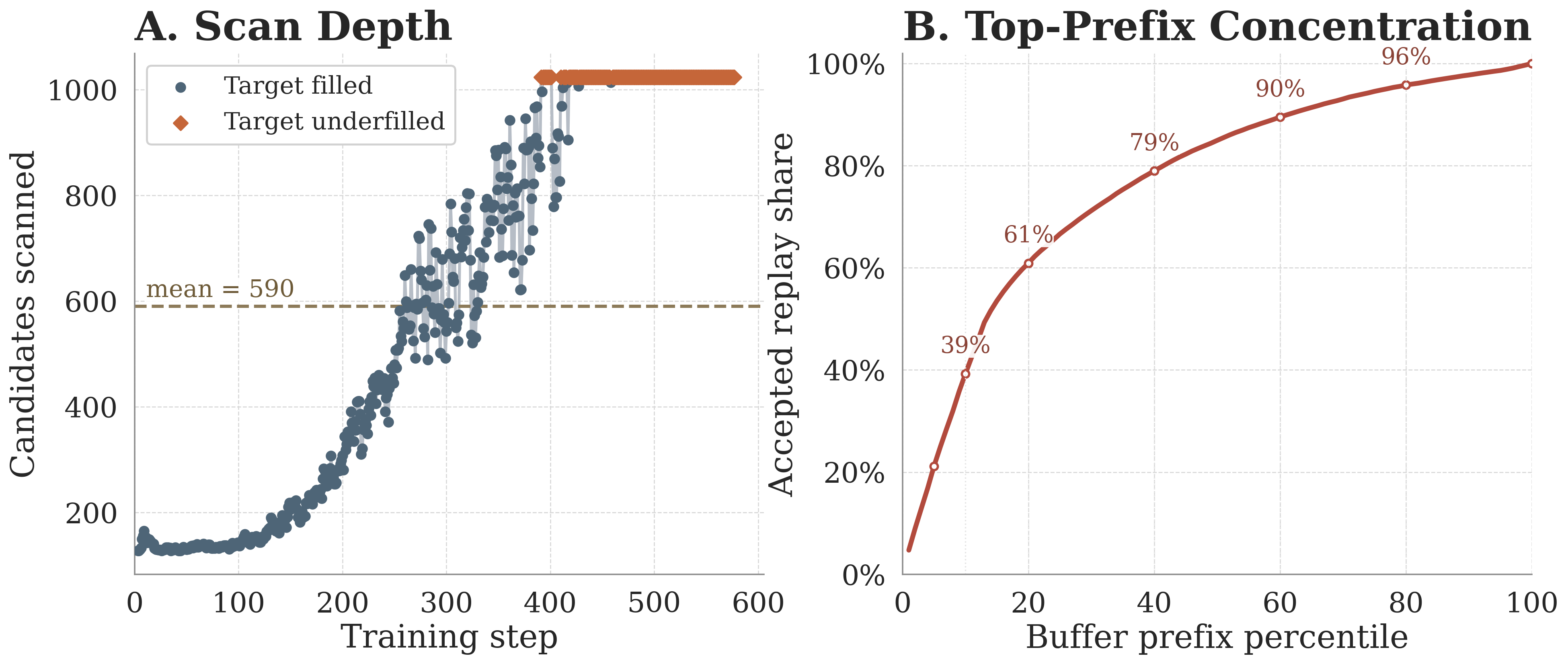}{1.00}{Headroom-ordered prefix truncation in practice. \textbf{(A)} Actual scan depth per training step, measured as the number of buffered candidates examined before replay selection terminates. Blue circles indicate steps that filled the replay target, and orange diamonds indicate underfilled steps. \textbf{(B)} Cumulative share of accepted replay recovered from the top Headroom-ordered buffer prefix. Accepted replay is strongly concentrated near the top prefix of the Headroom-ordered buffer, but later training increasingly requires deeper scans because the Policy Drift gate becomes more selective.}

Figure~\ref{fig:math_prefix_truncation} shows two complementary properties of the replay-selection mechanism. First, accepted replay is strongly concentrated near the top of the Headroom-ordered buffer. In Panel~(B), a relatively small top prefix already accounts for a large share of all accepted replay, indicating that cached Headroom ordering is not merely decorative but materially shapes which groups are reused.

Second, the scan depth needed to satisfy the replay budget grows over training. As shown in Panel~(A), early steps can often fill the replay target after scanning only a shallow prefix, whereas later steps increasingly require deep scans and sometimes fail to fill the target at all. This pattern is consistent with a tightening effective admissible set under the Policy Drift gate: as training progresses, fewer buffered groups remain simultaneously high-priority and sufficiently near-policy.

Taken together, the two panels show that replay selection is not well described as either a fixed top-$K$ rule or a uniform scan over the entire buffer. Rather, it is an adaptive prefix-truncation process in which Headroom ordering determines where replay mass is concentrated, while Policy Drift and the replay budget determine how deep the scan must go to recover enough admissible groups.

\subsection{Initial stored headroom and lifetime replay pressure}
\label{app:initial-headroom-lifetime-replay}

We next ask whether stored Headroom acts as a genuine replay-selection pressure over training, rather than merely serving as a definitional score attached to buffered groups. To test this, we bucket groups by the decile of their \emph{initial stored Headroom} at buffer entry, then measure how often those groups are replayed over their lifetime.

\begin{figure}[t]
  \centering
  \includegraphics[width=\linewidth]{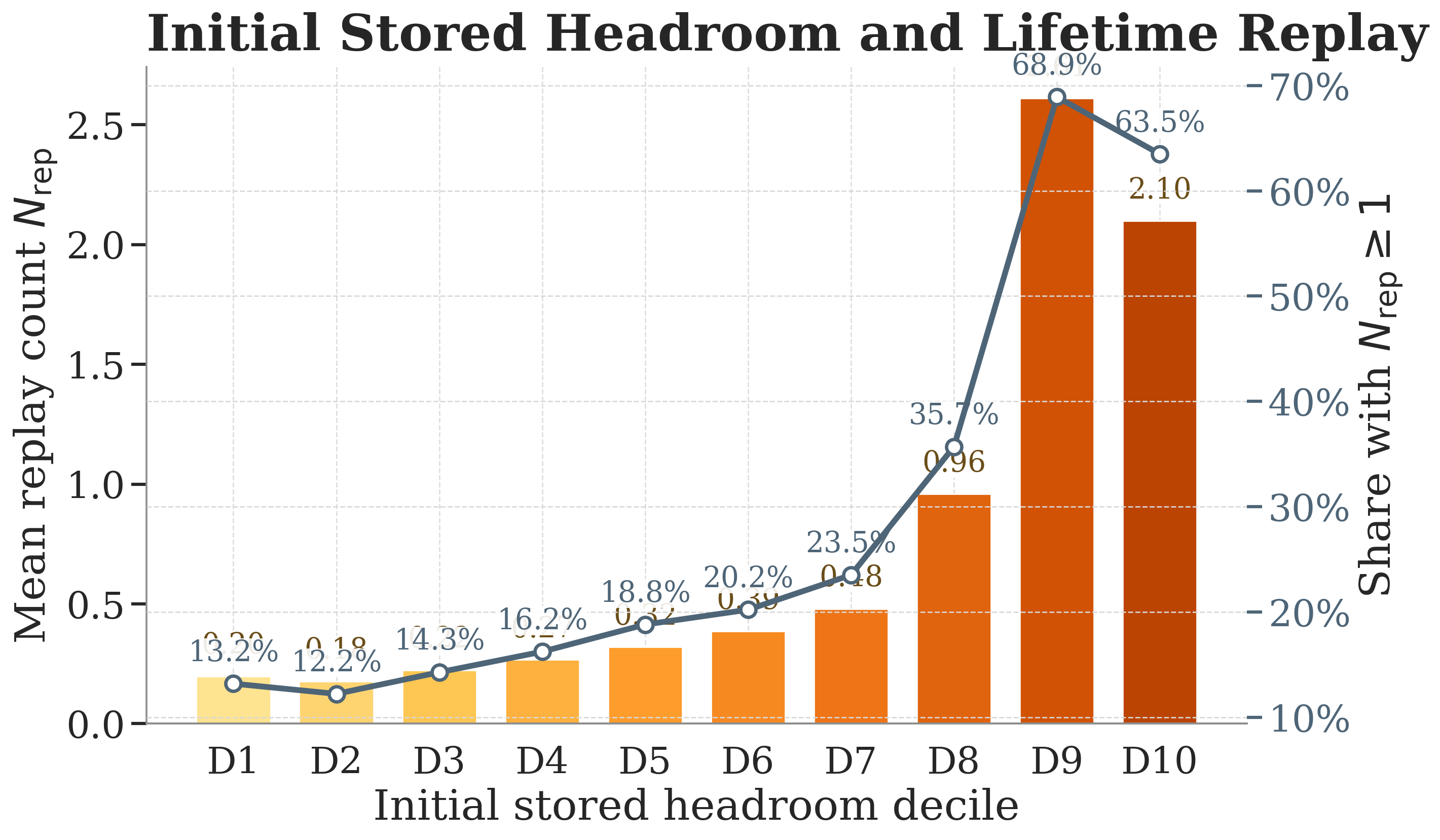}
  \caption{
    Initial stored Headroom and lifetime replay pressure.
    Bars show the mean lifetime replay count \(N_{\mathrm{rep}}\) for groups in each initial stored-Headroom decile, and the line shows the probability that a group is replayed at least once.
    Higher initial stored Headroom is associated with both more frequent reuse and a higher probability of entering replay at least once, indicating a real Headroom-driven selection pressure rather than a purely definitional score.
  }
  \label{fig:math_initial_headroom_deciles}
\end{figure}

Figure~\ref{fig:math_initial_headroom_deciles} shows a clear association between initial stored Headroom and subsequent lifetime replay exposure. Low-Headroom deciles exhibit both small mean replay counts and low replay-at-least-once probabilities, whereas the upper deciles show markedly stronger reuse. Thus, the stored Headroom score is not merely a conceptual ranking variable: it induces a real selection pressure on how often groups re-enter training.

Importantly, the relationship is strong but not perfectly deterministic. The uppermost deciles are clearly favored, but they do not form a strictly monotone sequence, and some lower-decile groups are still replayed. This is expected: lifetime reuse is shaped not only by Headroom ranking, but also by Policy Drift admissibility, scan-prefix competition, replay-budget truncation, and finite buffer residence. Higher initial stored Headroom is therefore associated with stronger lifetime replay exposure, though it does not fully determine the reuse process alone.

\subsection{Replay multiplicity and exposure concentration}
\label{app:replay-multiplicity-exposure}

Finally, we examine the overall distribution of replay reuse across buffered groups. The quantities \(N_T^{\mathrm{rep}}(g)\) and \(W_T(g)\) from Appendix~\ref{app:formal-headroom-drift} allow us to distinguish two different views of replay concentration: how many groups fall into each replay-count cohort, and how much total replay exposure mass those cohorts account for.

\begin{figure}[t]
  \centering
  \includegraphics[width=\linewidth]{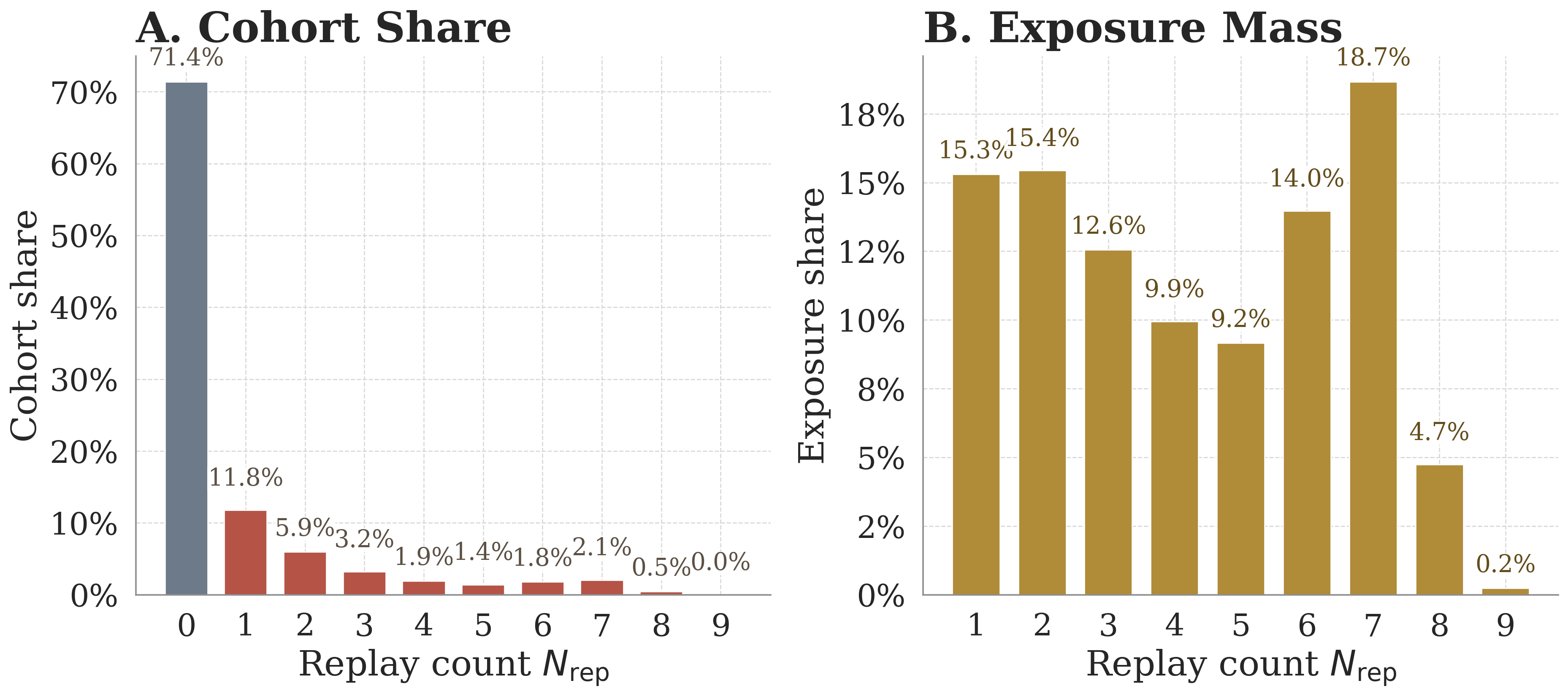}
  \caption{
    Replay multiplicity and exposure concentration.
    \textbf{(A)} Cohort share by lifetime replay count \(N_{\mathrm{rep}}\), showing what fraction of groups are replayed \(0,1,2,\dots\) times.
    \textbf{(B)} Exposure-mass share by lifetime replay count, showing what fraction of total replay-use events comes from each \(N_{\mathrm{rep}}\) cohort.
    Most groups are never replayed, but a much smaller subset of repeatedly reused groups accounts for a disproportionately large share of total replay exposure.
  }
  \label{fig:math_replay_lifetime_counts}
\end{figure}

Figure~\ref{fig:math_replay_lifetime_counts} shows that replay is far from uniformly distributed across buffered groups. Most groups are never replayed at all, and the cohort share decreases rapidly as the lifetime replay count grows. In this sense, Headroom-Drift Replay is highly selective at the cohort level: the majority of buffered groups never re-enter training through replay.

However, the exposure-mass view in Panel~(B) tells a more informative story. Although high-\(N_{\mathrm{rep}}\) cohorts are small in population share, they account for a much larger share of total replay-use events. Thus, replay selection is not merely deciding one-shot inclusion versus exclusion; it is redistributing \emph{lifetime training exposure} toward a comparatively small subset of repeatedly reused groups.

This distinction between cohort share and exposure mass is important for interpreting replay as a training mechanism. A replay rule may appear sparse if viewed only through the fraction of groups that are ever replayed, yet still exert substantial influence on optimization if the repeatedly reused groups receive a large share of total replay exposure. In this sense, Headroom-ranked, Drift-gated replay induces a structured concentration of lifetime exposure rather than a uniform spreading of replay across all buffered groups.

\clearpage
\section{Replay dynamics and the delay of entropy collapse}
\label{app:entropy-collapse}

\subsection{Why entropy collapse matters in GRPO}

In GRPO-style reinforcement learning, late-stage failures such as entropy collapse, policy contraction, reward hacking, and exploration failure are widely reported. Terminology differs across studies, but the core concern is consistent: even when reward or score improves, the policy distribution can become overly concentrated, making subsequent training brittle. From this perspective, entropy collapse is not merely a benign sharpening of the policy; it is a practical indicator of broader training failure \citep{miao2025energy,yuan2026cappo,shen2026bapo,liu2026entropy}.

Prior work addresses this issue from several complementary directions, including optimizer and clipping design \citep{yuan2026cappo,shen2026bapo}, explicit regularization against reward hacking and late-stage degradation \citep{miao2025energy}, and theoretical analyses of entropy dynamics in reinforcement fine-tuning \citep{liu2026entropy}. Taken together, these studies establish entropy collapse as an informative signal of broader training failure. Building on this line of work, this appendix treats replay as an operational factor and asks whether replay changes collapse timing itself, rather than merely shifting the learning curve in absolute training steps. In line with Contribution~3, we keep the analysis observational and focus on how replay shapes training dynamics through gate behavior, mismatch, and entropy/performance evolution.

In our experiments, the replay run appeared to enter entropy collapse later, even after reaching a comparable training-score level. However, this pattern alone is insufficient evidence of more stable learning. A run can also appear to collapse later in absolute-step time simply because it learns more slowly and reaches high-training-score regimes later. We therefore reinterpret replay effects under training-score-matched comparison and ask whether replay merely shifts the trajectory in time or meaningfully delays collapse onset.

\begin{rqbox}
  \textbf{RQ2.} \textbf{Under training-score-matched comparison, does replay meaningfully delay entropy collapse?}
\end{rqbox}

\subsection{Experimental setting and comparison axes}

We compare three training configurations: the standard on-policy baseline (\emph{GRPO on-policy matched (b256)}); \emph{Headroom-Drift Replay (b256+r128)}, which adds Headroom-Drift Replay to the same on-policy setup; and \emph{GRPO on-policy larger (b384)}, which increases the on-policy batch size without replay. All three runs use the same Qwen2.5-Math-1.5B model, GRPO training framework, and dataset. They differ only in replay usage, actor-batch composition, and the number of GRPO mini-batches per update.

\begin{table}[htbp]
  \centering
  \small
  \setlength{\tabcolsep}{4pt}
  \renewcommand{\arraystretch}{1.15}
  \begin{tabular}{lccc}
    \hline
    Setting & \shortstack{GRPO on-policy\\matched (b256)} & \shortstack{Headroom-Drift Replay\\(b256+r128)} & \shortstack{GRPO on-policy\\larger (b384)} \\
    \hline
    Replay & No & Yes & No \\
    On-policy batch size & 256 & 256 & 384 \\
    Replay target groups & 0 & 128 & 0 \\
    GRPO mini-batch size & 64 & 64 & 64 \\
    \shortstack[l]{Number of GRPO mini-batches\\per update} & 4 & 6 & 6 \\
    \hline
  \end{tabular}
  \caption{Runs compared in the entropy-collapse analysis.}
  \label{tab:entropy_collapse_runs}
\end{table}

A GRPO mini-batch here denotes one sequential subdivision of the full actor batch within a single update step. Concretely, \emph{Headroom-Drift Replay (b256+r128)} and \emph{GRPO on-policy larger (b384)} both use six GRPO mini-batches per update, whereas \emph{GRPO on-policy matched (b256)} uses four. We therefore treat \emph{GRPO on-policy larger (b384)} as the stricter on-policy control, because it better matches \emph{Headroom-Drift Replay (b256+r128)} in per-update optimization load. This comparison helps separate the structural effect of replay from the effect of simply processing more mini-batches.

\subsection{Empirical characterization of the observed delay}

Although the replay run appears to enter entropy collapse later in absolute training progress, this observation alone does not establish collapse-free, stable learning. A run may look delayed simply because it progresses more slowly: if it reaches comparable training-score levels later, its transition into low-entropy regions will also appear later on the raw progress axis, even when the underlying collapse dynamics are unchanged. We therefore move beyond raw-progress comparison and use training-score-matched analysis to distinguish two possibilities: replay may merely shift the trajectory in time, or it may delay collapse onset at matched performance levels. We first state the limitation of raw-progress comparison and then summarize the training-score-matched patterns across the three runs.

\subsubsection{From raw-progress comparison to matched-score analysis}

To remove learning-speed confounds, we compare runs at matched training-score levels rather than at absolute progress indices. Let \(S\) denote the training score and \(H\) denote policy entropy. For a training-score anchor \(s_0\), define the first matching index as
\[
  t_{\mathrm{first}}(S \ge s_0).
\]
We then define entry into and exit from the low-entropy regime as
\[
  t_{\mathrm{entry}}(H \le \theta),
  \qquad
  t_{\mathrm{rebound}}(H \ge \theta').
\]
To reduce sensitivity to short-lived fluctuations, \(t_{\mathrm{entry}}\) is detected using 10 consecutive log points, and \(t_{\mathrm{rebound}}\) using 5 consecutive log points.

Using these events, we define two training-score-conditioned delays:
\[
  \Delta_{\mathrm{entry}}(\theta \mid s_0)
  =
  t_{\mathrm{entry}}(H \le \theta)
  -
  t_{\mathrm{first}}(S \ge s_0),
\]
\[
  \Delta_{\mathrm{rebound}}(\theta,\theta' \mid s_0)
  =
  t_{\mathrm{rebound}}(H \ge \theta')
  -
  t_{\mathrm{entry}}(H \le \theta).
\]
\(\Delta_{\mathrm{entry}}\) measures how long a run takes to enter the low-entropy regime after reaching the shared training-score anchor, and \(\Delta_{\mathrm{rebound}}\) measures how long it remains in that regime before rebound.

\subsubsection{Observed patterns across the three runs}

\begin{figure}[!htbp]
  \centering
  \includegraphics[width=0.84\linewidth]{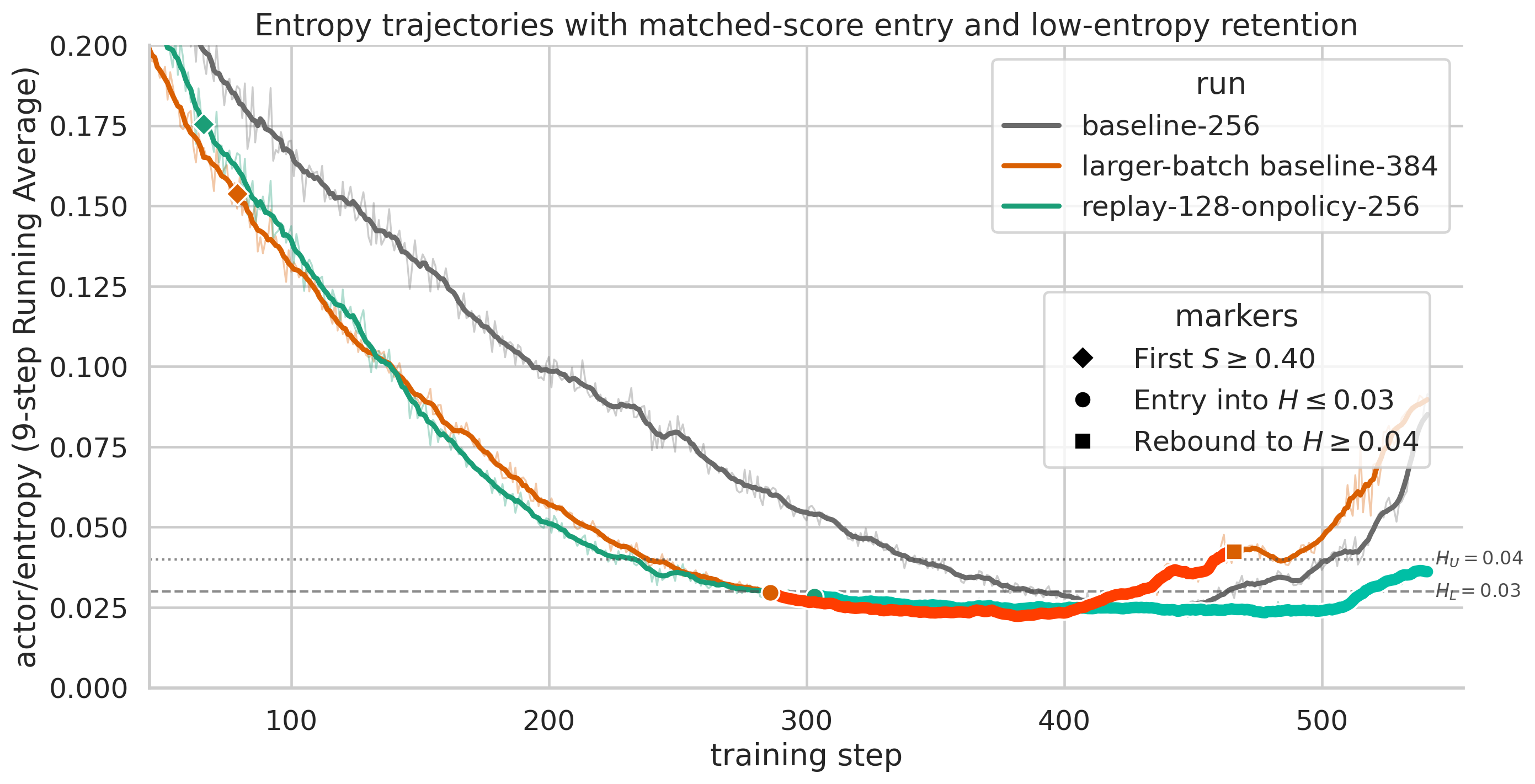}
  \caption{Entropy trajectories for the three runs. The gray \emph{GRPO on-policy matched (b256)} is shown only as a reference. Diamonds denote first arrival at the shared training-score anchor, circles denote entry into the low-entropy regime, squares denote rebound, and the highlighted segments indicate the retained low-entropy window.}
  \label{fig:qwen25_math_entropy_trajectories}
\end{figure}

In this analysis, we use the representative score anchor \(S \ge 0.40\), the low-entropy plateau entry threshold \(H_L=0.03\), and the exit threshold \(H_U=0.04\). Table~\ref{tab:entropy_collapse_summary} summarizes the score-conditioned delays of the three runs under these criteria, and Figure~\ref{fig:qwen25_math_entropy_trajectories} visualizes the corresponding entropy trajectories and marker timings.

\begin{table}[t]
  \centering
  \small
  \setlength{\tabcolsep}{4pt}
  \renewcommand{\arraystretch}{1.15}
  \begin{tabular}{lcc}
    \hline
    Run & \shortstack{Entry delay\\ \(\Delta_{\mathrm{entry}}\)} & \shortstack{Rebound delay\\ \(\Delta_{\mathrm{rebound}}\)} \\
    \hline
    GRPO on-policy matched (b256) & 318 & 121 \\
    \shortstack[l]{Headroom-Drift Replay\\(b256+r128)} & 237 & 261 \\
    GRPO on-policy larger (b384) & 207 & 180 \\
    \hline
  \end{tabular}
  \caption{Entry and rebound delays for the three runs under the representative thresholds defined in the text.}
  \label{tab:entropy_collapse_summary}
\end{table}

Figure~\ref{fig:qwen25_math_entropy_trajectories} shows that \emph{Headroom-Drift Replay (b256+r128)} reaches the shared score anchor \(S \ge 0.40\) earlier than \emph{GRPO on-policy larger (b384)}. The replay run therefore does not lag in entering the high-score regime. Yet the same trajectory indicates later entry into the low-entropy regime, suggesting that score improvement and entropy compression are not strictly synchronized.

The matched-score comparison reinforces this point and extends beyond entry timing. Under \(H_L=0.03\), \emph{GRPO on-policy larger (b384)} shows a shorter post-anchor entry delay than \emph{Headroom-Drift Replay (b256+r128)} (207 vs.\ 237 in Table~\ref{tab:entropy_collapse_summary}). Under \(H_U=0.04\), \emph{GRPO on-policy larger (b384)} also rebounds sooner, while \emph{Headroom-Drift Replay (b256+r128)} sustains the low-entropy regime longer. In this run pair, replay reaches the shared high-score regime earlier, but enters the low-entropy regime later and remains there longer.

Taken together, these observations indicate that the delay cannot be explained only by slower learning in the replay run. At matched score levels, both low-entropy entry and rebound are delayed under replay. The observed pattern is therefore more consistent with a change in post-anchor entropy dynamics than with a simple shift in optimization speed.

\subsection{Interpretation of the observed delay}

The previous subsection suggests that the delayed collapse in the replay run is not simply a by-product of slower optimization; entropy dynamics differ even after runs reach comparable score levels. This subsection develops a working interpretation of that difference and then examines why the effect weakens in the late stage.

\subsubsection{Replay as a near-policy corrective mechanism}

As shown above in Figure~\ref{fig:qwen25_math_entropy_trajectories} and Table~\ref{tab:entropy_collapse_summary}, the replay run reaches the shared score anchor earlier, yet enters and exits the low-entropy regime later. This pattern makes a purely slower-learning explanation unlikely and motivates a working interpretation.

Figure~\ref{fig:replay_selection_in_practice}(c) provides the corresponding conceptual pattern: replay under Headroom-Drift remains concentrated on recent sources, but is not reducible to trivial previous-step reuse.

\begin{table}[htbp]
  \centering
  \footnotesize
  \setlength{\tabcolsep}{4pt}
  \renewcommand{\arraystretch}{1.1}

  \begin{tabular}{lrrrrrr}
    \toprule
    \multicolumn{7}{c}{(A) Source staleness summary} \\
    \midrule
    Population & Mean & Median & P90 & Max & \% age $\ge 2$ & \% age $\ge 3$ \\
    \midrule
    Replay pool & 3.32 & 3 & 6 & 8 & 73.51 & 60.22 \\
    Used replay subset & 2.67 & 2 & 5 & 7 & 66.00 & 44.06 \\
    \bottomrule
  \end{tabular}

  \vspace{0.35em}

  \begin{tabular}{p{0.72\linewidth}r}
    \toprule
    \multicolumn{2}{c}{(B) Mixing structure of the used replay subset} \\
    \midrule
    Statistic & Value \\
    \midrule
    Mean distinct source ages per update & 4.07 \\
    Fraction of updates with at least 3 source ages & 85.71\% \\
    Fraction of updates with both \(age=1\) and \(age \ge 3\) & 76.79\% \\
    \bottomrule
  \end{tabular}

  \caption{True source staleness of replay groups. Panel (A) shows that the replay groups actually used for updates are more concentrated on recent source samples than the replay pool as a whole, while still spanning a non-trivial short horizon. Panel (B) shows that the used replay subset typically mixes multiple source ages within the same update.}
  \label{tab:qwen25_math_replay_staleness}
\end{table}

Table~\ref{tab:qwen25_math_replay_staleness}(A)\footnote{Source staleness is reconstructed from checkpoint snapshots because replay-group insertion times are directly available only in checkpointed replay states. Since checkpoints are saved every 10 training steps in this run, the reported statistics summarize checkpoint-aligned observations rather than every individual step.}, together with the conceptual view in Figure~\ref{fig:replay_selection_in_practice}(c), indicates that replay in this run is neither trivial one-step reuse nor indiscriminate long-range reuse. The replay groups actually used for updates span source ages from \(1\) to \(7\) steps; \(66.00\%\) come from sources at least \(2\) steps old, and \(44.06\%\) from sources at least \(3\) steps old. At the same time, the used subset is more recent than the replay pool as a whole (mean source staleness: \(2.67\) vs. \(3.32\) steps), indicating a clear recent-source preference rather than uniform reuse across the full buffer. Importantly, this recency bias does not collapse to a single age: Table~\ref{tab:qwen25_math_replay_staleness}(B) shows a mixed-age structure (mean distinct source ages per update: \(4.07\); updates with at least three ages: \(85.71\%\); updates containing both \(age=1\) and \(age\ge3\): \(76.79\%\)).

These statistics are consistent with a near-policy corrective interpretation. Replay mostly reuses recent samples but still incorporates selected older ones, so each update reflects a short-horizon multi-age mixture rather than only the immediately preceding rollout. Relative to pure on-policy learning, replay therefore introduces trajectories generated at nearby policy states that remain useful under current criteria. Within this process, Headroom prioritizes higher-value groups in the replay pool, while the Policy Drift gate filters out trajectories that have drifted too far from the current policy. Under this interpretation, replay is less a pull toward stale policies than a constrained reintroduction of still-compatible trajectories from nearby policy snapshots, which can moderate one-step acceleration while retaining informative signal.

\subsubsection{Late-stage transition and failure mode}

\replayfigure[fig:qwen25-math-ingress-vs-replay-used]{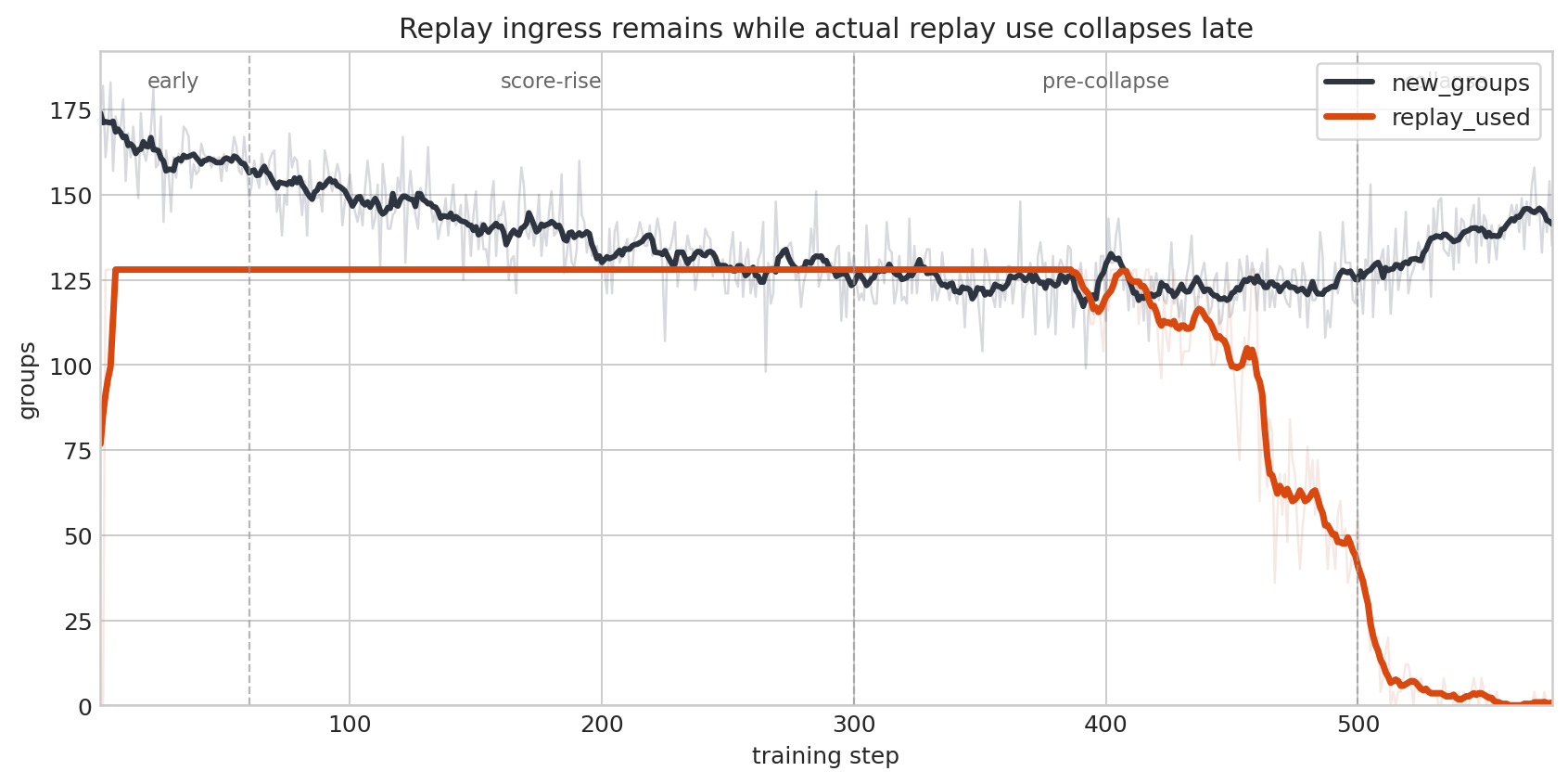}{0.92}{Replay ingress and actual replay use over training. The gray curve shows the number of newly ingressed groups added to the replay buffer at each step, while the orange curve shows the number of replay groups actually used for the update. Although ingress remains substantial into the late stage, actual replay use drops sharply near collapse, suggesting that the late-stage transition is better understood as a contraction of the usable replay subset rather than as a simple ingress collapse.}

\begin{table}[htbp]
  \centering
  \small
  \setlength{\tabcolsep}{4pt}
  \renewcommand{\arraystretch}{1.15}
  \begin{tabular}{lcccc}
    \hline
    Metric & Early learning & Score-rise & Pre-collapse & Collapse \\
    \hline
    Mean source age & 2.77 & 3.26 & 2.24 & 1.26 \\
    \% age $=1$ & 21.15 & 22.90 & 42.13 & 76.47 \\
    \% age $\ge 3$ & 48.08 & 59.35 & 33.50 & 2.94 \\
    Mean distinct ages / update & 4.33 & 4.83 & 3.80 & 1.67 \\
    \% updates with $\ge 3$ ages & 100.00 & 100.00 & 85.00 & 16.67 \\
    \shortstack[l]{\% updates with $age=1$\\and $age\ge3$} & 66.67 & 87.50 & 85.00 & 16.67 \\
    \hline
  \end{tabular}
  \caption{Phase-wise contraction of the used replay mixture. The replay subset is most diverse during the score-rise regime and collapses toward a narrow age-1/2-dominated mixture near the failure phase.}
  \label{tab:qwen25_math_phase_replay_mixture}
\end{table}

Figure~\ref{fig:qwen25-math-ingress-vs-replay-used} shows that the late-stage transition is not well explained by a simple ingress collapse. The number of newly ingressed groups does not drop sharply, even in the late stage. By contrast, the amount of replay actually used in updates falls markedly in the same region, and steps that fail to fill the replay target become frequent near collapse. The transition is therefore better interpreted as a rapid contraction of the subset that remains reusable under current criteria.

This interpretation is reinforced by the phase-wise mixture changes in Table~\ref{tab:qwen25_math_phase_replay_mixture}. During the score-rise phase, the used replay subset has mean source age \(3.26\), share \(age \ge 3\) of \(59.35\%\), mean distinct ages per update of \(4.83\), and \(100\%\) of updates with at least three ages, indicating a rich multi-age mixture during rapid improvement. This structure weakens in pre-collapse and then contracts sharply in collapse: mean source age falls to \(2.24\) and then \(1.26\), the \(age=1\) share rises to \(42.13\%\) and then \(76.47\%\), the \(age \ge 3\) share drops to \(33.50\%\) and then \(2.94\%\), and mean distinct ages decrease to \(3.80\) and then \(1.67\). The late-stage issue is therefore not only reduced replay volume; the corrective mixture itself narrows from a multi-age structure to a highly concentrated age range.

\replayfigure[fig:qwen25-math-policy-drift-proxy]{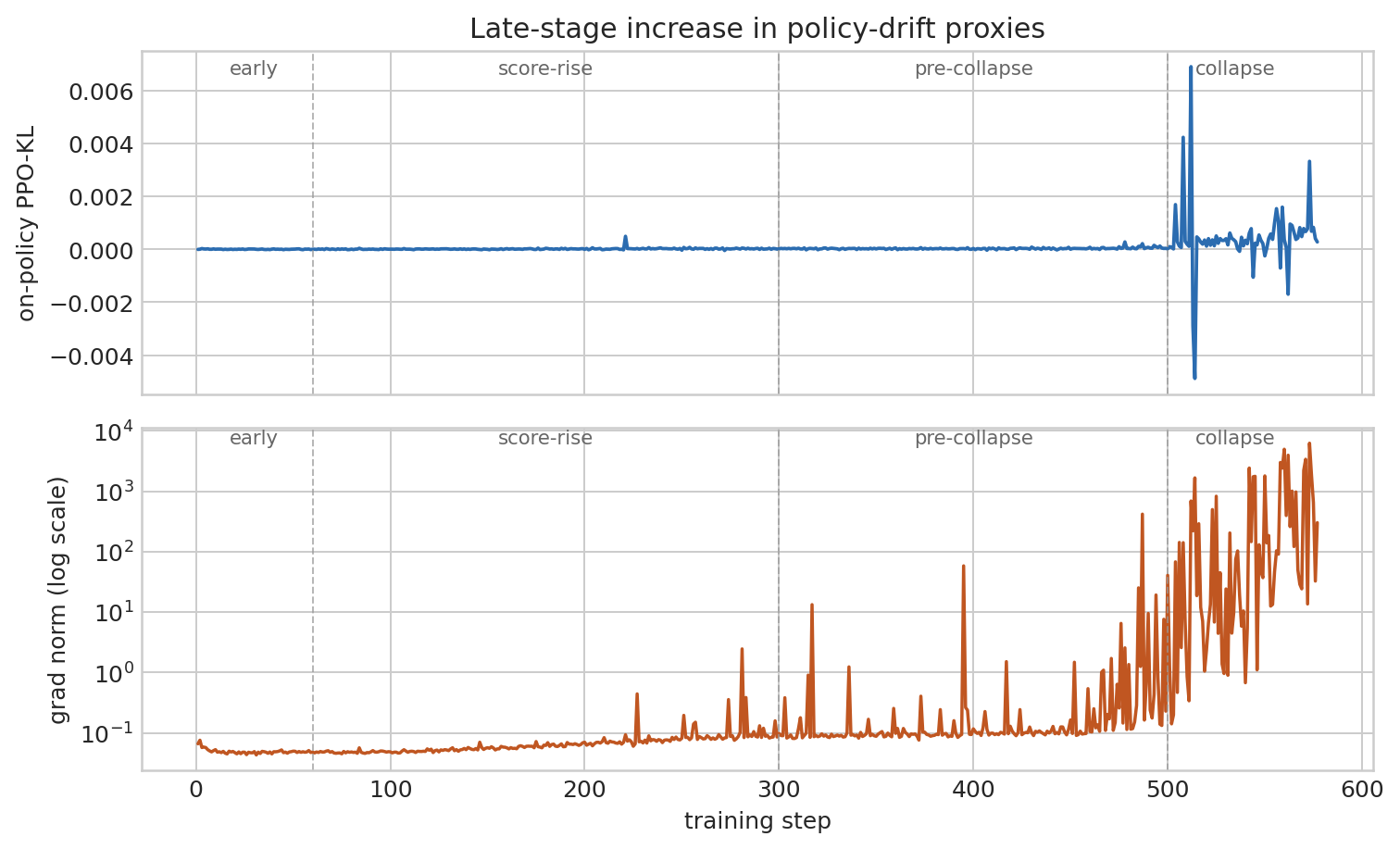}{0.90}{Supplementary policy-drift proxies over training. The upper panel shows on-policy GRPO-KL, and the lower panel shows gradient norm on a log scale. Both quantities rise sharply in the late stage, consistent with a phase in which the current policy begins to move more abruptly while the Drift-gate-admissible replay subset simultaneously contracts.}

This transition also co-occurs with larger policy movement in the late stage. As shown in Supplementary Figure~\ref{fig:qwen25-math-policy-drift-proxy}, on-policy GRPO-KL and gradient norm rise together during collapse. At the same time, the main logs show a sharp drop in Drift-gate acceptance rate relative to pre-collapse, along with repeated steps that fail to fill the replay target. This combination is consistent with increasing mismatch between the current policy and replayed trajectories, which shrinks the replay groups admissible under the Policy Drift threshold.

Taken together, the late-stage failure in this run is consistent with a phase in which the near-policy corrective mechanism can no longer be sustained in sufficiently rich form. In the early-to-middle stage, replay appears to combine useful trajectories from multiple recent policy steps, coinciding with both score improvement and delayed entropy collapse. In the late stage, replay ingress remains present, but the used replay subset drops sharply and its internal source-age mixture contracts into a narrow age-1/2-dominant range. We therefore interpret the transition as being linked less to replay presence per se and more to the duration over which a usable corrective mixture can be maintained.

\clearpage

\section{Implementation details for headroom-drift replay}
\label{app:impl-details}

This appendix provides implementation-level detail for the replay-augmented training procedure described in Algorithm~\ref{alg:hdr_grpo_patch}. The main text focuses on the conceptual design of replay selection (Headroom ranking, Policy Drift gating, FIFO eviction); this section supplements it with the engineering choices required to run the system stably in practice. Algorithm~\ref{alg:impl-details} expands each step of Algorithm~\ref{alg:hdr_grpo_patch} to this level, with concrete instantiations from the mathematical reasoning setting (\(|\widetilde{\mathbf{R}}_t|{=}128\) replay groups, \(|\mathcal{G}_t^{\mathrm{on}}|{=}256\) on-policy groups, mini-batch size 64, 8 GPUs).

\begin{algorithm}[htbp]
  \caption{Two-column implementation view of Algorithm~\ref{alg:hdr_grpo_patch}. Left: pseudocode. Right: concrete instantiation in the mathematical reasoning setting.}
  \label{alg:impl-details}
  \footnotesize
  \setlength{\tabcolsep}{4pt}
  \renewcommand{\arraystretch}{1.08}
  \begin{tabular}{p{0.58\linewidth} p{0.38\linewidth}}
    \textbf{Left: General Procedure (Pseudocode)} & \textbf{Right: Concrete Instantiation (Math Setting)} \\

    \begin{minipage}[t]{\linewidth}\vspace{0pt}
      \textbf{A. Rollout, Reward, Advantage}\\
      Same as Algorithm~\ref{alg:hdr_grpo_patch}.
    \end{minipage}
    &
    \begin{minipage}[t]{\linewidth}\vspace{0pt}
      \textbf{A/C. Shared with Alg.~\ref{alg:hdr_grpo_patch}}\\
      Rollout, reward, advantage, ingress, and retrieval are unchanged.
    \end{minipage} \\

    \begin{minipage}[t]{\linewidth}\vspace{0pt}
      \textbf{B. Frozen Reference State}\\
      On-policy groups: compute old log-probs via forward pass.\\
      Replay groups: reuse stored $\log\pi_{\mathrm{gen}(g)}$.\\
      Replay advantages: reuse stored $A_i$.
    \end{minipage}
    &
    \begin{minipage}[t]{\linewidth}\vspace{0pt}
      \textbf{B. Frozen references}\\
      Replay groups: 128 groups reuse stored $\log\pi_{\mathrm{gen}(g)}$ and $A_i$ (no recomputation).
    \end{minipage} \\

    \begin{minipage}[t]{\linewidth}\vspace{0pt}
      \textbf{D. Selection with Early Termination}\\
      Sort buffer by cached Headroom (descending).\\
      Initialize $\mathrm{selected} \gets \emptyset$.\\
      \texttt{for} each candidate $g$ in sorted order \texttt{do}\\
      \quad \texttt{if} $|\mathrm{selected}| \ge K_{\mathrm{rep}}$ \texttt{then break}\\
      \quad Compute $\log\pi_{\theta_t}$ for $g$.\\
      \quad Compute $\mathrm{Drift}(g;\theta_t)$ using current and stored log-probs.\\
      \quad Refresh cached Headroom from the same current-policy pass.\\
      \quad \texttt{if} $\mathrm{Drift}(g;\theta_t) \le \tau$ append $g$, else reject $g$ (cached Headroom still refreshed).\\
      \texttt{end for}
    \end{minipage}
    &
    \begin{minipage}[t]{\linewidth}\vspace{0pt}
      \textbf{D. Selection}\\
      Stop when $K_{\mathrm{rep}}{=}128$ is filled (early termination).\\
      Use one parallelized forward pass per scanned group to compute both Policy Drift and refreshed cached Headroom.
    \end{minipage} \\

    \begin{minipage}[t]{\linewidth}\vspace{0pt}
      \textbf{E. Batch Construction}\\
      $\widetilde{\mathbf{R}}_t \gets \mathrm{selected}[\,1 : u\lfloor m/u \rfloor\,]$.\\
      $\mathrm{actor\_batch} \gets [\,\widetilde{\mathbf{R}}_t \;;\; \mathcal{G}_t^{\mathrm{on}}\,]$.\\
      Split into $k$ mini-batches (preserving order).
    \end{minipage}
    &
    \begin{minipage}[t]{\linewidth}\vspace{0pt}
      \textbf{E. Batch construction}\\
      $u{=}1$ (no truncation).\\
      $[128\ \text{replay};\ 256\ \text{on-policy}] = 384$ total groups.\\
      $k{=}6$ mini-batches: MB 1--2 replay, MB 3--6 on-policy.
    \end{minipage} \\

    \begin{minipage}[t]{\linewidth}\vspace{0pt}
      \textbf{F. Sequential PPO Update + Eviction}\\
      \texttt{for} $i = 1, \dots, k$ \texttt{do}\\
      \quad Partition MB$_i$ across GPUs with length balancing.\\
      \quad Compute $\log\pi_\theta$ on stored trajectories.\\
      \quad Compute $r = \pi_\theta / \pi_{\mathrm{gen}(g)}$, clipped PPO objective, and update $\theta$.\\
      \texttt{end for}\\
      Append ingress groups and apply FIFO eviction.
    \end{minipage}
    &
    \begin{minipage}[t]{\linewidth}\vspace{0pt}
      \textbf{F. PPO update}\\
      Compute numerator $\log\pi_\theta$ only, then clipped PPO update; ingress append + FIFO eviction after update.
    \end{minipage}
  \end{tabular}
\end{algorithm}

\subsection{Frozen reference state}

When a replay group is admitted to the buffer, its generation-time per-token log-probabilities \(\log\pi_{\mathrm{gen}(g)}(a_{i,j}\mid s_{i,j})\) and response-level advantages \(A_i\) are stored as immutable state (Algorithm~\ref{alg:impl-details}, Frozen Reference State). The importance-ratio denominator is therefore always anchored at the generation-time policy \(\pi_{\mathrm{gen}(g)}\); only the numerator is recomputed under the current policy \(\pi_{\theta_t}\) at each PPO update. Advantages are likewise reused without recomputation. As described in Section~\ref{subsec:replay-components}, the Policy Drift gate measures how far this frozen reference state has drifted from the current policy and blocks groups that exceed threshold \(\tau\).

\paragraph{Replay-buffer storage.}
In the mathematical-reasoning configuration, the CPU-resident FIFO buffer holds at most 512 query groups (8,192 responses). At the configured maximum sequence lengths, the stored token sequences, generation-time log-probabilities, and GRPO advantages total approximately 448 MiB at full capacity.

\subsection{Selection with early termination and cached-headroom refresh}

Replay selection scans buffer candidates in descending cached-Headroom order and accepts groups satisfying \(\mathrm{Drift}(g;\theta_t) \le \tau\) until the replay budget \(K_{\mathrm{rep}}\) is filled, at which point the scan terminates immediately (Algorithm~\ref{alg:impl-details}, Selection with Early Termination). This early termination ensures that selection cost scales with \(K_{\mathrm{rep}}\) and the number of candidates actually scanned, rather than with the full buffer size. When the budget cannot be filled---for instance, when \(\tau\) is tight and many candidates are rejected---the accepted count is used as-is.

For each scanned candidate, the current-policy forward pass---which is parallelized over stored trajectories---simultaneously computes Policy Drift and the runtime Headroom needed to refresh cached Headroom (Section~\ref{subsec:replay-components}). This cached-Headroom refresh is applied to every scanned group regardless of whether it passes or fails the Policy Drift gate, and requires no additional forward pass. As a result, the selection procedure performs Policy Drift gating and cached-Headroom refresh in a single same-pass reevaluation.

\subsection{Replay-prefix layout}

When forming the mixed actor batch \(\mathcal{A}_t\), replay groups are placed before on-policy groups (Algorithm~\ref{alg:impl-details}, Batch Construction). Because the actor batch is split into mini-batches by sequential slicing, this prefix layout causes replay groups to occupy the earlier mini-batches and on-policy groups to occupy the later ones. In the mathematical reasoning setting, for example, the six mini-batches are partitioned as MB~1--2 (replay) and MB~3--6 (on-policy). Since PPO processes mini-batches sequentially and updates the policy parameters after each one, the relatively more off-policy replay samples are consumed first within each training step.

\subsection{Batch alignment}

\paragraph{Micro-batch truncation.}
The number of accepted replay groups may not be a multiple of the micro-batch processing unit \(u\). To avoid misalignment during gradient accumulation, the accepted sequence is truncated to the largest multiple of \(u\) that does not exceed the accepted count (Algorithm~\ref{alg:impl-details}, Batch Construction). Formally, given the accepted replay sequence \(\mathbf{R}_t = (g_1, \dots, g_m)\), the groups actually used for training are
\[
  \widetilde{\mathbf{R}}_t = (g_1, \dots, g_{u\lfloor m/u \rfloor}).
\]
Because cached-Headroom ordering is preserved, only the lowest-priority tail is removed. The number of discarded groups is at most \(u - 1\); when \(u = 1\), no truncation occurs.

\paragraph{Sequence-length balanced partitioning.}
Each mini-batch is distributed across GPUs for data-parallel processing (Algorithm~\ref{alg:impl-details}, Sequential PPO Update). A naive sequential partition can concentrate long sequences on a single GPU, leaving other GPUs idle while that GPU finishes. This imbalance is amplified in the mixed batch because replay groups and on-policy groups are generated at different training steps and may have different response-length distributions. To mitigate this, each mini-batch is repartitioned so that the total sequence length per GPU is approximately balanced. In practice, this reduces the maximum-minus-minimum sequence-length gap across GPUs from approximately 73K tokens to under 5K tokens. This repartitioning does not change which samples are trained on or the objective used; it affects only the mapping of samples to GPUs within each mini-batch, reducing idle time caused by uneven workloads.

\clearpage

\section{Ablation studies}

\subsection{L1-style versus L2-style policy drift: formal interpretation and empirical comparison}
\label{app:l1-vs-l2-ablation}

The main text adopts a squared (L2-style) aggregation for Policy Drift and notes that an L1 variant yields a tighter formal bound but underperforms empirically (\S\ref{subsec:replay-components}). This appendix develops both sides of that claim: we first show that the two gates define qualitatively different compatibility regions, and then compare them empirically on mathematical reasoning.

\subsubsection{Formal interpretation: variance-sensitive gating}
\label{app:squared-drift-geometry}

The main method defines Policy Drift as
\[
  \mathrm{Drift}_{L2}(g;\theta_t)
  :=
  \frac{1}{|\mathcal{T}(g)|}
  \sum_{(i,j)\in\mathcal{T}(g)}
  \Delta_{i,j}(g;\theta_t)^2.
\]
The corresponding L1-style alternative replaces the squared terms with absolute values:
\[
  \mathrm{Drift}_{L1}(g;\theta_t)
  :=
  \frac{1}{|\mathcal{T}(g)|}
  \sum_{(i,j)\in\mathcal{T}(g)}
  \left|\Delta_{i,j}(g;\theta_t)\right|.
\]
A natural question is whether the squared gate is simply a stricter version of the absolute gate. It is not. The squared gate is better understood as a \emph{variance-sensitive} compatibility score.

\paragraph{Proposition (mean--variance decomposition).}
Let $\mu_g$ denote the mean absolute tokenwise drift and $v_g$ the within-group variance of absolute tokenwise drift. Then
\[
  \mathrm{Drift}_{L2}(g;\theta_t) = \mu_g^2 + v_g.
\]

\emph{Proof.} Immediate from the standard second-moment identity $\mathbb{E}[X^2] = (\mathbb{E}[X])^2 + \mathrm{Var}(X)$.

\paragraph{Corollary (acceptance-region geometry).}
An L1 gate with threshold $\tau_1$ accepts a group whenever $\mu_g \le \tau_1$. The L2 gate with threshold $\tau_2$ accepts whenever $\mu_g^2 + v_g \le \tau_2$. The L2 gate is therefore not uniformly stricter: it may reject a group with small mean drift if that drift is concentrated on a few tokens, while accepting another group with larger mean drift if the tokenwise mismatch is diffuse.

\paragraph{Illustrative example.}
Consider two four-token groups with identical L1 drift:
\[
  (0.02,\; 0.02,\; 0.02,\; 0.02)
  \quad\text{vs.}\quad
  (0.08,\; 0,\; 0,\; 0).
\]
Both have $\mathrm{Drift}_{L1} = 0.02$, but $\mathrm{Drift}_{L2} = 0.0004$ for the uniform case and $0.0016$ for the spike case---a fourfold difference invisible to the L1 gate.

In our experiments, the L1 threshold is $\tau_1 = 0.005$ and the L2 threshold is $\tau_2 = 0.001$. Because the L2 gate operates on a second moment, its threshold is more naturally compared on the root-mean-square scale: $\sqrt{\tau_2} \approx 0.032$. The two gates therefore define different compatibility regions rather than tighter or looser versions of the same region.

\subsubsection{Empirical comparison on mathematical reasoning}
\label{app:squared-drift-empirics}

We compare L1-style and L2-style Drift gating under otherwise identical training configurations on mathematical reasoning. Figure~\ref{fig:m1m2-gating-trends} summarizes step-aligned trajectories over 500 steps. Figures~\ref{fig:m2-selection-dynamics}--\ref{fig:m2-selection-snapshots} characterize the replay groups selected by the L2 gate.

\replayfigure[fig:m1m2-gating-trends]{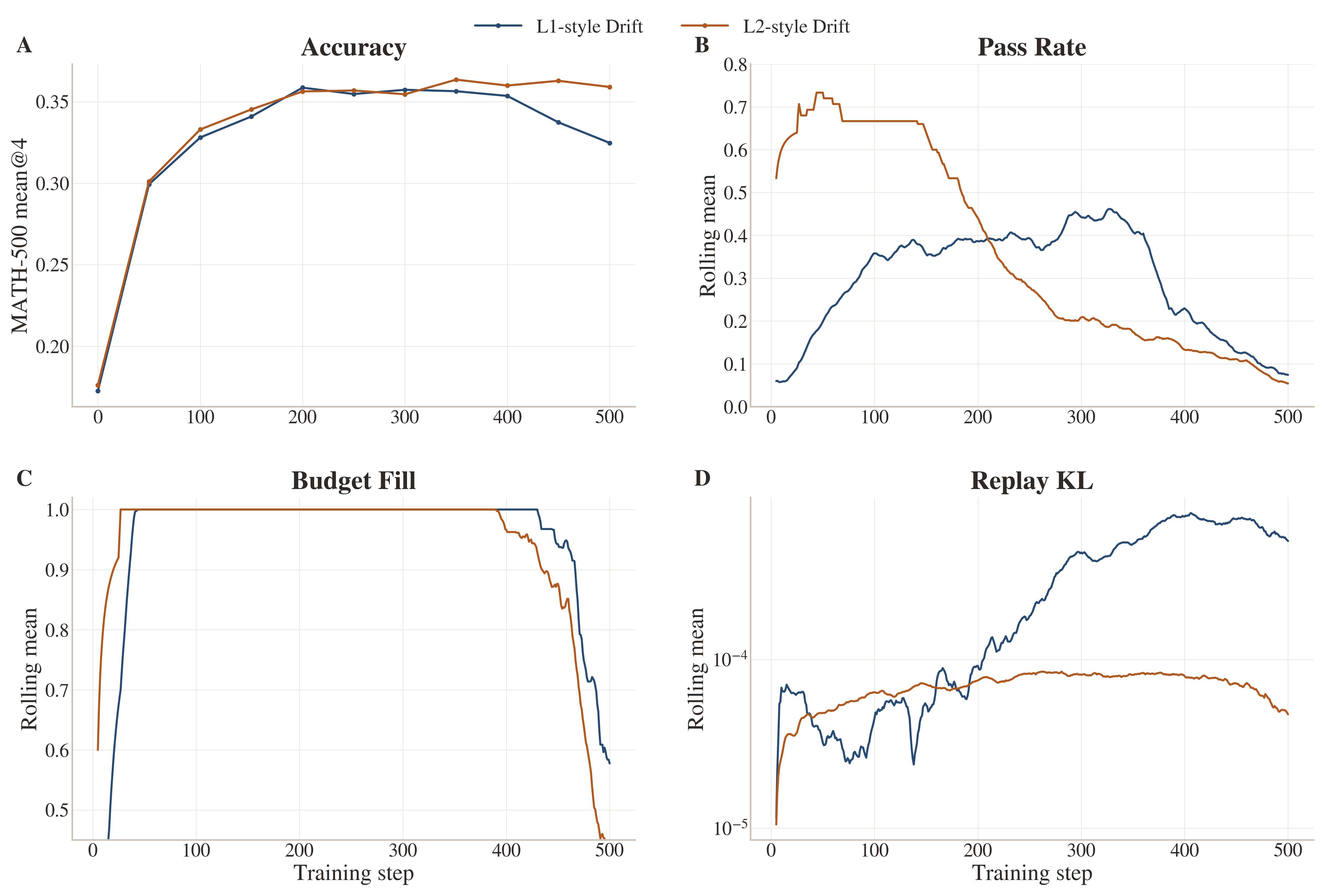}{0.92}{Step-aligned comparison of L1-style and L2-style Drift gating. L2 attains better late-stage validation accuracy while maintaining lower replay mismatch, despite matched or lower replay-budget fulfillment.}

\paragraph{L2 wins through selection quality, not replay volume.}
The two gates achieve similar replay-budget fulfillment through the middle phase of training. At step 350, both fill the full budget, yet L2 already yields higher validation accuracy (Mean@4: 0.3636 vs.\ 0.3565) with an order-of-magnitude lower replay KL ($8.5 \times 10^{-5}$ vs.\ $5.9 \times 10^{-4}$). The gap widens in the late phase. Averaged over steps 350--500, L2 achieves higher Mean@4 (0.3613 vs.\ 0.3430) despite lower average budget fulfillment (0.81 vs.\ 0.88), while maintaining replay KL roughly nine times smaller ($7.0 \times 10^{-5}$ vs.\ $6.3 \times 10^{-4}$). L2 does not use more replay; it uses better-compatible replay.

\replayfigure[fig:m2-selection-dynamics]{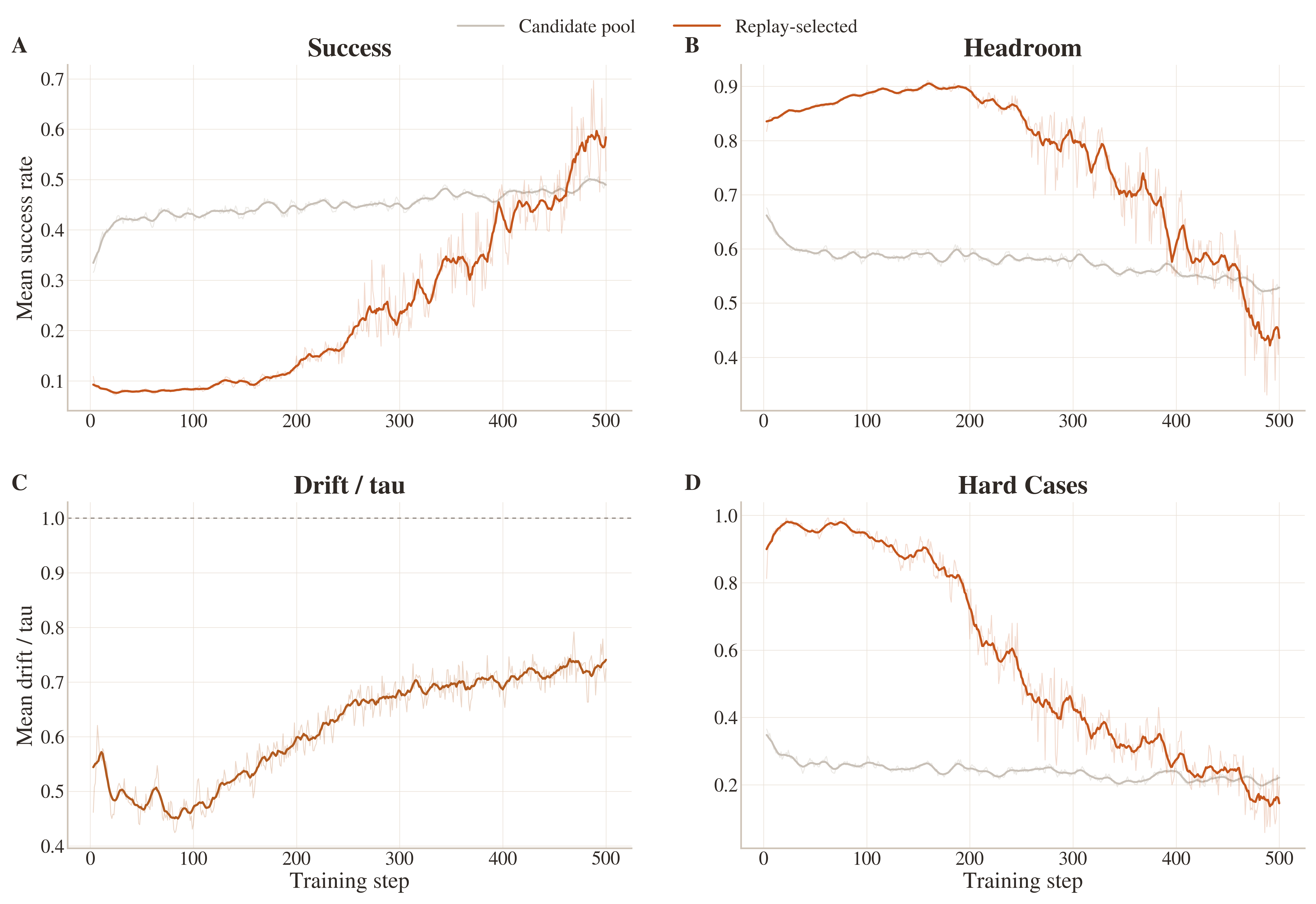}{0.92}{Aggregate properties of the replay groups selected by L2-style Drift gating. Early training concentrates replay on hard, high-Headroom groups; later training moves toward the remaining frontier, and replay usage contracts as that frontier is depleted.}

\paragraph{Selection dynamics within the L2 gate.}
Figure~\ref{fig:m2-selection-dynamics} shows how the L2-selected replay subset evolves. Early in training (step 50), the selected groups have much lower success rate (0.080 vs.\ 0.423 in the candidate pool) and higher Headroom (0.863 vs.\ 0.592), indicating strong concentration on hard, high-value cases. This concentration weakens as training progresses: by step 350, the selected subset approaches the pool boundary (success rate 0.390 vs.\ 0.474 and Headroom 0.643 vs.\ 0.555), and by step 500 the gap has largely closed. Once the high-value frontier is depleted, the L2 gate contracts replay usage rather than forcing the budget full with less compatible groups.

\replayfigure[fig:m2-selection-snapshots]{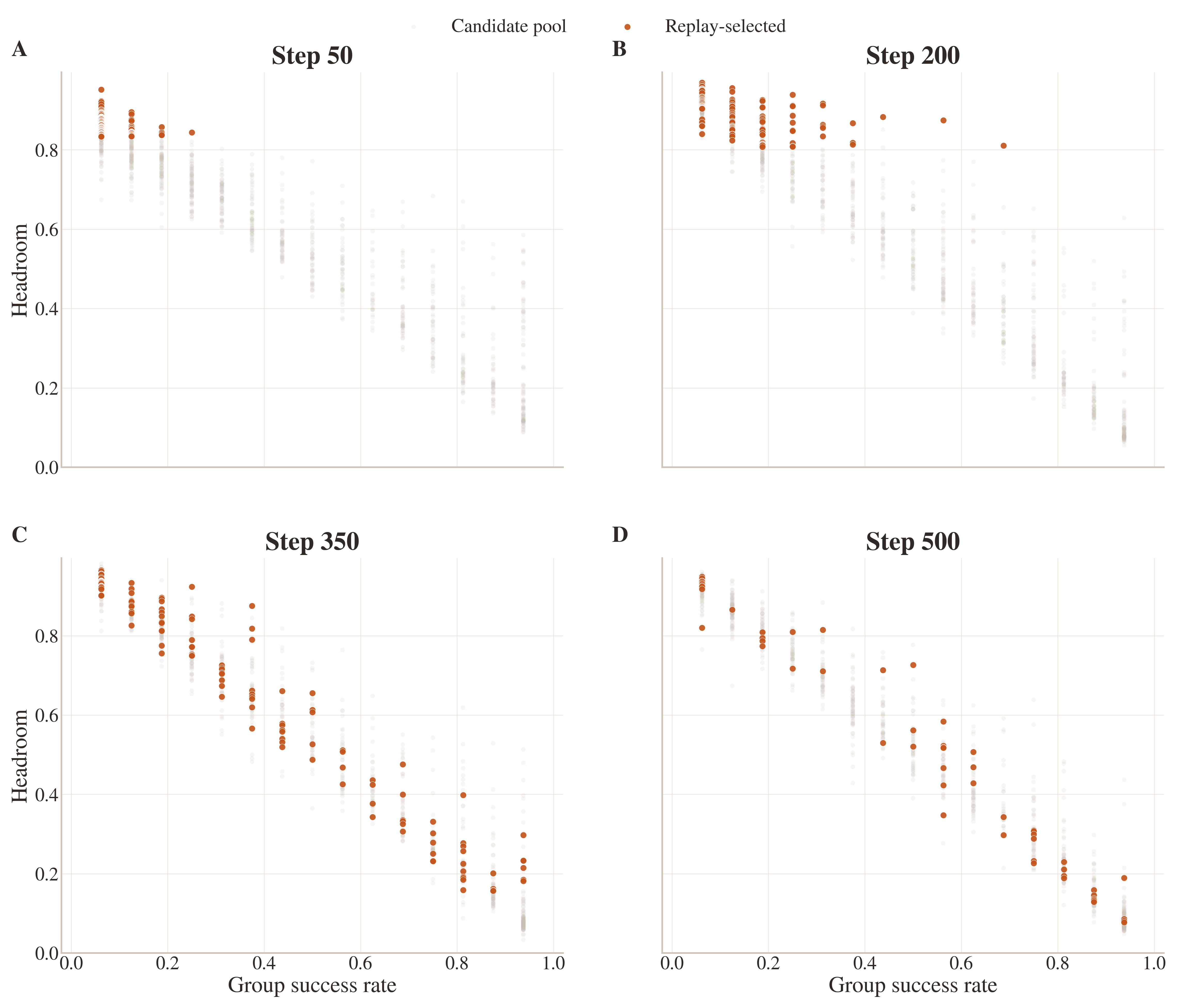}{0.92}{Selection snapshots in the success-rate / Headroom plane at four training stages. Selected groups (orange) concentrate near the upper-left frontier of the candidate cloud (gray), particularly in early and middle training.}

Figure~\ref{fig:m2-selection-snapshots} visualizes this pattern as snapshots. At steps 50 and 200, selected groups cluster tightly in the low-success, high-Headroom corner. By step 350, coverage broadens as the high-value region thins. By step 500, selection spans a wider range of success rates, reflecting the depleted frontier. The L2 gate adapts its effective selectivity to the available pool rather than maintaining a fixed selection profile.

\clearpage
\subsection{Practical calibration and sensitivity of the Policy Drift threshold \texorpdfstring{$\tau$}{τ}}
\label{app:tau-sensitivity}

We use \(\tau\) to place the Policy Drift gate in a viable operating regime, not to locate a narrow performance optimum. Table~\ref{tab:tau-settings} summarizes the thresholds used across the evaluated runs. The values fall into two task-family settings: the single-turn reasoning runs and the CISPO-style portability study share one value, while the 3B and 7B Search-R1 runs share another. In particular, the 7B Search-R1 run reuses the 3B threshold without scale-specific retuning.

\begin{table}[htbp]
  \centering
  \small
  \setlength{\tabcolsep}{4pt}
  \renewcommand{\arraystretch}{1.12}
  \begin{tabular}{L{0.25\linewidth}L{0.21\linewidth}cL{0.30\linewidth}}
    \toprule
    Evaluated setting & Scope & \(\tau\) & Reuse note \\
    \midrule
    Mathematical reasoning & Single-turn reasoning & \(10^{-3}\) & Shared single-turn setting \\
    Multimodal reasoning & Single-turn reasoning & \(10^{-3}\) & Shared single-turn setting \\
    CISPO-style portability & Objective portability & \(10^{-3}\) & Replay hyperparameters unchanged \\
    Search-R1 (3B) & Agentic Search & \(10^{-2}\) & Agentic setting \\
    Search-R1 (7B) & Agentic scale check & \(10^{-2}\) & Reused without scale-specific retuning \\
    \bottomrule
  \end{tabular}
  \caption{Drift thresholds used across the evaluated settings. The values reduce to two task-family settings, and the 7B Search-R1 run reuses the 3B threshold without scale-specific retuning.}
  \label{tab:tau-settings}
\end{table}

We then use a controlled Search-R1 sweep to examine how a viable regime can be identified without a dense full-training search. Agentic Search is a useful testbed because its multi-turn interaction with an external retrieval environment makes replay valuable while also exposing training instability more clearly than the single-turn settings \citep{igpo2026,istar2026,gem2026}.

\begin{rqbox}
  \textbf{RQ.} \textbf{How can a useful operating regime for \(\tau\) be identified without an exhaustive full-training sweep, and what diagnostic signatures appear when the gate is too tight or too loose?}
\end{rqbox}

\subsubsection{Experimental setup}

Experiments are conducted on Agentic Search (Qwen2.5-3B-Instruct, 8\(\times\)H100). The three decade-spaced conditions are designed to probe distinct gate states rather than locate a narrow performance optimum. All other training settings---including batch size, learning rate, replay buffer size, \(n\), and \(K_{\mathrm{rep}}\)---are held fixed.

\subsubsection{Gate behavior and calibration diagnostics}

We interpret replay acceptance and replay KL jointly: acceptance measures whether the gate supplies enough replay, while replay KL measures whether the admitted groups remain compatible with the current policy. Table~\ref{tab:tau-diagnostic-summary} summarizes the three conditions.

\begin{table}[H]
  \centering
  \small
  \setlength{\tabcolsep}{3pt}
  \renewcommand{\arraystretch}{1.12}
  \begin{tabular}{L{0.12\linewidth}c c c L{0.30\linewidth}}
    \toprule
    Condition & \(\tau\) & Mean acceptance & \shortstack{Gate rejections\\per update} & Replay-KL behavior \\
    \midrule
    \textsc{tight} & 0.001 & 6.4\% & 450.3 & \(0.0005 \rightarrow 0.00003\) as replay vanishes \\
    \textsc{moderate} & 0.01 & 40.2\% & 119.4 & Stable at approximately \(0.002\) \\
    \textsc{loose} & 0.1 & 94.4\% & 1.9 & \(0.0045 \rightarrow 0.0170\) \\
    \bottomrule
  \end{tabular}
  \caption{Diagnostic summary of the Search-R1 threshold sweep. Acceptance rate and gate rejections per update are training averages; replay-KL entries summarize the early-to-final trajectory.}
  \label{tab:tau-diagnostic-summary}
\end{table}

The three settings produce qualitatively different gate states. Under \textsc{tight}, only 6.4\% of candidates are accepted, effectively starving replay. Its low replay KL does not indicate better compatibility; it results from admitting very few replay groups. Under \textsc{loose}, 94.4\% of candidates pass, leaving the gate nearly inactive, and replay KL rises over training. \textsc{moderate} instead admits replay selectively while keeping replay KL stable at approximately 0.002. Neither diagnostic is sufficient alone: a useful threshold must preserve replay supply and control policy mismatch at the same time.

\subsubsection{Accumulation of distribution mismatch}

The clearest difference across \(\tau\) settings appears in replay KL. Replay KL is computed at each training step over the replay samples and measures how far the current policy \(\pi_\theta\) has drifted from the policy \(\pi_{\theta_{\mathrm{old}}}\) that generated each replay sample. Specifically, it is the mean of \(D_{\mathrm{KL}}(\pi_\theta \| \pi_{\theta_{\mathrm{old}}})\) across token positions in the replay batch. A higher value indicates that the replay data lies farther from the current policy distribution. This replay KL is a related but distinct diagnostic from the Policy Drift gate quantity itself: the gate uses token-level squared log-probability drift on stored actions (the Policy Drift score), whereas replay KL summarizes mismatch over the accepted replay batch.

\begin{figure}[htbp]
  \centering
  \includegraphics[width=0.85\linewidth]{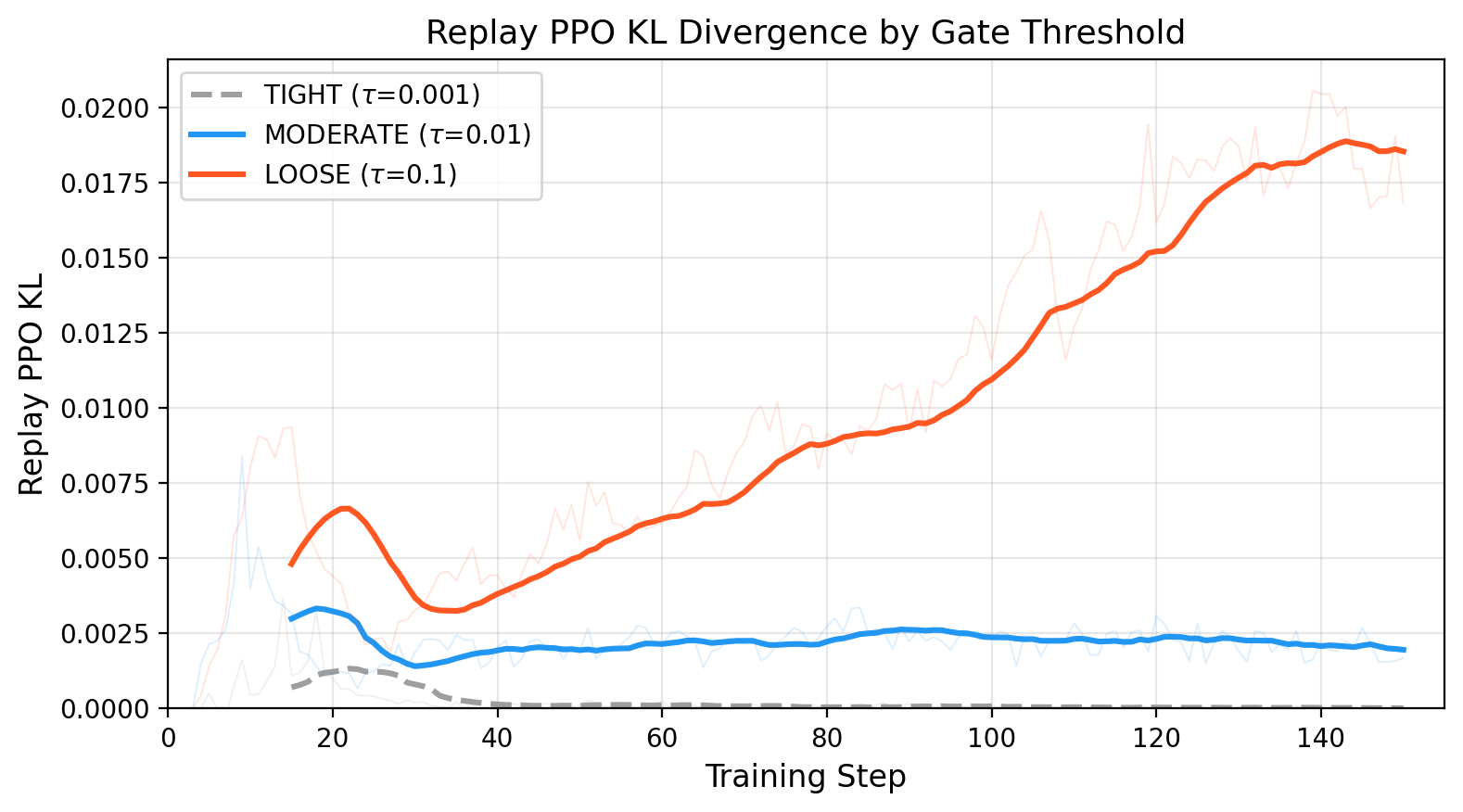}
  \caption{Replay KL over training for the three \(\tau\) conditions (running average, window\,=\,15). Under \textsc{moderate}, replay KL remains stable throughout training. Under \textsc{loose}, it increases monotonically.}
  \label{fig:tau-replay-kl}
\end{figure}

Figure~\ref{fig:tau-replay-kl} shows the replay KL trajectories, and Table~\ref{tab:tau-kl} reports interval-wise averages.

\begin{table}[htbp]
  \centering
  \small
  \setlength{\tabcolsep}{5pt}
  \renewcommand{\arraystretch}{1.12}
  \begin{tabular}{lcccc}
    \toprule
    Interval & \textsc{tight} & \textsc{moderate} & \textsc{loose} & \textsc{loose}\,/\,\textsc{mod.} \\
    \midrule
    Early interval  & 0.0005  & 0.0021 & 0.0045 & 2.1$\times$ \\
    Middle interval & 0.0001  & 0.0023 & 0.0089 & 3.8$\times$ \\
    Final interval  & 0.00003 & 0.0022 & 0.0170 & 7.7$\times$ \\
    \bottomrule
  \end{tabular}
  \caption{Mean replay KL by training interval. Under \textsc{moderate}, replay KL remains flat at ${\sim}0.002$ throughout training, whereas under \textsc{loose} it grows monotonically and reaches its largest gap from \textsc{moderate} in the final interval.}
  \label{tab:tau-kl}
\end{table}

Under \textsc{moderate}, replay KL remains stable at approximately 0.002 throughout training. Under \textsc{loose}, it increases monotonically and reaches a large multiple of the \textsc{moderate} level in the final interval. This reflects the continued admission of replay data that has progressively diverged from the current policy, without filtering by the gate. We report this growth as a distributional diagnostic; the extent to which it translates into gradient-level effects after clipping is not isolated in this analysis.

The replay KL under \textsc{tight} falls below 0.0001 late in training. This does not indicate superior distribution compatibility; rather, it reflects the scarcity of replay samples admitted to the training batch.

\subsubsection{Performance and entropy dynamics}

\begin{figure}[htbp]
  \centering
  \includegraphics[width=0.85\linewidth]{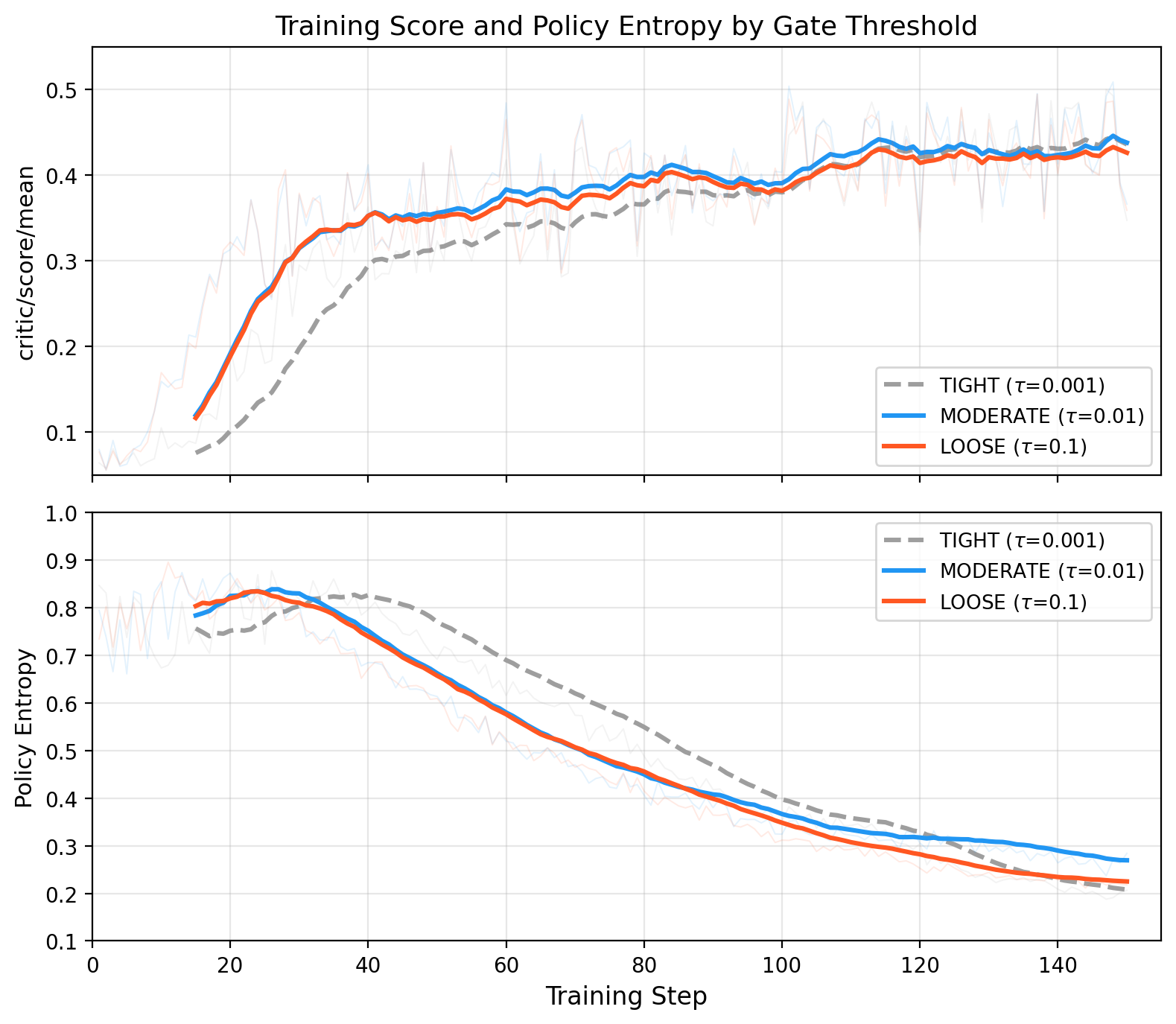}
  \caption{Training score (upper) and policy entropy (lower) over training for the three \(\tau\) conditions (running average, window\,=\,15).}
  \label{fig:tau-score-entropy}
\end{figure}

Figure~\ref{fig:tau-score-entropy} compares the training score and policy entropy trajectories, and Table~\ref{tab:tau-score-entropy} reports interval-wise averages.

\begin{table}[htbp]
  \centering
  \small
  \setlength{\tabcolsep}{5pt}
  \renewcommand{\arraystretch}{1.12}
  \begin{tabular}{l cc cc cc}
    \toprule
    & \multicolumn{2}{c}{\textsc{tight} (\(\tau{=}0.001\))} & \multicolumn{2}{c}{\textsc{moderate} (\(\tau{=}0.01\))} & \multicolumn{2}{c}{\textsc{loose} (\(\tau{=}0.1\))} \\
    \cmidrule(lr){2-3} \cmidrule(lr){4-5} \cmidrule(lr){6-7}
    Interval & Score & Entropy & Score & Entropy & Score & Entropy \\
    \midrule
    Early interval  & 0.206 & 0.783 & 0.272 & 0.757 & 0.269 & 0.756 \\
    Middle interval & 0.365 & 0.527 & 0.392 & 0.447 & 0.382 & 0.441 \\
    Final interval  & 0.432 & 0.271 & 0.434 & 0.300 & 0.424 & 0.255 \\
    \bottomrule
  \end{tabular}
  \caption{Training score and policy entropy by interval for the three \(\tau\) conditions.}
  \label{tab:tau-score-entropy}
\end{table}

\textsc{tight} records the lowest score in the middle interval (0.365 vs.\ 0.392 for \textsc{moderate} and 0.382 for \textsc{loose}). This appears to be due to the near-absence of replay, which prevents the run from benefiting from replay-based acceleration. \textsc{tight} approaches \textsc{moderate} by the final interval, suggesting that on-policy learning alone can converge gradually, though with slower progress in the early-to-mid phase.

The score gap between \textsc{moderate} and \textsc{loose} is approximately 1 percentage point, which is modest in absolute terms but consistent across all intervals in favor of \textsc{moderate}.

All three conditions exhibit a common decreasing trend in entropy over the course of training. In the final interval, the ordering of entropy across conditions (\textsc{moderate} 0.300 $>$ \textsc{tight} 0.271 $>$ \textsc{loose} 0.255) coincides with the ordering of score (\textsc{moderate} 0.434 $>$ \textsc{tight} 0.432 $>$ \textsc{loose} 0.424). \textsc{moderate} maintains both the highest score and the highest entropy among the three conditions at this stage. We note that this alignment does not imply a simple monotonic relationship between entropy and score across all training phases: in the middle interval, \textsc{tight} exhibits the highest entropy yet the lowest score, reflecting slow learning due to the near-absence of replay rather than a beneficial preservation of exploration. Nonetheless, the co-occurrence of faster entropy decline and lower score under \textsc{loose} is consistent with prior findings that rapid entropy reduction in LLM reinforcement learning tends to be associated with degraded learning outcomes \citep{petrenko2026entropy,liu2026entropy,yu2025dapo}.

\subsubsection{Summary}

\begin{rqbox}
  \textbf{Answer to RQ.} \textbf{A short log-spaced sweep is sufficient to identify a useful operating regime for \(\tau\). Collapsing acceptance rules out thresholds that starve replay, while rising replay KL rules out thresholds that leave the gate too permissive. The remaining regime admits replay selectively and keeps replay KL stable. Because these signals are available before final performance is known, unsuitable candidates need not be trained to completion. Together with threshold reuse across settings and model scales, these results support treating \(\tau\) as a coarse calibration choice rather than a finely tuned performance parameter.}
\end{rqbox}

The same diagnostics could support an adaptive schedule that loosens the gate when replay is starved and tightens it when replay KL rises.

\subsection{Component ablation on mathematical reasoning}
\label{app:math-component-ablation}

Table~\ref{tab:math_ablation} reports a focused component ablation on MATH-500 at the evaluation endpoint. Headroom-only is the closest PER-style priority-only analogue in this group-level setting: it ranks buffered groups by Headroom without the current-policy compatibility gate. Under the same training configuration, the full Headroom-Drift combination records the highest endpoint values on both Best@32 and Mean@32, while Headroom-only and Policy-Drift-only are lower.

\begin{table}[H]
  \centering
  \small
  \setlength{\tabcolsep}{8pt}
  \renewcommand{\arraystretch}{1.12}
  \begin{tabular}{lcc}
    \toprule
    Variant & Best@32 ($\uparrow$) & Mean@32 ($\uparrow$) \\
    \midrule
    Policy Drift only & 0.7820 & 0.6794 \\
    Headroom only & 0.8100 & 0.7140 \\
    Headroom-Drift Replay (full) & \textbf{0.8200} & \textbf{0.7215} \\
    \bottomrule
  \end{tabular}
  \caption{MATH-500 ablation results at the evaluation endpoint. Arrows indicate metric direction ($\uparrow$ higher is better). Best@32 denotes the average of the highest score among 32 samples per input, and Mean@32 denotes the average score across all 32 samples per input.}
  \label{tab:math_ablation}
\end{table}

\vspace{-0.5em}

\begin{figure}[H]
  \centering
  \includegraphics[width=0.88\linewidth]{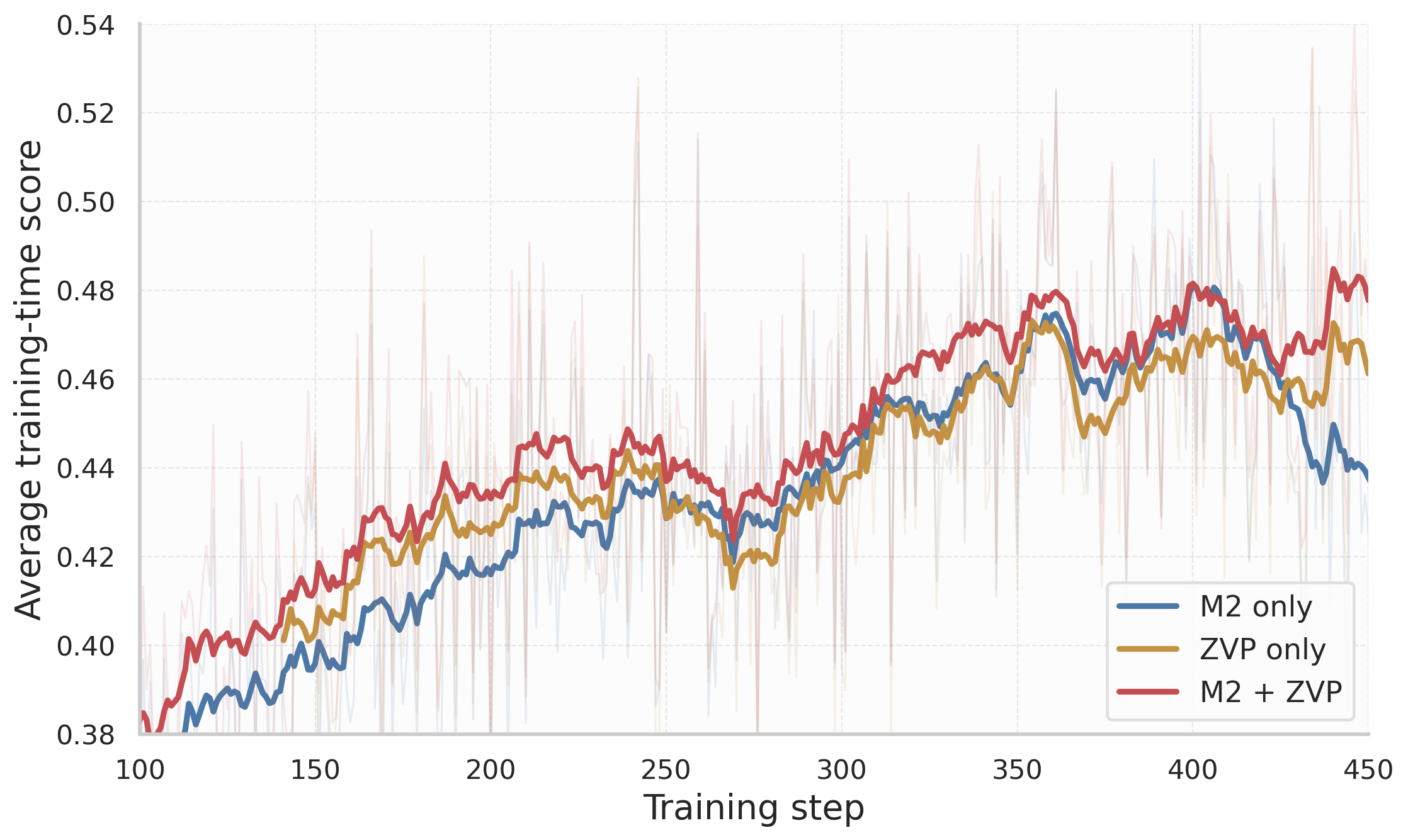}
  \caption{Training-score curves for the mathematical-reasoning component ablation. In the figure legend, M2 denotes Policy Drift only, ZVP denotes Headroom only, and M2 + ZVP denotes full Headroom-Drift Replay.}
  \label{fig:math_ablation_training_curves}
\end{figure}

Figure~\ref{fig:math_ablation_training_curves} provides the corresponding training-score trajectories for the same three variants. All three curves improve over training, and the late-stage ordering is consistent with Table~\ref{tab:math_ablation}: the full variant remains on top, Headroom-only follows closely, and Policy-Drift-only stays lower. The margin between full and Headroom-only is modest, while the gap to Policy-Drift-only is visibly larger in the later phase. These trajectories provide complementary context to the endpoint comparison.

\subsection{Sign-only versus advantage-weighted headroom}
\label{app:sign-vs-advmag-headroom}

The main text defines Headroom as a token-level score that depends only on the \emph{sign} of the response-level advantage $A_i$: tokens in positively-advantaged responses contribute $1-\pi(a_{i,j}\mid s_{i,j})$, tokens in negatively-advantaged responses contribute $\pi(a_{i,j}\mid s_{i,j})$, and the group score is a flat average over all tokens. A natural question is whether reweighting each response's contribution by $|A_i|$ would produce a better ranking.

\paragraph{What changes mathematically.}
Let $s_g$ denote the group success rate, and let $H_g^{+}$, $H_g^{-}$ denote the mean per-token headroom of the positive-advantage and negative-advantage responses respectively. Under equal response lengths, the two variants reduce to
\[
  Z_{\mathrm{sign}}(g)
  \;\approx\;
  s_g\,H_g^{+} \;+\; (1-s_g)\,H_g^{-},
  \qquad
  Z_{\mathrm{adv}}(g)
  \;\approx\;
  \tfrac{1}{2}\,H_g^{+} \;+\; \tfrac{1}{2}\,H_g^{-}.
\]
The approximation for $Z_{\mathrm{adv}}$ follows because, under group-centered binary rewards, positive responses carry $|A_i|\approx 1-s_g$ and negative responses carry $|A_i|\approx s_g$, so the count imbalance between the two sides is cancelled to first order by the advantage magnitudes.

The key difference is that $Z_{\mathrm{sign}}$ contains the success ratio $s_g$ directly as a mixing weight. In failure-heavy groups ($s_g \ll 0.5$), the negative branch dominates the score, and high-confidence wrong tokens (large $\pi$ on negatively-advantaged responses) can push the group to the top of the ranking. $Z_{\mathrm{adv}}$ removes this structural dependence on $s_g$, so groups are ranked more by branch-specific correction room than by failure ratio.

In short, the two variants do not define slightly different weightings of the same ranking; they define qualitatively different ranking geometries along the success-rate axis.

\paragraph{Empirical comparison.}
We conducted a small-scale preliminary experiment comparing the two variants under otherwise identical training configurations. Over the shared evaluation window (steps 1--38), sign-only Headroom achieves a higher average training score (0.3288 vs.\ 0.3022). Sign-only leads on 32 of 38 steps; advantage-weighted leads on 5. The gap is negligible in the earliest steps (at step 10: 0.2754 vs.\ 0.2783) but widens once replay becomes active. Over steps 19--38, where both runs use replay, sign-only leads on every step (mean 0.3863 vs.\ 0.3397). Representative checkpoints:

\begin{center}
\small
\setlength{\tabcolsep}{6pt}
\begin{tabular}{lccc}
  \toprule
  Step & Sign-only & Adv-weighted & $\Delta$ \\
  \midrule
  10 & 0.2754 & 0.2783 & $-$0.0029 \\
  20 & 0.3481 & 0.3108 & $+$0.0373 \\
  35 & 0.4177 & 0.3572 & $+$0.0605 \\
  38 & 0.4119 & 0.3501 & $+$0.0618 \\
  \bottomrule
\end{tabular}
\end{center}

The mathematical analysis above suggests a possible explanation. By removing the structural dependence on $s_g$, advantage weighting deprioritizes failure-heavy groups that sign-only would rank highly. In this setting, those failure-heavy groups appear to carry useful corrective signal; deprioritizing them reduces the effectiveness of replay. We retain sign-only Headroom in the main method accordingly.

\section{Preliminary portability to a CISPO-style objective}
\label{app:cispo-portability}

\begin{figure}[htbp]
  \centering
  \includegraphics[width=0.92\linewidth]{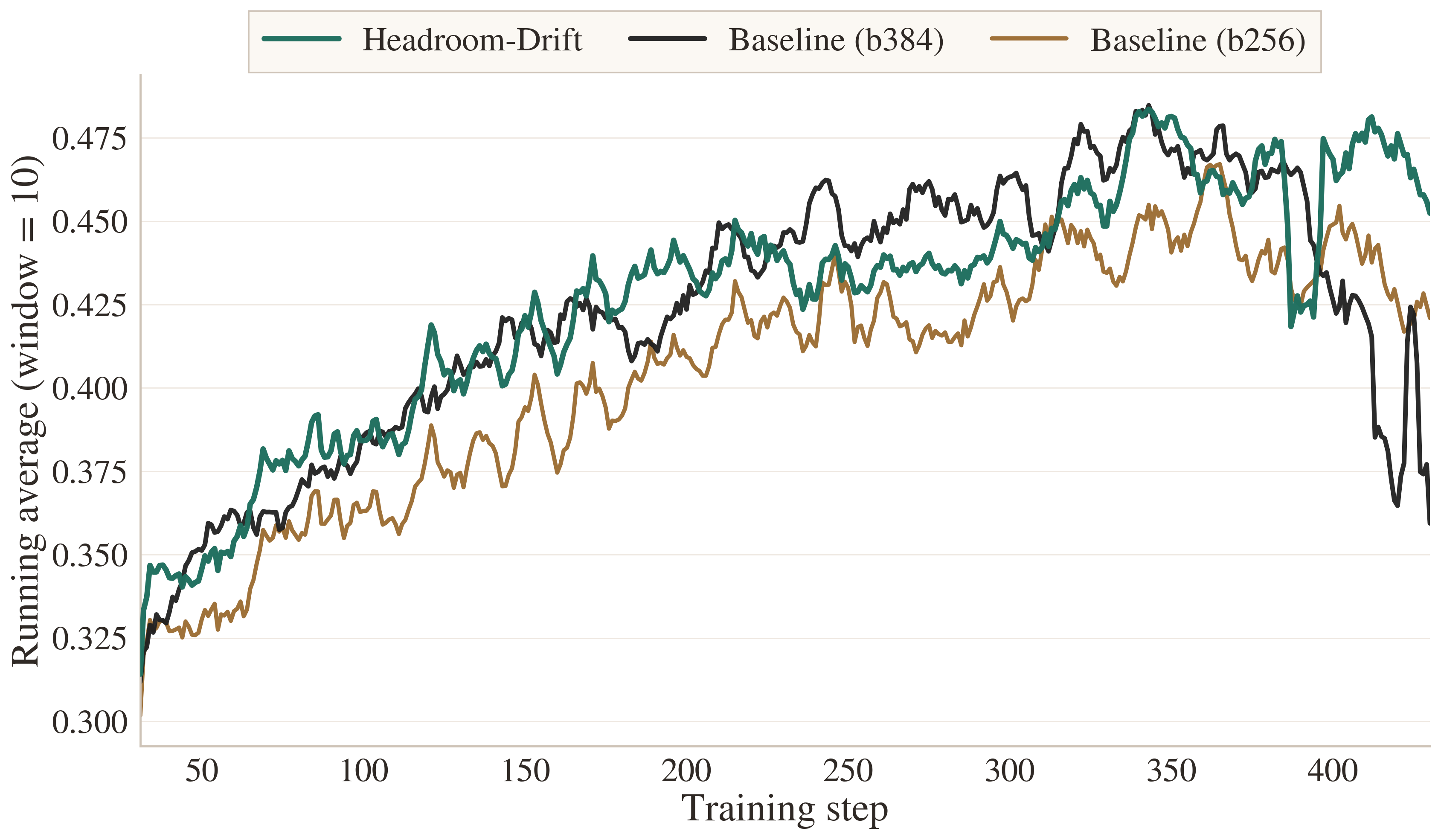}
  \caption{Training-score trajectories under a CISPO-style objective. 
  Headroom-Drift denotes full Headroom-Drift Replay (b256+r128). 
  Baseline (b256) is the on-policy matched control; 
  Baseline (b384) is the on-policy larger control. 
  Running average with window 10.}
  \label{fig:cispo_training_curves}
\end{figure}

The Conclusion identifies objective-level portability as an open question: whether the same two-axis decomposition transfers beyond GRPO-style clipped objectives. We provide preliminary training-score evidence on a CISPO-style objective using the same mathematical reasoning setup as the main experiments (Qwen2.5-Math-1.5B, identical dataset and evaluation protocol). The only change is the policy-optimization objective; Headroom ranking, Policy Drift gating, buffer mechanics, and all replay hyperparameters remain unchanged.

Figure~\ref{fig:cispo_training_curves} shows three patterns consistent with the GRPO results in the main text. First, Headroom-Drift Replay outperforms the on-policy matched baseline (b256) across nearly the entire training trajectory, confirming that principled replay adds value beyond the matched fresh-rollout budget under CISPO as well. Second, Headroom-Drift tracks above the on-policy larger baseline (b384) in the late phase and reaches a higher raw peak score (0.522 vs.\ 0.510 at steps 380 and 343 respectively), despite using fewer fresh responses. Third, the on-policy larger baseline exhibits a sharp score collapse after step 400, whereas the replay trajectory remains more stable through the same region---a pattern that mirrors the delayed entropy-collapse observation under GRPO (Appendix~\ref{app:entropy-collapse}).

\end{document}